\documentclass[lettersize,journal]{IEEEtran}
\usepackage{amsmath,amsfonts,bm}
\usepackage{algorithm}
\usepackage{algpseudocode}
\usepackage{array}
\usepackage[caption=false,font=normalsize,labelfont=sf,textfont=sf]{subfig}
\usepackage{textcomp}
\usepackage{stfloats}
\usepackage{url}
\usepackage{verbatim}
\usepackage{graphicx}
\usepackage{cite}

\usepackage{multirow}
\usepackage{booktabs}
\usepackage{xcolor}
\usepackage{colortbl}
\usepackage{hyperref}
\hypersetup{
  colorlinks=true,
  citecolor=green,
  linkcolor=red,
  urlcolor=blue}
\usepackage{amssymb}
\usepackage{pifont}
\usepackage{algpseudocode}
\algrenewcommand\algorithmicrequire{\textbf{Input:}}
\algrenewcommand\algorithmicensure{\textbf{Output:}}
\begin{document}

\title{Fine-grained Recognition with Learnable Semantic Data Augmentation}

\author{Yifan~Pu*,
        Yizeng~Han*,
        Yulin~Wang,
        Junlan~Feng,~\IEEEmembership{Fellow,~IEEE,}
        Chao~Deng,
        and~Gao~Huang,~\IEEEmembership{Member,~IEEE}
\thanks{Y. Pu, Y. Han, Y. Wang and G. Huang are with the Department of Automation, BNRist, Tsinghua University, Beijing 100084, China. 
Email: \{pyf20, hanyz18, wang-yl19\}@mails.tsinghua.edu.cn; gaohuang@tsinghua.edu.cn.}
\thanks{J. Feng and C. Deng are with the China Mobile Research Institute, Beijing 100053, China.
Email: \{fengjunlan, dengchao\}@chinamobile.com.}
\thanks{* Equal contrubution.}
}



\maketitle

\begin{abstract}
Fine-grained image recognition is a longstanding computer vision challenge that focuses on differentiating objects belonging to multiple subordinate categories within the same meta-category. Since images belonging to the same meta-category usually share similar visual appearances, mining discriminative visual cues is the key to distinguishing fine-grained categories. Although commonly used \emph{image-level} data augmentation techniques have achieved great success in generic image classification problems, they are rarely applied in fine-grained scenarios, because their \emph{random editing-region} behavior is prone to destroy the discriminative visual cues residing in the subtle regions. 
In this paper, we propose diversifying the training data at the \emph{feature-level} to alleviate the discriminative region loss problem. Specifically, we produce diversified augmented samples by translating image features along semantically meaningful directions. The semantic directions are estimated with a covariance prediction network, which predicts a sample-wise covariance matrix to adapt to the large intra-class variation inherent in fine-grained images. Furthermore, the covariance prediction network is jointly optimized with the classification network in a meta-learning manner to alleviate the degenerate solution problem.
Experiments on four competitive fine-grained recognition benchmarks (CUB-200-2011, Stanford Cars, FGVC Aircrafts, NABirds) demonstrate that our method significantly improves the generalization performance on several popular classification networks (e.g., ResNets, DenseNets, EfficientNets, RegNets and ViT). Combined with a recently proposed method, our semantic data augmentation approach achieves state-of-the-art performance on the CUB-200-2011 dataset. The source code will be released.
\end{abstract}


\section{Introduction}\label{sec:introduction}

\IEEEPARstart{F}{ine-grained} image recognition aims to distinguish objects with subtle differences in visual appearance within the same general category, e.g., different species of animals \cite{wah2011caltech, van2015building, KhoslaYaoJayadevaprakashFeiFei_FGVC2011}, different models of aircraft \cite{krause20133d}, different kinds of retail products \cite{wei2022rpc, bai2020products}. The key challenge therefore lies in comprehending fine-grained visual differences that sufficiently discriminate between objects that are highly similar in overall appearance but differ in subtle traits \cite{zhao2017survey, zheng2018survey, wei2021fine}. In recent years, deep learning \cite{lecun2015deep} has emerged as a powerful tool to learn discriminative image representations and has achieved great success in the field of fine-grained visual recognition \cite{wei2021fine, zhao2017survey}.


\begin{figure}[t]
    \vskip -0.12 in
    \begin{center}
    \centerline{\includegraphics[width=\columnwidth]{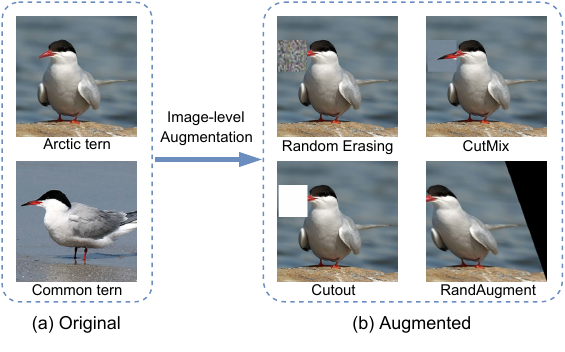}}  
    \vspace{-3ex}
    \caption{An example of the discriminative region loss problem. (a) The original images belong to the arctic tern and the common tern categories, respectively. The primary visual difference between them is the color of the \emph{bill} (common terns have a dark tip in the bill, while arctic terns do not). (b) The image samples augmented by some popular image-level data augmentation techniques. The random editing region behaviors inherited in these techniques can potentially hurt the discriminative region (e.g. tern's bill) for fine-grained images. Furthermore, the random crop and paste behavior of CutMix may even replace the critical region with that from another class. As a result, image-level data augmentation approaches might induce the noisy labels, which downgrades the model performance in the fine-grained scenario.}
    \label{fig:motivation}
    \end{center}
    \vskip -0.30 in
\end{figure}

\begin{figure*}[ht]
  \vskip -0.05 in
  \begin{center}
     \includegraphics[width=0.95\linewidth]{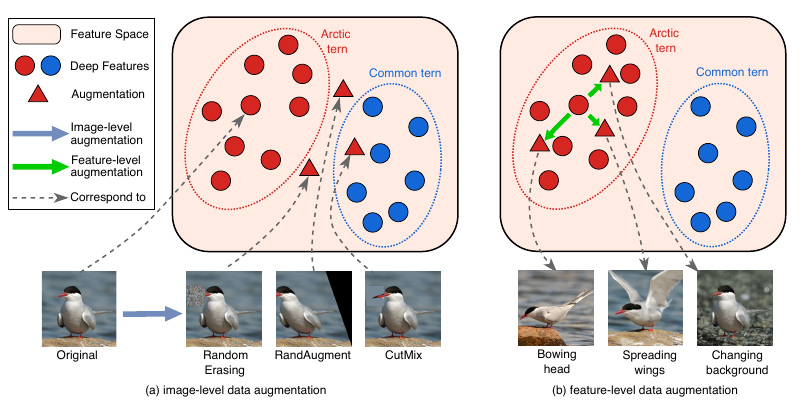}
  \end{center}
  \vskip -0.2in
  \caption{Illustration of the difference between image-level data augmentation and feature-level data augmentation. The random editing behavior of image-level data augmentation is prone to destroy the discriminative subtle regions of fine-grained images. Feature-level data augmentation alleviate this problem by directly translating deep features into meaningful directions in the feature space.}
  \label{fig:feature_space}
  \vskip -0.15 in
\end{figure*}

As deep neural networks dominate the field of visual object recognition \cite{huang2022glance, wang2021adaptive, wang2021not, wang2020glance, han2022latency, han2022learning, pu2023adaptive, han2023dynamic}, data augmentation techniques further boost the generalization ability of neural networks in the generic image classification scenario. Popular data augmentation techniques, e.g., Mixup \cite{zhang2018mixup}, CutMix \cite{yun2019cutmix}, RandAugment\cite{cubuk2020randaugment}, and Random Erasing \cite{zhong2020random}, have become a standard recipe in training modern convolution networks \cite{liu2022convnet, tan2021efficientnetv2, wang2022efficienttrain} and vision transformers \cite{liu2021swin, touvron2021training, wang2021not}. However, in the scenario of fine-grained image recognition, these data augmentation techniques are rarely applied because of the \emph{discriminative region loss} problem. Specifically, as illustrated in \figurename~\ref{fig:motivation}, the random editing behavior of \emph{image-level} data augmentation approaches have a risk of destroying discriminative regions, which is of great significance in performing fine-grained recognition. For example, random drop based data augmentations, such as Cutout \cite{devries2017improved}, Random Erasing \cite{zhong2020random},  have the potential to drop the discriminative regions of fine-grained objects. Geometric transformation, which is contained in AutoAugment \cite{cubuk2019autoaugment} and RandAugment \cite{cubuk2020randaugment}, is also likely to cause a loss to discriminative visual cues. Besides, mix-based techniques (e.g. Mixup \cite{zhang2018mixup}, CutMix \cite{yun2019cutmix}) would even result in a \emph{noisy label} problem by replacing discriminative regions of the current image with those of another class. \figurename~\ref{fig:feature_space} (a) further shows such limitations of the \emph{image-level} augmentation techniques in the feature space: the deep feature of an image-level \emph{augmented} image would probably distribute on the classification boundary or even intrude into the feature space of another category.

To cope with the aforementioned problem, we propose to augment the training samples at the \emph{feature} level (\figurename~\ref{fig:feature_space} (b)) rather than the image level. By translating data samples in the deep feature space along their corresponding meaningful semantic directions, the feature-level data augmentation would produce diversified augmented image features, which correspond to semantic meaningful images in the pixel space. In this way, the implicit data augmentation method alleviates the discriminative region loss problem caused by the random editing manner of image-level data augmentation techniques.

\begin{figure}[h]
    \begin{center}
    \centerline{\includegraphics[width=0.9\columnwidth]{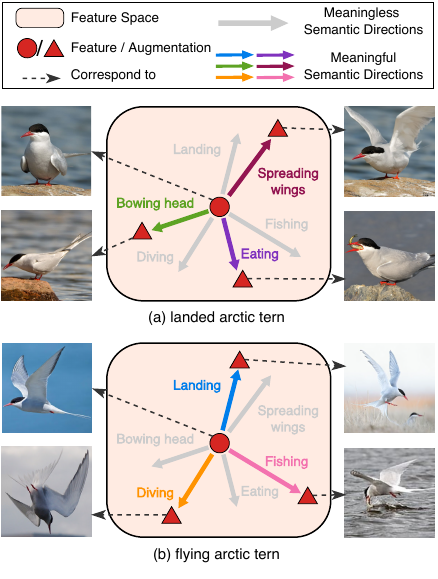}}  
    \vspace{-3ex}
    \caption{An example of semantic directions for different image samples within the same subordinate category (arctic tern). (a) A landed arctic tern has the semantic transform of bowing its head, spreading wings, and eating, while the flying ones do not. This is because a flying bird cannot eat, bow its head or spread its wings after it has spread its wings. (b) A flying arctic tern has the semantic transform of landing, diving, and fishing, while the birds on the land cannot fish in the lake, dive down in the air, or land again.}
    \label{fig:samplewise_good}
    \end{center}
    \vskip -0.25in
\end{figure}

The performance of implicit data augmentation techniques heavily relies on the quality of the semantic directions. The existing feature-level data augmentation method (implicit semantic data augmentation, ISDA~\cite{wang2019implicit}) verifies that, in the generic image classification problem, a global set of semantic directions shared by all classes is inferior to maintaining a number of sets of semantic directions for each category. Therefore, ISDA \cite{wang2019implicit} takes the \emph{class-conditional} covariance matrices of deep features as the candidate semantic directions of each category and estimates the covariance matrices statistically in an \emph{online} manner. However, in the fine-grained scenario, the limitations of the augmentation approach in \cite{wang2019implicit} are two-fold: 1) the \emph{online estimation} strategy is sub-optimal due to the limited amount of training data in each class; 2) the \emph{class-conditional} semantic directions are not suitable in the fine-grained recognition for its large intra-class variation and small inter-class variation \cite{wei2021fine, wei2019deep}. Specifically, the meaningful semantic directions of the image samples within the same sub-category vary because of the large intra-class variance. For example, in \figurename~\ref{fig:samplewise_good} the landed birds have some meaningful semantic directions that flying birds do not have, and vice versa.
As a result, augmenting all samples within one sub-category along the same set of semantic directions is improper. Intuitively, diversifying different training samples along their corresponding semantic directions is preferable.

In this paper, we propose a learnable semantic data augmentation method for the fine-grained image recognition problem.
The meaningful semantic directions is automatically \emph{learned} in a sample-wise manner based on a \emph{covariance matrix prediction network} (CovNet) rather than estimated class-conditionally with statistical method~\cite{wang2019implicit}. The CovNet takes in the deep features of each training sample and predicts their meaningful semantic directions. The covariance matrix prediction network and the classification network are jointly trained in a meta-learning manner. The meta-learning framework optimizes the two networks in an alternate way with different objectives, which solves the degeneration problem when optimizing them jointly (see the theoretical and empirical analysis in Sec.~\ref{ssec:CovNet} and Sec.~\ref{sec_ablation}, respectively). Compared with the online estimation approach \cite{wang2019implicit}, our proposed sample-wise prediction method can effectively produce appropriate semantic directions for each training sample, and therefore boost the network performance in the fine-grained image classification task.

We evaluate our method on four popular fine-grained image recognition benchmarks (i.e. CUB-200-2011 \cite{wah2011caltech}, Stanford Cars \cite{krause20133d}, FGVC Aircrarts \cite{maji2013fine} and NABirds \cite{van2015building}). Experimental results show that our approach effectively enhances the intra-class compactness of learned features and significantly improves the performance of mainstream classification networks (e.g. ResNet \cite{he2016deep}, DenseNet \cite{huang2017densely}, EfficientNet \cite{tan2019efficientnet}, RegNet \cite{radosavovic2020designing} and ViT \cite{dosovitskiy2020image}) on various fine-grained classification datasets. Combined with a recent proposed fine-grained recogntion method (P2P-Net \cite{yang2022fine}), the proposed method achieves state-of-the-art performance on CUB-200-2011.

\section{Related work}\label{sec:related}

\textbf{Fine-grained recognition}.
Fine-grained image recognition aims to discriminate numerous visually similar subordinate categories that belong to the same basic category \cite{wei2019deep, wei2021fine, zhao2017survey, wang2019survey, zhang2014part, yang2018learning, ding2021ap, chang2020devil, koniusz2021power, xu2016webly, deng2015leveraging, he2022transfg, yu2023difficulty, li2018domain}. 
Recognizing fine-grained categories is difficult due to the challenges of discriminative region localization and fine-grained feature learning. Broadly, existing fine-grained recognition approaches can be divided into two main paradigms: localization methods and feature encoding methods. The former usually creates models that capture the discriminative semantic parts of fine-grained objects and then construct a mid-level representation corresponding to these parts for the final classification. Common methods could be divided as employ detection or segmentation techniques \cite{zhang2014part, ge2019weakly, wang2020graph, liu2020filtration}, utilize deep filters \cite{wang2018learning, ding2019selective, huang2020interpretable} and leverage attention mechanisms \cite{zhang2019learning, zheng2019looking, zheng2019learning, ji2020attention, he2022transfg}. The feature encoding methods aim to learn a unified, yet discriminative, image representation for modeling subtle differences between fine-grained categories. Common practices include performing high-order feature interactions \cite{yu2018hierarchical, wei2018grassmann, zheng2019learningdeep, min2020multi, sun2020text} and designing novel loss functions \cite{dubey2018maximum, sun2020fine, zhuang2020learning, chang2020devil, xu2023multi}. However, because image-level data augmentation techniques tend to be harmful to discriminative subtle regions of fine-grained images, the designing of data augmentation strategies for fine-grained vision tasks is rarely explored.

\textbf{Data Augmentation}.
In recent years, data augmentation techniques have been widely used in training deep neural networks \cite{shorten2019survey, perez2017effectiveness, cubuk2019autoaugment, zhong2020random, devries2017improved, taylor2018improving, inoue2018data, summers2019improved, takahashi2019data, mikolajczyk2018data, devries2017dataset, wong2016understanding, bowles2018gan, lim2018doping, zhang2019dada, verma2019manifold, zhang2023style}. Basic data augmentation, such as rotation, translation, cropping, flipping \cite{simonyan2015very}, are commonly used to increase the diversity of training samples. Beyond these, Cutout \cite{devries2017improved}, Mixup \cite{zhang2018mixup} and CutMix \cite{yun2019cutmix} are manually design with domain knowledge. Recently, inspired by the neural architecture search algorithms, some works attempt to automate learning data augmentation policies, such as AutoAugment \cite{cubuk2019autoaugment} and RandAugment \cite{cubuk2020randaugment}. Although these data augmentation methods are commonly used and some of them have even become a routine in training deep neural networks in the generic classification problem, they are rarely adopted in the fine-grained scenario because random crop, mix and deform operations in the image level would easily destroy the discriminative information of fine-grained objects. 
Inspired by a recently proposed technique \cite{wang2019implicit}, recent studies have achieved significant success by applying feature-level data augmentation to augment minority classes in the long-tailed recognition problem (MetaSAug \cite{li2021metasaug}) and enhancing classifier adaptability in domain adaptation through the generation of source features aligned with target semantics (TSA \cite{li2021transferable}). These methods demonstrate the ability to achieve state-of-the-art performance in their problems. In contrast, this paper focus on the longstanding fine-grained recognition problem and proposes an important extension of ISDA \cite{wang2019implicit}, aiming to address the discriminative region loss problem through the augmentation of training data at the feature level.

\textbf{Meta-learning}. The field of meta-learning has seen a dramatic rise in interest in recent years \cite{finn2017model, franceschi2018bilevel, liu2018darts, snell2017prototypical, metz2018meta, duan2016rl, houthooft2018evolved, alet2019meta, real2019regularized, zoph2017neural, lemke2015metalearning, hospedales2021meta, ravi2017optimization, mishra2018simple, li2021metasaug, li2021meta}. Contrary to the conventional deep learning approach which solves the optimization problem with a fixed learning algorithm, meta-learning aims to improve the learning algorithm itself. Under the meta-learning framework, a machine learning model could gain experience over multiple learning episodes, and uses this experience to improve its future learning performance. Meta-learning has proven useful both in multi-task scenarios where task-agnostic knowledge is extracted from a family of tasks and used to improve the learning of new tasks from that family \cite{thrun1998learning, finn2017model}, and in  single-task scenarios where a single problem is solved repeatedly and improved over multiple episodes \cite{franceschi2018bilevel, liu2018darts, andrychowicz2016learning}. Successful applications have been demonstrated in areas spanning few-shot image recognition \cite{finn2017model, snell2017prototypical}, unsupervised learning 
\cite{metz2018meta}, data-efficient \cite{duan2016rl, houthooft2018evolved} and self-directed \cite{alet2019meta} reinforcement learning, hyperparameter optimization \cite{franceschi2018bilevel}, domain generalizable person ReID\cite{zhang2023style}, and neural architecture search \cite{liu2018darts, zoph2017neural}. In this paper, we design a single-task meta-learning algorithm, which learns the meaningful semantic directions for each training sample during the classification model training process.

\section{Method}


\begin{table}[t]
    \vskip -0.10in
    \caption{Notations used in this paper}
    \vskip -0.10in
    \label{tab:notations}
    \begin{center}
    \begin{tabular}{c|c}
        \toprule
        \textbf{Notations} & \textbf{Descriptions} \\
        \midrule
        $\bm{x}_i, \bm{a}_i$ & The $i$-th sample and its deep feature \\
        $y_i, \hat{y}_i$ & The label and the prediction logits for the $i$-th sample \\
        $\mathbf{X}, \bm{y}$ & Mini-batch of input samples and labels \\
        \midrule
        $f(\cdot, \bm{\theta}_f)$ & Classification network parameterized by $\bm{\theta}_f$ \\
        $f^b, f^h$ & Feature extractor (\textbf{b}ackbone) and classification \textbf{h}ead of $f$ \\
        $g(\cdot, \bm{\theta}_g)$ & Covariance prediction network parameterized by $\bm{\theta}_g$ \\
        \midrule
        $\bm{\theta}_f^{(t)}$ & The parameter of $f$ at time step $t$ \\
        $\tilde{\bm{\theta}}_f^{(t)}$ & The \emph{pseudo-updated} parameter of $f$ at time step $t$ \\
        \midrule
        $\hat{\Sigma}_c$ & The \emph{estimated} covariance matrix for class $c$ in  \cite{wang2019implicit} \\
        $\Sigma_i^g$ & Our covariance matrix \emph{predicted} by $g$ for sample $i$ \\
        \midrule
        $\ell, \mathcal{L}$ & The loss for one sample, and for a set of samples \\
        \bottomrule
        \end{tabular}
    \end{center}
    \vskip -0.10in
\end{table}

In this section, we first introduce the preliminaries of our method, implicit semantic data augmentation (ISDA) \cite{wang2019implicit} and its online estimation algorithm for the covariance matrices. Then our meta-learning-based framework will be presented. We further present the convergence proof of our proposed meta-learning algorithm. For better readability, we list the notations used in this paper in Table \ref{tab:notations}.

\vspace{-3.2mm}
\subsection{Implicit Semantic Data Augmentation}

Most conventional data augmentation methods \cite{yun2019cutmix, zhang2018mixup, zhong2020random, cubuk2020randaugment} make modifications directly on training \emph{images}. In contrast, ISDA performs data augmentation at the \emph{feature} level, i.e., translating image features along meaningful semantic directions. Such directions are determined based on the covariance matrices of deep features. Specifically, for a $C$-class classification problem, ISDA in \cite{wang2019implicit} statistically estimates the class-wise covariance matrices $ \bm{\hat{\Sigma}}\! = \! \{ \hat{\Sigma}_1, \hat{\Sigma}_2, ..., \hat{\Sigma}_C \}$ in an \emph{online} manner at each training iteration. For the $i$-th sample $\bm{x}_i$ with ground truth $y_i$, ISDA randomly samples transformation directions from the Gaussian distribution $\mathcal N(0, \lambda \hat{\Sigma}_{y_i})$ to augment the deep feature $\bm{a}_i$, where $\bm{a}_i$ is the learned feature to be fed to the last fully-connected layer in a deep network, and $\lambda$ is a hyperparameter controlling the augmentation strength. One should sample a large number ($M$) of directions in $\mathcal N(0, \lambda \hat{\Sigma}_{y_i})$ to get the sufficiently augmented features $\bm{a}_{i}^{m} \!\sim\! \mathcal N(\bm{a}_{i}, \lambda \hat{\Sigma}_{y_i}), m=1, 2, \cdots, M $. The modified cross-entropy loss on these augmented data can be written as




\begin{equation}
\setlength{\abovedisplayskip}{-1ex}
\ell_{M} = \frac{1}{M} \sum_{m=1}^{M} -\log \left(\frac{e^{\boldsymbol{w}_{y_{i}}^{\mathrm{T}} \boldsymbol{a}_{i}^{m}+b_{y_{i}}}}{\sum_{j=1}^{C} e^{\boldsymbol{w}_{j}^{\mathrm{T}} \boldsymbol{a}_{i}^{m}+b_{j}}}\right),
\end{equation}

\noindent
where $[\bm{w}_{1},\dots, \bm{w}_{C}]^{T} $ and $ [b_{1},\dots, b_{C}]^T$ are the trainable parameters of the last fully connected layer, $M$ is the number of sampled directions. Take a step further, if infinite directions are sampled, ISDA derives the upper bound of the expected cross-entropy loss on all augmented features:

\begin{equation}
\setlength{\abovedisplayskip}{-5pt}
\begin{aligned}
\ell_{\infty} &= \lim_{M \to \infty} \ell_{M} 
= \mathrm{\mathbb{E}}_{\bm{a}_{i}^{m}} \left [ 
-\log \left(\frac{e^{\boldsymbol{w}_{y_{i}}^{\mathrm{T}} \boldsymbol{a}_{i}^{m}+b_{y_{i}}}}{\sum_{j=1}^{C} e^{\boldsymbol{w}_{j}^{\mathrm{T}} \boldsymbol{a}_{i}^{m}+b_{j}}}\right)
\right ]
\\
&\leq -\log \left(\frac{e^{\boldsymbol{w}_{y_{i}}^{\mathrm{T}} \boldsymbol{a}_{i}+b_{y_{i}}}}{\sum_{j=1}^{C} e^{\boldsymbol{w}_{j}^{\mathrm{T}} \boldsymbol{a}_{i}+b_{j}+\frac{\lambda}{2} \boldsymbol{v}_{j y_{i}}^{\mathrm{T}} \hat{\Sigma}_{y_{i}} \boldsymbol{v}_{j y_{i}}}}\right) \triangleq \ell^{\text{ISDA}},
\end{aligned}
\end{equation}


\noindent
where $\bm{v}_{j y_{i}}\!=\!\bm{w}_{j}-\bm{w}_{y_{i}}$ and the upper bound is defined as the ISDA loss $\ell^{\text{ISDA}}$. By optimizing the upper bound $\ell^{\text{ISDA}}$, the feature-level semantic augmentation procedure is implemented efficiently. Equivalently, ISDA can be seen as a novel robust loss function, which is compatible with any neural network architecture training with the cross-entropy loss.

\vspace{-3.2mm}
\subsection{Covariance Matrix Prediction Network}\label{ssec:CovNet}

The performance of the aforementioned ISDA \cite{wang2019implicit} heavily relies on the \emph{estimated} class-wise covariance matrices, which directly affect the quality of semantic directions. In the fine-grained visual recognition scenario, the limited amount of training data, along with its large intra-class variance and small inter-class variance characteristic, poses great challenges on the covariance estimation. The unsatisfying covariance matrices can further cause limited improvement in model performance. To this end, we propose to automatically \emph{learn} \emph{sample-wise} semantic directions based on a {covariance matrix prediction network} instead of statistically \emph{estimating} the \emph{class-wise} covariance matrices as in the existing method \cite{wang2019implicit}.

Our covariance matrix prediction network (CovNet) is established as a multilayer perception (MLP), denoted as $g(\cdot; \bm{\theta}_g)$ parameterized by $\bm{\theta}_g$. The CovNet take the deep feature $\boldsymbol{a}_i$ as input, and predicts its \emph{sample-wise} semantic directions: $\Sigma_i^g = g(\boldsymbol{a}_i;\bm{\theta}_g).$ As a result, the {ISDA} loss function with our {predicted} covariance matrices can be rewritten as

\begin{equation}\label{joint_loss}
\setlength{\abovedisplayskip}{-3pt}
\begin{aligned}
&\ell^{\text{ISDA}} (\boldsymbol{x}_i, y_i; \Sigma_i^g(\bm{\theta}_g), \bm{\theta}_{f})\\
=& -\log \left(\frac{e^{\boldsymbol{w}_{y_{i}}^{\mathrm{T}} \boldsymbol{a}_{i}+b_{y_{i}}}}{\sum_{j=1}^{C} e^{\boldsymbol{w}_{j}^{\mathrm{T}} \boldsymbol{a}_{i}+b_{j}+\frac{\lambda}{2} \boldsymbol{v}_{j y_{i}}^{\mathrm{T}} \Sigma_i^g(\bm{\theta}_g) \boldsymbol{v}_{j y_{i}}}}\right),
\end{aligned}
\end{equation}

\noindent
where $\bm{a}_{i}\!=\!f^b(\bm{x}_{i}; \bm{\theta}_{f^b})$ is the deep feature extracted by $f^b$, and $\bm{\theta}_{f^{h}} = \{[\boldsymbol{w}_{1},\dots, \boldsymbol{w}_{C}]^{T}, [b_{1},\dots, b_{C}]^T\}$ refers to the parameters of the classification head (fully-connected layer). In practice, due to GPU memory limitation, the CovNet only predicts the diagonal elements of the covariance matrices and we set all other elements as zero following~\cite{wang2019implicit}. We use the Sigmoid activation in the last layer of $g$ to ensure the produced covariance matrices are positive definite.






It is worth noting that if we optimize the covariance matrix prediction network $g$ and the classification network $f$ simultaneously with the loss function in Eq.~(\ref{joint_loss}), a trivial solution will be derived for our CovNet $g$. Since the covariance matrix $\Sigma_i$ is positive definite, adding the term $\frac{\lambda}{2} \boldsymbol{v}_{j y_{i}}^{\mathrm{T}} \Sigma_i^g \boldsymbol{v}_{j y_{i}}$ to the denominator would always increase the loss value:

\begin{equation}
\label{eq:degenerate}
\begin{aligned}
\ell^{\text{ISDA}}
& \geq
-\log \left(\frac{e^{\boldsymbol{w}_{y_{i}}^{\mathrm{T}} \boldsymbol{a}_{i}+b_{y_{i}}}}{\sum_{j=1}^{C} e^{\boldsymbol{w}_{j}^{\mathrm{T}} \boldsymbol{a}_{i}+b_{j}}}\right).
\end{aligned}
\end{equation}

\noindent
Therefore, the naive joint training strategy would trivially encourage the CovNet to produce a zero-valued matrix $\Sigma_i^g$, and limited improvement for the classification accuracy will be obtained (see the results in Section~\ref{sec_ablation}).

\vspace{-3.2mm}
\subsection{The Meta-learning Method}

We propose a meta-learning-based~\cite{finn2017model, shu2019meta} approach to deal with the degenerate solution problem illustrated in Eq.~\eqref{eq:degenerate}. By setting a meta-learning objective and optimizing $g$ with meta-gradient~\cite{finn2017model}, the CovNet $g$ could learn to produce appropriate covariance matrices by mining the meta knowledge from metadata. In this subsection, we first formulate the forward pass and the training objectives of the classification network $f$ and the CovNet $g$, respectively. Then the detailed optimization pipeline of the two networks is presented.

\subsubsection{The meta-learning objective}

\begin{figure*}[ht]
  \begin{center}
     \includegraphics[width=0.75\linewidth]{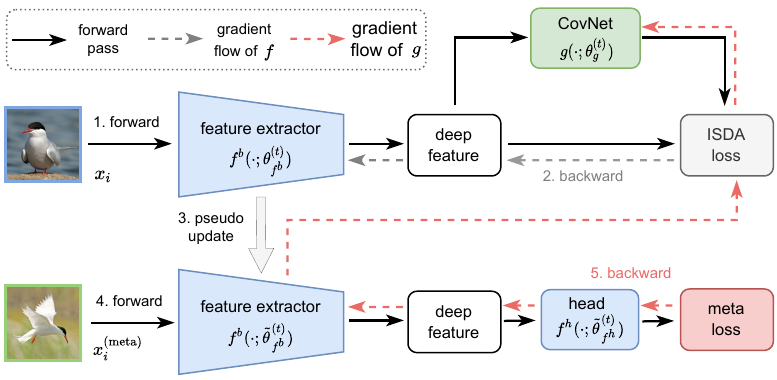}
  \end{center}
  \vskip -0.2in
  \caption{The forward data flow and backward gradient flow for updating the classification model $f$ and the CovNet $g$. We first compute the ISDA loss (Eq.~\eqref{eq:train_loss}) and take a backward step to get the pseudo updated classification network $\tilde{\bm{\theta}}_f$ (step 1 $\sim$ 3). Then, the meta loss is computed to update $\bm{\theta}_g$ (Eq.~\eqref{eq:meta_loss}, step 4 and 5). Note that the parameters of the classification head $\bm{\theta}_{f^h}$ is implicitly contained in the ISDA loss (Eq.~(\ref{joint_loss})) in step 1.} 
  \label{fig:grad_flow}
  \vskip -0.2in
\end{figure*}

With our learned covariance matrix scheme, the optimization objective of the classification network parameter $\bm{\theta}_f$ is minimizing the ISDA loss under the learned covariance matrices. The classification network $\bm{\theta}_f$ is optimized with the training data. Given a training sample $\bm{x}_i$, the feature extractor $f^b$ (backbone of the classification network $f$) produces its feature $\bm{a}_i = f^b(\bm{x}_i; \bm{\theta}_{f^b})$. The corresponding covariance matrix  $\Sigma_i^g = g(\bm{a}_i; \bm{\theta}_{g})$ is predicted by feeding the deep feature $\bm{a}_i$ into the CovNet, and serves as a parameter of ISDA loss function. The loss of training sample $\bm{x}_i$ can be formulated as $\ell^{\text{ISDA}}(\bm{x}_i, y_i; \Sigma_i^g(\bm{\theta}_g), \bm{\theta}_{f})$. Therefore, the training loss over the whole training dataset is

\begin{equation}
\setlength{\abovedisplayskip}{-3pt}
\label{eq:train_loss}
\mathcal{L}^{\text{train}}(\bm{\theta}_{f} ; \bm{\theta}_{g}) = \frac{1}{N_{1}} \sum_{i=1}^{N_{1}} \ell^{\text{ISDA}}(\bm{x}_i, y_i; \Sigma_i^g(\bm{\theta}_g), \bm{\theta}_{f}),
\end{equation}

\noindent
where $N_1$ is the number of training data. The optimization target of the classification network  $\bm{\theta}_f$ is minimizing the training loss under the learned covariance matrices $\Sigma_i^g$

\begin{equation}
\label{eq:cls_obj}
\setlength{\abovedisplayskip}{-3pt}
\begin{aligned}
\bm{\theta}_{f}^{*}(\bm{\theta}_{g}) &= \underset{\bm{\theta}_{f}}{\arg \min } \; \mathcal{L}^{\text{train}}(\bm{\theta}_{f} ; \bm{\theta}_{g}). \\
\end{aligned}
\end{equation}

\noindent
It can be observed from Eq.~\eqref{eq:cls_obj} that the optimized $\bm{\theta}_f^*$ is a function of the parameter of the CovNet $\bm{\theta}_g$.

The network parameter of the CovNet $\bm{\theta}_g$ is learned in a meta-learning approach. We use the meta data from the meta dataset $\{\bm{x}_{i}^{\text{(meta)}}, y_{i}^{\text{(meta)}}\}_{i=1}^{N_2}$ to optimize the CovNet. The meta loss is the cross-entropy loss between the network prediction $\hat{y}_{i}^{\text{(meta)}}$ and corresponding ground truth $y_{i}^{\text{(meta)}}$

\begin{equation}
\setlength{\abovedisplayskip}{-3pt}
\setlength{\belowdisplayskip}{-3pt}
\label{eq:meta_loss}
\begin{aligned}
\mathcal{L}^{\text{meta }}(\bm{\theta}_{f}(\bm{\theta}_{g})) 
= \frac{1}{N_2} \sum_{i=1}^{N_2} \ell^{\text{CE}}\left(\hat{y}_{i}^{\text{(meta)}},y_{i}^{\text{(meta)}}\right), \\
\end{aligned}
\end{equation}

\noindent
where $N_2$ denotes the number of samples of the meta set, $\hat{y}_{i}^{\text{(meta)}} = f(\bm{x}_{i}^{\text{(meta)}}; \bm{\theta}_{f}(\bm{\theta}_{g}))$ is the network output of the meta sample $\bm{x}_{i}^{\text{(meta)}}$ and $\ell^{\text{CE}}(\cdot, \cdot)$ means the cross-entropy loss. Therefore, the optimal parameter of the CovNet $\bm{\theta}_g$ is obtained by minimizing the meta-learning objective $\mathcal{L}^{\text{meta }}$

\begin{equation}
\setlength{\abovedisplayskip}{-3pt}
\setlength{\belowdisplayskip}{-3pt}
\label{eq:meta_obj}
\begin{aligned}
\bm{\theta}_{g}^{*}
&=\underset{\bm{\theta}_{g}}{\arg \min } \; \mathcal{L}^{\text {meta }}(\bm{\theta}_{f}(\bm{\theta}_{g})). \\
\end{aligned}
\end{equation}

Since the CovNet $g$ is optimized by backpropagating the meta-gradient~\cite{finn2017model} through the gradient operator in the meta-objective and the meta-objective is established with cross-entropy loss rather than ISDA loss, the degeneration problem (illustrated in Eq.~\eqref{eq:degenerate}) when training $\bm{\theta}_{f}$ an $\bm{\theta}_{g}$ in one gradient step with the same optimization objective will not happen.

\subsubsection{The optimization pipeline}
\label{subsec:the_optimization_pipeline}

\begin{figure}[t]
  \begin{center}
     \includegraphics[width=0.6\linewidth]{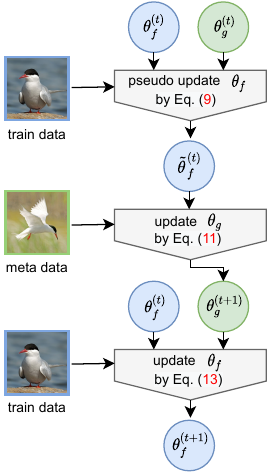}
  \end{center}
  \vskip -0.1in
  \caption{The optimization pipeline in iteration t. It contains three steps: (1) the classification network $\bm{\theta}_{f}$ is pseudo-updated by the training samples; (2) the covariance prediction network $\bm{\theta}_{g}$ is updated by the gradient of the pseudo-updated $\hat{\bm{\theta}}_{f}$; (3) $\bm{\theta}_{f}$ take a real update step based on the predicted covariance matrix by the new CovNet $\bm{\theta}_{g}^{(t+1)}$.}
  \label{fig:pipeline}
  \vskip -0.2in
\end{figure}

Under our meta-learning framework, the classification network $\bm{\theta}_f$ and the covariance prediction network $\bm{\theta}_g$ are jointly optimized in a nested way. Specifically, the CovNet $\bm{\theta}_g$ is updated by the meta data on the basis of the pseudo updated classification network $\tilde{\bm{\theta}}_f$. The pseudo updated classification networks $\tilde{\bm{\theta}}_f$ is constructed not only for the gradient provided by the training data, but also for finding the appropriate  $\bm{\theta}_g$ under the current batch of training data. After the $\bm{\theta}_g$ is updated, the classification network conduct a real update using the covariance matrices predicted by the updated $\bm{\theta}_g$. For clarity, we sum the meta-learning algorithm into three phases (illustrated in \figurename~\ref{fig:pipeline}): the pseudo update process of the classification network, the meta update process of the CovNet and the real update process of the classification network. The detailed optimization procedure within a learning iteration is presented as follows.

\textit{\textbf{The pseudo update process}} updates the classification network with the training data. The pseudo updated classification network $\tilde{\bm{\theta}}_f$ only serves as a transient state in each optimization iteration. It helps the CovNet in the meta update phase find the proper parameters under the current batch of training data and provide the gradients to constructs a computational graph for the meta update process to compute meta-gradients.

Starting from a initial state $\bm{\theta}_f^{(t)}$ of the $t$-th iteration, the classification network take a pseudo update step under the current CovNet parameter $\bm{\theta}_g^{(t)}$ using training data

\begin{equation}
\setlength{\abovedisplayskip}{-3pt}
\setlength{\belowdisplayskip}{-3pt}
\bm{\theta}_{f}^{*}(\bm{\theta}_{g}^{(t)})=\underset{\bm{\theta}_{f}}{\arg \min } \; \mathcal{L}^{\text{train}}(\bm{\theta}_{f} ; \bm{\theta}_{g}^{(t)}).
\end{equation}

\noindent
The updated parameter by gradient descent is 

\begin{equation}
\setlength{\abovedisplayskip}{-2pt}
\setlength{\belowdisplayskip}{2pt}
\label{eq:pseudo_update}
\tilde{\bm{\theta}}_{f}^{(t)}(\bm{\theta}_{g}^{(t)})=\bm{\theta}_{f}^{(t)}-\left.\alpha_t \nabla_{\bm{\theta}_{f}} \mathcal{L}^{\text{train}}(\bm{\theta}_{f}; \bm{\theta}_{g}^{(t)})\right|_{\bm{\theta}_{f}^{(t)}},
\end{equation}

\noindent
where $\alpha_t$ is the learning rate of the classification network at the current time step $t$. Note that the pseudo update process only update the classification network. The gradients would not pass backward to the CovNet. See \figurename~\ref{fig:grad_flow} (grey dashed lines) for the gradient flow of this pseudo update.


\textit{\textbf{The meta update process}} updates the CovNet using the meta data. By minimizing the meta-objective over the metadata

\begin{equation}
\bm{\theta}_{g}^{*}=\underset{\bm{\theta}_{g}}{\arg \min } \; \mathcal{L}^{\text {meta }}(\tilde{\bm{\theta}}_{f}^{(t)}(\bm{\theta}_{g})),
\end{equation}

\noindent
we could get the updated CovNet parameter

\begin{equation}
\label{eq:meta_update}
\bm{\theta}_{g}^{(t+1)}=\bm{\theta}_{g}^{(t)}-\left.\beta_t \nabla_{\bm{\theta}_{g}} \mathcal{L}^{\text{meta}}\left(\tilde{\bm{\theta}}_{f}^{(t)}(\bm{\theta}_{g})\right)\right|_{\bm{\theta}_{g}^{(t)}},
\end{equation}

\noindent
where $\beta_t$ is the current learning rate of $\bm{\theta}_{g}$. The meta knowledge contained in the metadata helps the CovNet $g$ learn the appropriate semantic directions for each samples in the deep feature space. The red dashed lines in \figurename~\ref{fig:grad_flow} illustrate the gradient flow for updating the parameter of CovNet $\bm{\theta}_{g}$.

\textit{\textbf{The real update process}} updates the classification network based on the prediction of the updated CovNet $\bm{\theta}_g^{(t+1)}$. The updated CovNet helps to predict the semantic directions of each training sample and facilitate the optimization procedure of adopting feature-level data augmentation

\begin{equation}
\bm{\theta}_{f}^{*}(\bm{\theta}_{g}^{(t+1)})=\underset{\bm{\theta}_{f}}{\arg \min } \; \mathcal{L}^{\operatorname{train}}(\bm{\theta}_{f} ; \bm{\theta}_{g}^{(t+1)}).
\end{equation}

\noindent
Finally, we could get the updated classification network parameter  $\bm{\theta}_{f}^{(t+1)}$ by taking a optimization step

\begin{equation}
\label{eq:real_update}
\bm{\theta}_{f}^{(t+1)}(\bm{\theta}_{g}^{(t+1)})=\bm{\theta}_{f}^{(t)}-\left.\alpha_t \nabla_{\bm{\theta}_{f}} \mathcal{L}^{\operatorname{train}}(\bm{\theta}_{f}; \bm{\theta}_{g}^{(t+1)})\right|_{\bm{\theta}_{f}^{(t)}}.
\end{equation}

In short, in each optimization iteration, the CovNet is updated via the pseudo gradients provided by $\tilde{\bm{\theta}}_{f}$. Then, the classification network get a real optimization step using the updated CovNet. These two networks are optimized alternately under our proposed meta-learning approach.

In contrast to existing meta-learning methods that require a standalone meta dataset, following the recent meta-learning technique~\cite{han2022learning}, we simply reuse the training dataset as our meta dataset. In our batch-based optimization algorithm, the training batch and the meta batch are ensured to be totally different. To be specific, in each training iteration we sample a batch of data $\mathbb{X}$ from the training set and chunk it into the training batch $\bm{X}$ and the meta batch $\bm{X}^{\text{(meta)}}$. After a optimization iteration, we exchange $\bm{X}$ and $\bm{X}^{\text{(meta)}}$ to reuse them. We summarize the learning algorithm in Algorithm~\ref{algo} and illustrate the training pipeline in \figurename~\ref{fig:pipeline}.

\begin{algorithm}[t]
    \caption{The meta-learning Algorithm}\label{algo}
    \begin{algorithmic}
    \Require Training data $\mathcal{D}$, batch size $B$, max iteration $T$, augmentation strength $\lambda$.
    \Ensure Backbone network parameters $\bm{\theta}_f$.
    \For{$t=0$ \textbf{to} $T-1$}
        \State $\{\mathbb{X},\mathbb{Y}\} \gets \mathrm{SampleMiniBatch}(\mathcal{D},B)$.
        \State Split $\{\mathbb{X},\mathbb{Y}\}$ into $\{\mathbf{X},\mathbf{X}^{\mathrm{(meta)}},\mathbf{y},\mathbf{y}^{\mathrm{(meta)}}\}$.
        
        \State Pseudo update for $\mathbf{\tilde{\theta}}_f^{(t)}$ by Eq.~(\ref{eq:pseudo_update}) using $\mathbf{X}$.
        \State Update $\mathbf{\theta}_g^{(t+1)}$ by Eq.~(\ref{eq:meta_update}) using $\mathbf{X}^{\mathrm{(meta)}}$.
        \State Update $\mathbf{\theta}_f^{(t+1)}$ by Eq.~(\ref{eq:real_update}) using $\mathbf{X}$.
        \State Exchange $\mathbf{X}$ and $\mathbf{X}^{\mathrm{(meta)}}$ and repeat the three steps above.
    \EndFor
    \end{algorithmic}
\end{algorithm}

\vspace{-3.2mm}
\subsection{Convergence of The Meta-learning Algorithm}

The proposed meta-learning-based algorithm involves a bi-level optimization procedure. We proved that our method converges to some critical points for both the training loss and the meta loss under some mild conditions. Specifically, we proved that the expectation of the meta loss gradient would be smaller than an infinitely small quantity in finite step, and the expectation of the gradient of the training loss will converge to zero. The theorems is demonstrated as the following. The complete proof is present in the Appendix~\ref{sec:appendix}.

\textbf{Theorem 1.} Suppose the ISDA loss function  $\ell^{\text{ISDA}}$ and the cross-entropy loss function $\ell^{\text{CE}}$ are both differentiable, Lipschitz continuous with constant $L$ and have $\rho$-bounded gradients with respect to the training/meta data. The learning rate satisfies $a_t = \min\{1, \frac{k}{T}\}$, for some $k>0$, such that $\frac{k}{T} < 1$. The meta learning rate $\beta_t (1<t<N)$ is monotone descent sequence. $\beta_t = \min \{\frac{1}{L}, \frac{c}{\sigma\sqrt{T}}\}$ for some $c>0$, such that $\frac{\sigma\sqrt{T}}{c} \geq L$ and $\sum_{t=1}^{\infty} \beta_t \leq \infty$, $\sum_{t=1}^{\infty} \beta_t^2 \leq \infty$. Then

\noindent 1) the proposed algorithm can always achieve
\begin{align}
\begin{split}
\min _{0 \leq t \leq T} \mathbb{E}\left[\left\|\nabla \mathcal{L}^{\text{meta}}\left(\bm{\theta}_{g}^{(t)}\right)\right\|_{2}^{2}\right] \leq \mathcal{O}\left(\frac{1}{\sqrt{T}}\right)
\end{split}
\end{align}
\noindent
in $T$ steps.

\noindent
2) the training loss is convergent
\begin{align}
\begin{split}
\lim_{t \to \infty} \mathop{\mathbb{E}} \left [ \| \nabla \mathcal{L}^{\text{train}}(\bm{\theta}_{f}^{(t)};\bm{\theta}_{g}^{(t)})\|_2^2 \right ] = 0.
\end{split}
\end{align}

\vspace{-3.2mm}
\subsection{Accelerating the meta-learning framework}

The cost of the meta update process is relatively high because updating $\bm{\theta}_g$ requires computing second-order gradients (see Eq.~\eqref{eq:meta_update}). In order to make the training procedure more efficient, we adopt an approximate method to accelerate the meta update process by freezing part of the classification network. In the pseudo update process, we freeze the first several blocks and only the late blocks will have gradients. Consequently, during the meta update process, the meta gradient will be computed solely from the gradient of a subset of the pseudo network. In this way, the algorithm significantly improves the training efficiency without sacrificing model performance. The detailed analysis is presented in Sec.~\ref{sec:froze}.

\section{Experiments}

In section, we empirically evaluate our semantic data augmentation method on different fine-grained visual recognition datasets. We first introduce the detailed experiment setup in Section~\ref{sec_settings}, including datasets and training configurations. Then the main results of our method with various backbone architectures on different datasets are presented in Section~\ref{sec_main_results}. We also conduct comparison experiments with competing approaches in Section~\ref{sec_compare_with_sota}. Finally, the ablation study (Section~\ref{sec_ablation}), the experiment results on the general recognition task (Section~\ref{tab:imagenet}), and the visualization results (Section~\ref{sec_visualization}) further validate the effectiveness of the proposed method.

\begin{table}[t]
\centering
\caption{Experiment results of our method on four popular fine-grained datasets (CUB-200-2011, FGVC Aircraft, Stanford Cars and NABirds). Basic data augmentation in the table refers to random horizontal flipping.}
\label{tab:datasets}
\begin{tabular}{c l l}
    \toprule
    Dataset & Data Augmentation & Top-1 Accuracy \\
    \midrule
    \multirow{3}{*}{CUB-200-2011}
        &\quad \quad Basic                              & $\quad \quad 84.5$ \\
        &\quad \quad ISDA~\cite{wang2019implicit}       & $\quad \quad 85.3_{(\uparrow 0.8)}$ \\
        &\quad \quad \cellcolor{lightgray!50}Our method & $\quad \quad \cellcolor{lightgray!50}86.7_{( \textcolor{blue}{\uparrow 2.2})}$ \\
    \midrule
    \multirow{3}{*}{FGVC Aircraft}
        &\quad \quad Basic & $\quad \quad 91.5$ \\
        &\quad \quad ISDA~\cite{wang2019implicit} & $\quad \quad 91.7_{(\uparrow 0.2)}$ \\
        &\quad \quad \cellcolor{lightgray!50}Our method & $\quad \quad \cellcolor{lightgray!50}92.7_{( \textcolor{blue}{\uparrow 1.2})}$ \\
    \midrule
    \multirow{3}{*}{Stanford Cars}
        &\quad \quad Basic & $\quad \quad 92.9$ \\
        &\quad \quad ISDA~\cite{wang2019implicit} & $\quad \quad 93.2_{(\uparrow 0.3)}$ \\
        &\quad \quad \cellcolor{lightgray!50}Our method & $\quad \quad \cellcolor{lightgray!50}94.3_{(\textcolor{blue}{\uparrow 1.4})}$ \\
    \midrule
    \multirow{3}{*}{NABirds}
        &\quad \quad Basic & $\quad \quad 83.5$ \\
        &\quad \quad ISDA~\cite{wang2019implicit} & $\quad \quad 83.9_{(\uparrow 0.4)}$ \\
        &\quad \quad \cellcolor{lightgray!50}Our method & $\quad \quad \cellcolor{lightgray!50}85.5_{(\textcolor{blue}{\uparrow 2.0})}$ \\
    \bottomrule
\end{tabular}
\vskip - 0.2 in
\end{table}

\vspace{-3.2mm}
\subsection{Experiment settings}\label{sec_settings}

\noindent\textbf{Datasets.} We evaluate our proposed methods on four widely used fine-grained benchmarks, i.e, CUB-200-2011 \cite{wah2011caltech}, FGVC Aircraft \cite{krause20133d}, Stanford Cars \cite{maji2013fine} and NABirds \cite{van2015building}. The CUB-200-2011 \cite{wah2011caltech} is the most wildly-used dataset for fine-grained visual categorization tasks, which contains 11,788 photographs of 200 subcategories belonging to birds, 5,994 for training and 5,794 for testing. The FGVC Aircraft \cite{krause20133d} dataset contains 10,200 images of aircraft, with 100 images for each of 102 different aircraft model variants, most of which are airplanes. The Stanford Cars \cite{maji2013fine} dataset contains 16,185 images of 196 classes of cars and is split into 8,144 training images and 8,041 testing images. The NABirds dataset \cite{van2015building} is a collection of 48,000 annotated photographs of the 400 species of birds that are commonly observed in North America. In our experiments, we only use the category label as supervision, although some datasets have additional available annotations.

For all the fine-grained datasets, we first resize the image to 600 $\times$ 600 pixels and crop it into 448 $\times$ 448 resolution (random cropping for training and center crop for testing), with a random horizontal flipping operation following behind.

\noindent\textbf{Implementation Details} For all the classification network structures, we load the pretrained network from the torchvision library, except that the ViT \cite{dosovitskiy2020image} pretrained weight is taken following TransFG \cite{he2022transfg}. We use stochastic gradient descent (SGD) optimizer to train the classification network $\bm{\theta}_{f}$, with a momentum of 0.9 and a weight decay of 0.0. The learning rate of the classification network is initialized as 0.03 with a batch size of 64, decaying with a cosine shape. All the models are trained for 100 epochs except that the combination experiment with P2P-Net \cite{yang2022fine} in Sec.~\ref{sec_compare_with_sota} follows the setting in \cite{yang2022fine}.

The covariance prediction network $\bm{\theta}_{g}$ is established as a multilayer perceptron with one hidden layer. The width of the hidden layer is set as a quarter of the final feature dimension. We also provide the ablation study on the structure of the CovNet in Sec.~\ref{sec:covnet_archi}.
Due to GPU memory limitation on high resolution images, following the practice of the original implicit data augmentation method \cite{wang2019implicit}, we approximate the covariance matrices by their diagonals, i.e., the variance of each dimension of the features. We also adopt the SGD optimizer with a learning rate of 0.001 to optimize $\bm{\theta}_{g}$. The CovNet is only used for training, and no extra computation cost will be brought during the inference procedure.

\begin{table}
\centering
\caption{Experiment results of our method when implemented with different network structures on the CUB-200-2011 dataset.}
\label{tab:nets}
\begin{tabular}{c l l}
    \toprule
    Classification Network & Data Augmentation & Top-1 Accuracy \\
    \midrule
    \multirow{3}{*}{ResNet-50}
        &\quad \quad Basic                              & $\quad \quad 84.5$ \\
        &\quad \quad ISDA~\cite{wang2019implicit}       & $\quad \quad 85.3_{(\uparrow 0.8)}$ \\
        &\quad \quad \cellcolor{lightgray!50}Our method & $\quad \quad \cellcolor{lightgray!50}86.7_{( \textcolor{blue}{\uparrow 2.2})}$ \\
    \midrule
    \multirow{3}{*}{DenseNet-161}
        &\quad \quad Basic                              & $\quad \quad 86.7$ \\
        &\quad \quad ISDA~\cite{wang2019implicit}       & $\quad \quad 87.2_{(\uparrow 0.5)}$ \\
        &\quad \quad \cellcolor{lightgray!50}Our method & $\quad \quad \cellcolor{lightgray!50}88.4_{( \textcolor{blue}{\uparrow 1.7})}$ \\
    \midrule
    \multirow{3}{*}{EfficientNet-B0}
        &\quad \quad Basic                              & $\quad \quad 85.3$ \\
        &\quad \quad ISDA~\cite{wang2019implicit}       & $\quad \quad 85.5_{(\uparrow 0.2)}$ \\
        &\quad \quad \cellcolor{lightgray!50}Our method & $\quad \quad \cellcolor{lightgray!50}86.3_{( \textcolor{blue}{\uparrow 1.0})}$ \\
    \midrule
    \multirow{3}{*}{MobileNetV2}
        &\quad \quad Basic                              & $\quad \quad 82.6$ \\
        &\quad \quad ISDA~\cite{wang2019implicit}       & $\quad \quad 83.1_{(\uparrow 0.5)}$ \\
        &\quad \quad \cellcolor{lightgray!50}Our method & $\quad \quad \cellcolor{lightgray!50}84.3_{( \textcolor{blue}{\uparrow 1.7})}$ \\
    \midrule
    \multirow{3}{*}{RegNetX-400MF}
        &\quad \quad Basic                              & $\quad \quad 82.9$ \\
        &\quad \quad ISDA~\cite{wang2019implicit}       & $\quad \quad 83.2_{(\uparrow 0.3)}$ \\
        &\quad \quad \cellcolor{lightgray!50}Our method & $\quad \quad \cellcolor{lightgray!50}84.1_{( \textcolor{blue}{\uparrow 1.2})}$ \\
    \midrule
    \multirow{3}{*}{ViT-B\_16}
        &\quad \quad Basic                              & $\quad \quad 90.3$ \\
        &\quad \quad ISDA~\cite{wang2019implicit}       & $\quad \quad 90.3_{(\uparrow 0.0)}$ \\
        &\quad \quad \cellcolor{lightgray!50}Our method & $\quad \quad \cellcolor{lightgray!50}90.7_{( \textcolor{blue}{\uparrow 0.4})}$ \\
    \bottomrule
\end{tabular}
\vskip -0.2 in
\end{table}

\vspace{-3.2mm}
\subsection{Effectiveness with different datasets and architectures}\label{sec_main_results}

We compare our proposed feature-level data augmentation method with counterparts that only use basic data augmentation (i.e. random horizontal flipping) on ResNet-50. The experiment results on the above-mentioned fine-grained datasets are shown in Table~\ref{tab:datasets}. From the results, we find that our method significantly improves the generalization ability of classification networks in various fine-grained scenarios, including birds \cite{wah2011caltech, van2015building}, aircraft \cite{krause20133d} and cars \cite{maji2013fine}. To be specific, on the most popular fine-grained dataset CUB-200-2011, the proposed method achieves 2.2\% improvement on the Top-1 Accuracy  compared to the baseline counterpart. Our method also earns more than 1.2\% Top-1 accuracy improvement on other popular fine-grained datasets. Compared with the semantic data augmentation with class-wise estimated semantic directions~\cite{wang2019implicit}, our sample-wise prediction approach shows its superiority in improving network performance.

We also apply our method on various popular classification network architectures, i.e, DenseNet \cite{huang2017densely}, EfficientNet \cite{tan2019efficientnet}, MobileNetV2 \cite{sandler2018mobilenetv2}, RegNet \cite{radosavovic2020designing}, Vision Transfromer (ViT) \cite{dosovitskiy2020image}. The results in Table~\ref{tab:nets} show that our method could improve the performance of fine-grained classification accuracy among various neural network architectures. Specifically, our method could improve the performance of networks designed for server (ResNet, DenseNet and EfficientNet) for more than 1.0\% Top-1 Accuracy. For mobile devices intended networks, our method could also get remarkable improvement (1.7\% accuracy improvement for MobileNetV2, 1.2\% for RegNetX-400MF). Our method is also effective on recent proposed vision transformer architectures (0.4\% accuracy improvement for ViT-B\_16). The results also show that, in the fine-grained scenario, sample-wise predicted semantic directions method is more effective than the class-wise estimated counterparts~\cite{wang2019implicit} among various neural network architectures.

\subsection{Comparisons with Competing Methods}\label{sec_compare_with_sota}

\begin{table}[t]
\centering
\caption{Comparison of our method with state-of-the-art (SOTA) approaches on the CUB-200-2011 dataset}
\label{tab:sota}
\begin{tabular}{l|c|c|c}
    \toprule
Method & Published in & Base Model & Accuracy \\ 
    \midrule
Bilinear CNN \cite{lin2015bilinear} & ICCV 2015 & VGG-16 & 84.1 \\
RA-CNN \cite{fu2017look} & CVPR 2017 & VGG-19 & 85.3 \\ 
MA-CNN \cite{zheng2017learning} & ICCV 2017 & VGG-19& 86.5 \\ 
Mask-CNN \cite{wei2018mask} & PR 2018 & VGG-16 & 85.7 \\
NTS-Net \cite{yang2018learning} & ECCV 2018 & ResNet-50 & 87.5 \\
MaxEnt \cite{dubey2018maximum} & NeurIPS 2018 & DenseNet-161 & 86.5 \\
DCL \cite{chen2019destruction} & CVPR 2019 & ResNet-50 & 87.8 \\
S3N \cite{ding2019selective} & ICCV 2019 & ResNet-50 & 88.5 \\
DF-GMM \cite{wang2020weakly} & CVPR 2020 & ResNet-50 & 88.8 \\
PMG \cite{du2020fine} & ECCV 2020 & ResNet-50 & 89.6 \\
FDL \cite{liu2020filtration} & AAAI 2020 & ResNet-50 & 88.6 \\
API-Net \cite{zhuang2020learning} & AAAI 2020 & ResNet-101 & 88.6 \\
MC-Loss \cite{chang2020devil} & TIP 2020 & ResNet-50 & 87.3 \\
PRIS \cite{du2021progressive} & TPAMI 2021 & ResNet-101 & 90.0 \\
CAL \cite{rao2021counterfactual} & ICCV 2021 & ResNet-101 & 90.6 \\
DTRG \cite{liu2022convolutional} & TIP 2022 & ResNet-50 & 88.8 \\
GDSMP-Net \cite{ke2023granularity} & PR 2023 & ResNet-50 & 89.9 \\
    \midrule
P2P-Net \cite{yang2022fine} & CVPR 2022 & ResNet-50 & 90.2  \\
P2P-Net + ISDA in \cite{wang2019implicit} & - & ResNet-50 & 90.4 \\
P2P-Net + Our Method & - & ResNet-50 & \cellcolor{lightgray!50}\textbf{91.0} \\
   \bottomrule
\end{tabular}
\end{table}

\noindent\textbf{Comparison and compatibility with state-of-the-arts.}
We combine our proposed sample-wise predicted semantic data augmentation method with a recent proposed fine-grained recognition technique P2P-Net \cite{yang2022fine}. 
In addition to the supervision from image label, the P2P-Net \cite{yang2022fine} further utilizes the localized discriminative parts to promote discrimination of image representations and learns a graph matching for part alignment in order to alleviate the variation of object pose. The P2P-Net \cite{yang2022fine} achieves state-of-the-art performance on the CUB-200-2011 dataset using the ResNet \cite{he2016deep} as base model.

In the P2P-Net \cite{yang2022fine}, there are totally $S+1$ classifiers, where the first $S-1$ classifiers are attached to $S-1$ intermediate feature maps of the base model, the $S^{th}$ classifier takes the final classification feature as input, and the last one make prediction depending on the combination of all the features mentioned previous. For simplicity, we only combine our method with the cross-entropy loss of the final classifier and keep other settings the same as P2P-Net \cite{yang2022fine}. We set the augmentation strength $\lambda_0$ as 5.0 and construct the CovNet as a MLP with one hidden layer, whose size is a half of the deep feature size. The meta update process in Eq.~\eqref{eq:meta_update} only occurs every ten iterations for training efficiency consideration.

Experimental results in Table.~\ref{tab:sota} show that combined with our proposed method, the P2P-Net achieves a 91.0\% Top-1 Accuracy on CUB-200-2011, which is 0.8 higher than the original P2P-Net. This result verifies the effectiveness when combining our method with other competing methods.

\begin{figure}
  \vskip -0.05in
  \begin{center}
     \includegraphics[width=1.00\linewidth]{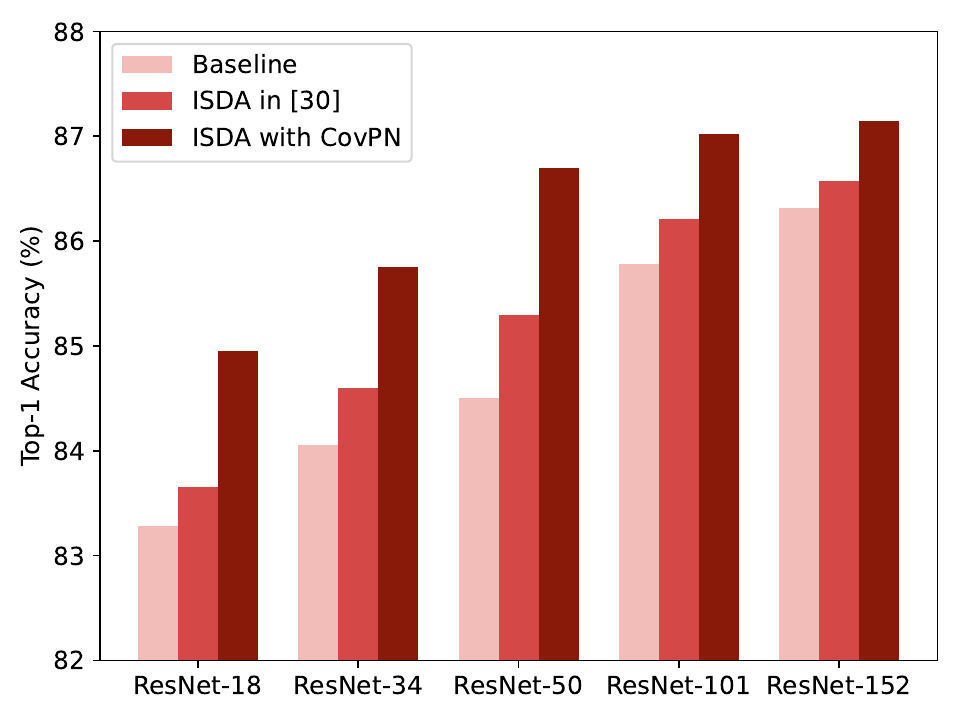}
  \end{center}
  \vskip -0.2in
  \caption{Comparison with basic data augmentation (Baseline), ISDA with online estimated covariance matrices (ISDA in \cite{wang2019implicit}) and our method (ISDA with CovNet) on CUB-200-2011. The experiments are conducted on ResNets of various depths (ResNet-\{18, 34, 50, 101, 152\}).}
  \label{fig_resnets}
  \vskip -0.2in
\end{figure}

\begin{table}[b]
\centering
\caption{Comparison of our method with image-level data augmentation techniques on the CUB-200-2011 dataset}
\label{tab:com_imglvl}
\begin{tabular}{c|l|c}
    \toprule
Classification Network & Data Augmentation & Top-1 Accuracy \\ 
    \midrule
    \multirow{6}{*}{ResNet-50}
        & {Basic } & $ 84.5 $ \\
        & {Mixup \cite{zhang2018mixup} } & $ 85.4 $ \\
        & {CutMix  \cite{yun2019cutmix} } & $ 85.9 $ \\
        & {Random Erasing \cite{zhong2020random} } & $ 83.8 $ \\
        & {RandAugment \cite{cubuk2020randaugment} } & $ 82.8$ \\
        &\cellcolor{lightgray!50}{Our method} & $ \cellcolor{lightgray!50}86.7$ \\
   \bottomrule
\end{tabular}
\end{table}

\begin{figure}
  \vskip 0.15 in
  \begin{center}
     \includegraphics[width=1.0\linewidth]{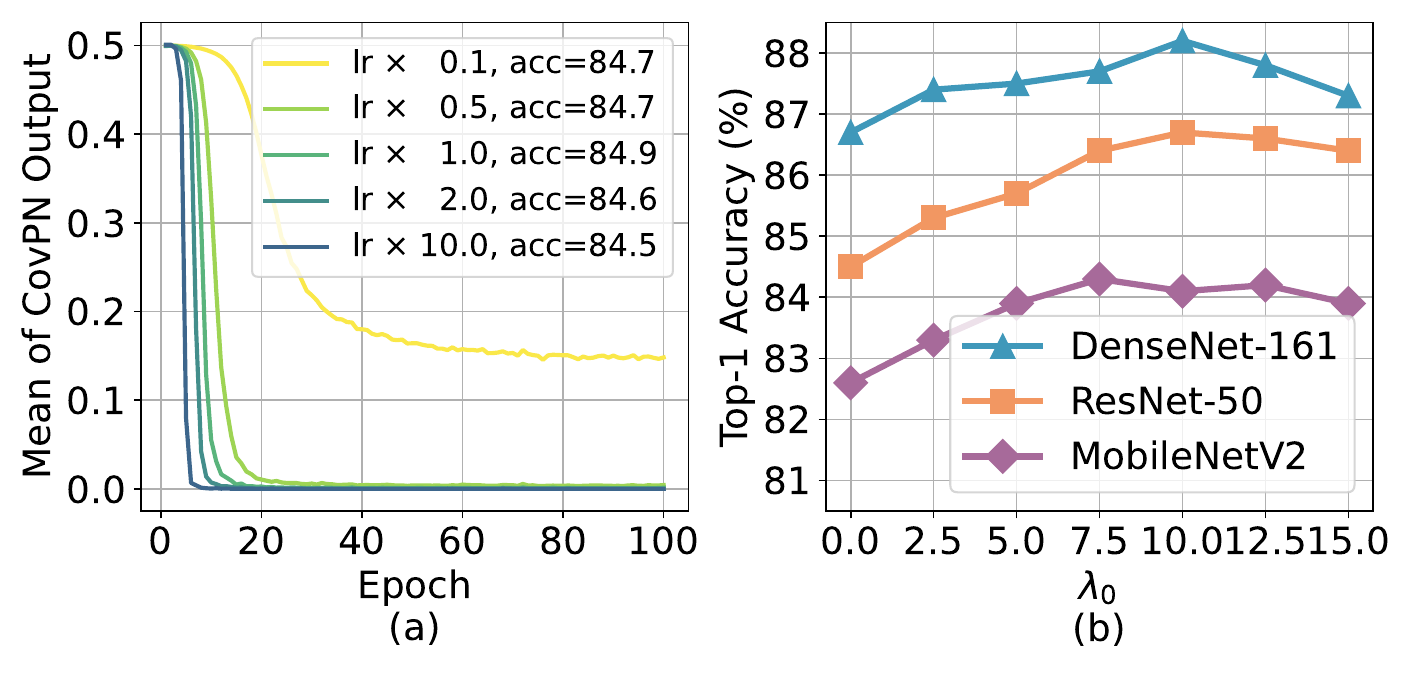}
  \end{center}
  \vskip -0.2in
  \caption{(a) Ablation studies of naive joint training. In the figure, \emph{lr} denotes the learning rate of $\bm{\theta}_{f}$, the number behind \emph{lr} is the learning rate proportion of $\bm{\theta}_{g}$ and \emph{acc} refers to the experiment results. (b) Ablation study on the strength of data augmentation $\lambda_0$. The best data augmentation strength is around 10.0.}
  \label{fig:abl_naive_train_and_lbd}
\end{figure}


\noindent\textbf{Superiority over online estimated covariance.} We compare our meta-learning-based sample-wise covariance matrix prediction method with the class-wise online estimation technique proposed in ISDA \cite{wang2019implicit}. When combining with P2P-Net \cite{yang2022fine}, ours method shows a superior performance comparing with the online estimation technique in ISDA \cite{wang2019implicit} (shown in Table.~\ref{tab:sota}).
We further conduct experiments on ResNets with different depths on CUB-200-2011, and the results are shown in \figurename~\ref{fig_resnets}. We could observe from the results that although ISDA in \cite{wang2019implicit} could improve the generalization ability over the baseline method, our proposed strategy could further surpass the it by a large margin. This phenomenon verifies the superiority of our sample-wise covariance matrix prediction over ISDA with online estimated covariance in the fine-grained scenario.

{
\noindent\textbf{Comparison with image-level data augmentation.} In Table~\ref{tab:com_imglvl}, we compare the performance of our method and some popular image-level data augmentation methods \cite{zhang2018mixup, yun2019cutmix, zhong2020random, cubuk2020randaugment}. It can be found that the proposed method is more effective than image-level counterparts in the fine-grained scenario.
}

\subsection{Ablation Studies}\label{sec_ablation}

We conduct ablation studies on our method to analyze how its variants affect the fine-grained visual classification result. We first show how optimizing the covariance matrix prediction network and the classification network simultaneously (mentioned in Section~\ref{ssec:CovNet}) would result in a zero-valued output. Then, we ablate the strength of augmentation $\lambda$, the growth scheduling of $\lambda$ and the structure of the proposed CovNet. Finally, the effect of accelerating the meta update process by freezing part of the pseudo network is presented.

\subsubsection{Naive joint training without meta-learning}
\label{sec:exp_naive}

\begin{figure*}[t]
    \begin{center}
    \includegraphics[width=\linewidth]{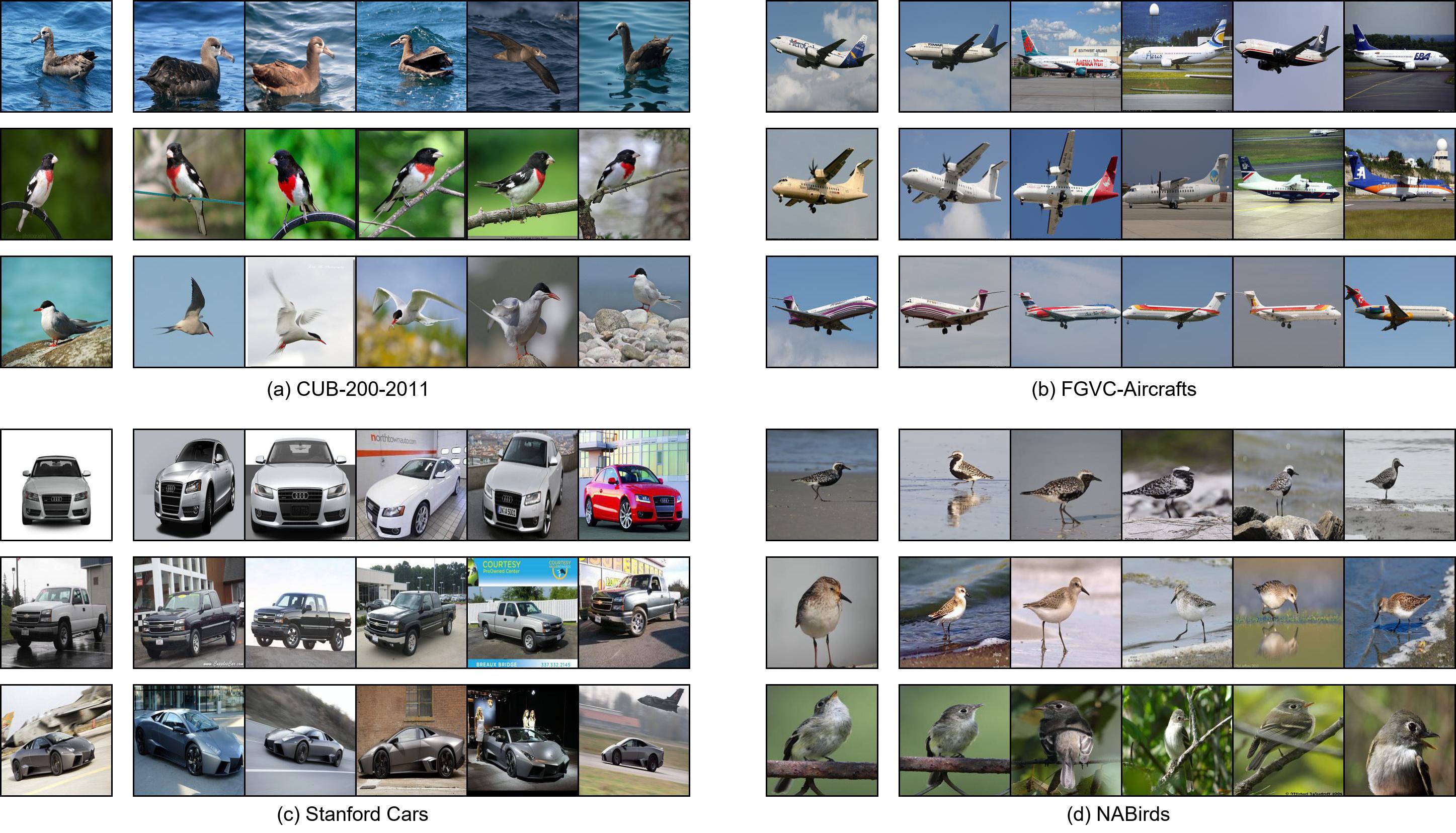}
    \end{center}
    \vskip -0.15in
	\caption{Visualization of the semantically augmented images. In each subfigure, the first column is the image corresponding to the original feature, and the following columns are the images corresponding to Top-5 nearest features of the augmented original one.}
	\label{fig:pixelspace}
\end{figure*}

\begin{figure*}[!h]
    \vskip 0.1in
    \begin{center}
    \includegraphics[width=\linewidth]{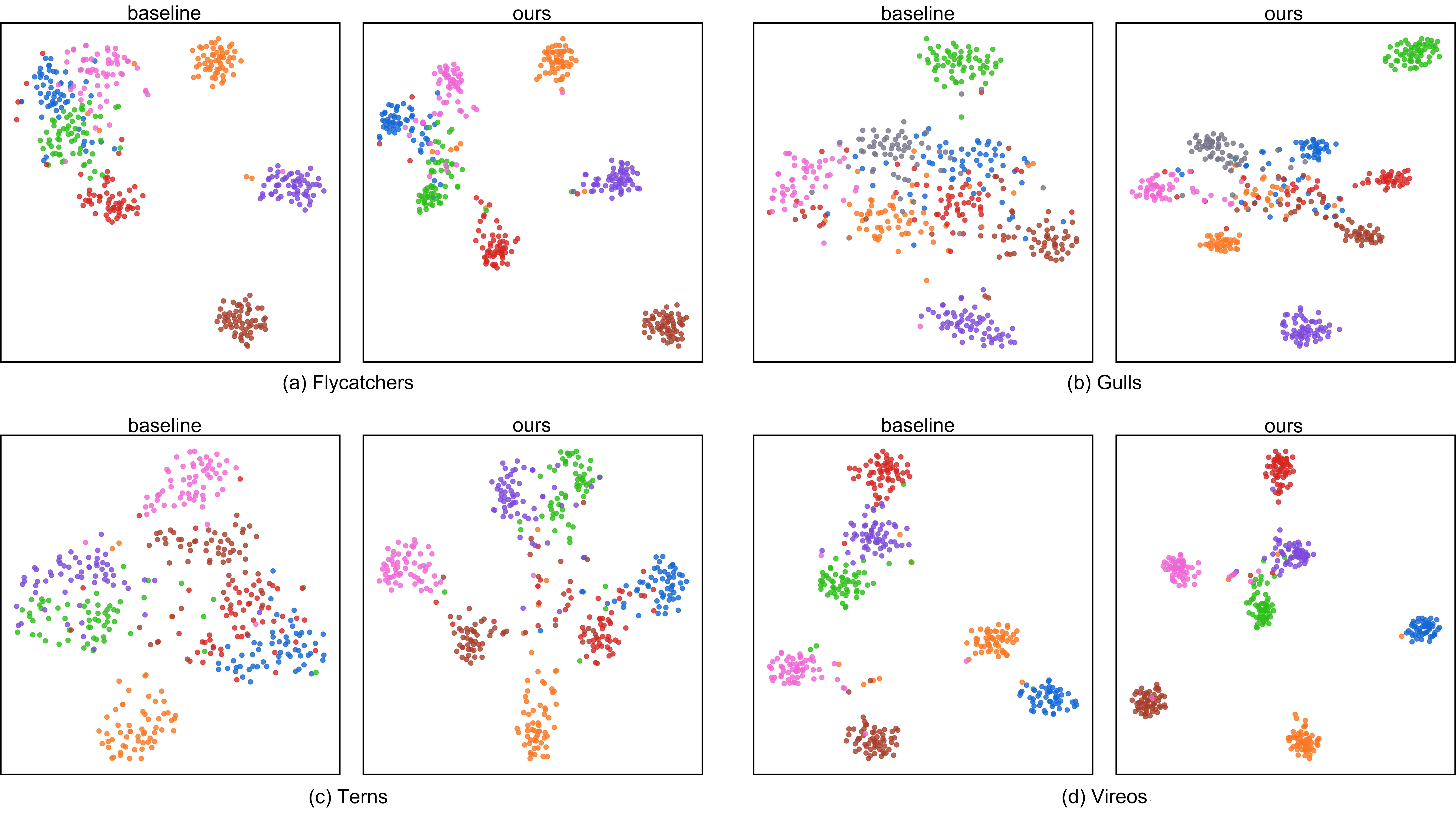}
    \end{center}
    \vskip -0.21in
	\caption{Visualization of the feature quality between the baseline training and our feature-level augmentation training on CUB-200-2011 using t-SNE  \cite{van2008visualizing}. Each subfigure corresponding to a meta-category (e.g. Flycatchers), and different colors refers to different sub-classes with the same meta-category (e.g. Acadian Flycatcher, Great Crested Flycatcher, Least Flycatcher, etc). In each subgraph, the left is the baseline result and the right is ours result, which demonstrates that our method enhances the intra-class compactnesss and the inter-class separability of the learned features.}
    \label{fig:tsne_bsl_ours}
\end{figure*}

We train the $\bm{\theta}_f$ and the $\bm{\theta}_g$ with the same target as described in Section~\ref{ssec:CovNet} and tuning the learning rate of $\bm{\theta}_g$ as a proportion of the learning rate of $\bm{\theta}_f$. The experiments are conducted on CUB-200-2011 dataset with a ResNet-50 classification network. In \figurename~\ref{fig:abl_naive_train_and_lbd}(a), we show the result from two perspectives: the mean of the CovNet output at the end of each epoch and the corresponding final accuracy. The mean of the CovNet output will converge to zero quickly if the learning rate is large and have a tendency to zero under a small learning rate, which verifies our claim in Section~\ref{ssec:CovNet}. Although the final classification accuracy is slightly higher than that of the baseline training method, it is inferior to ISDA~\cite{wang2019implicit} and remarkably worse than that of our proposed sample-wise predicted semantic data augmentation method.

\begin{figure*}[!ht]
    \begin{center}
    \includegraphics[width=\linewidth]{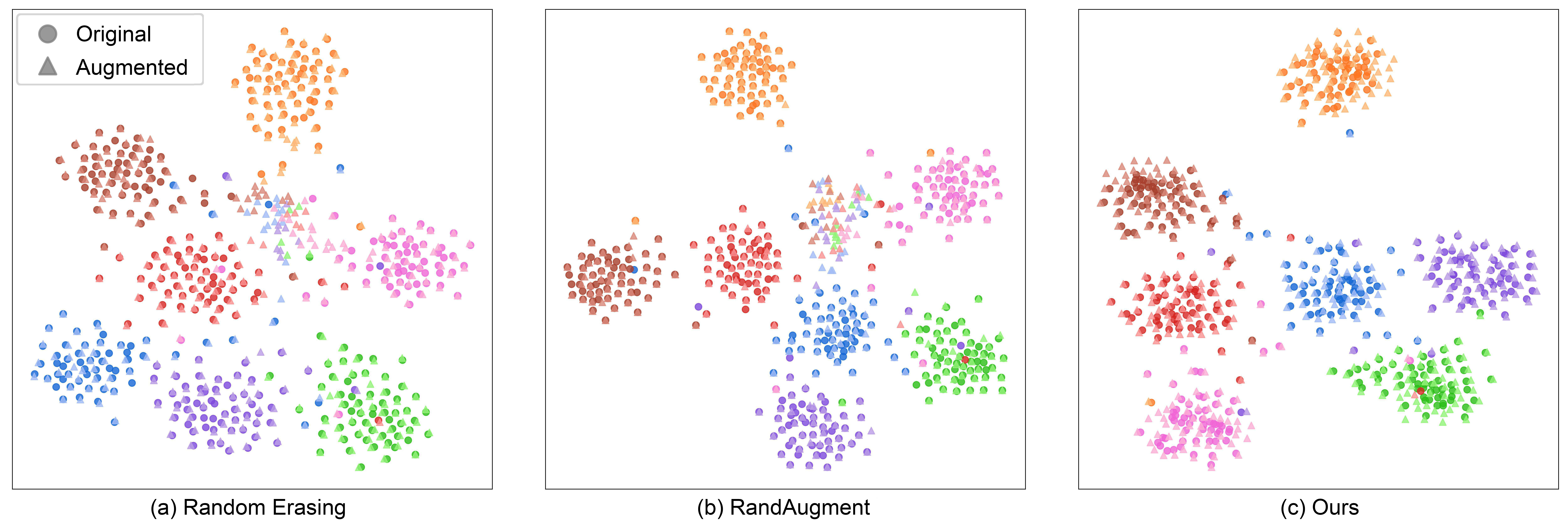}
    \end{center}
    \vskip -0.1in
	\caption{Visualization of the original image features ($\circ$) and the augmented features ($\vartriangle$) with different data augmentation techniques. Different colors refers to samples of different sub-categories. These data is obtained from the same meta-category (Wren) of the CUB-200-2011 dataset.}
	\label{fig:vis_aug}
\end{figure*}

\subsubsection{Influence of the strength of augmentation $\lambda$}

\begin{figure}
  \begin{center}
     \includegraphics[width=\linewidth]{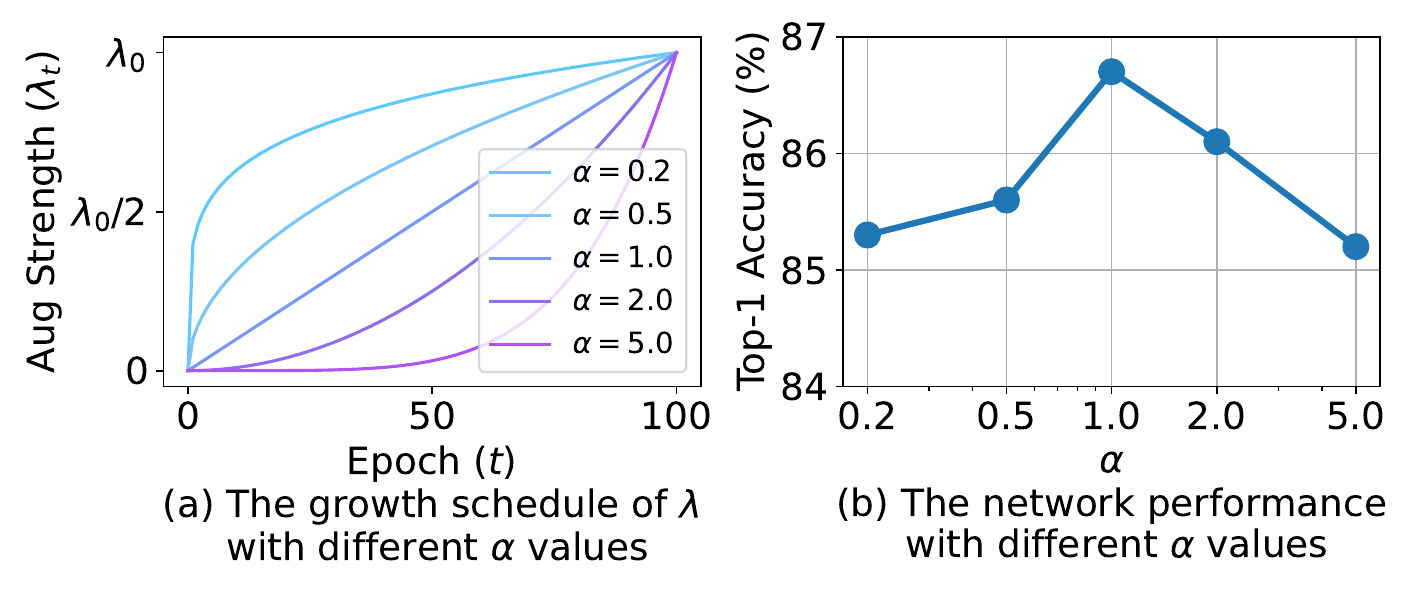}
  \end{center}
  \vskip -0.1in
  \caption{Ablation studies of the growth scheduling of the augmentation strength $\lambda$. (a) is the illustration of different scheduling strategies of $\lambda$. The scheduling function is $\lambda_t = (t/T)^{\alpha} \times \lambda_0$, where $\alpha$ controls the shape of the scheduling curve. Setting $\alpha=1.0$ leads to a linear growth scheduling for $\lambda$. (b) The network performance of different $\lambda$ schedule by changing $\alpha$ on ResNet-50.}
  \label{fig:abl_alpha}
  \vskip -0.10in
\end{figure}

\begin{figure}
  \begin{center}
     \includegraphics[width=\linewidth]{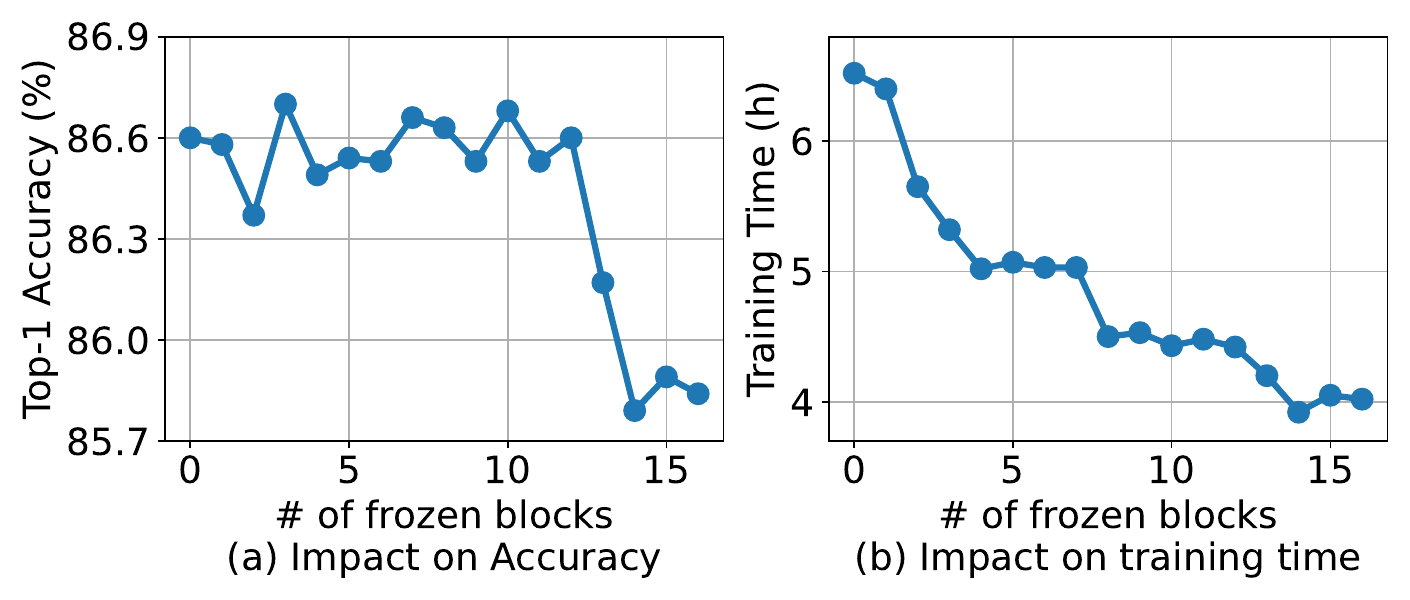}
  \end{center}
  \vskip -0.1in
  \caption{Ablation studies on the number of frozen blocks. A zero horizontal coordinate means we only freeze the network stem, and the horizontal coordinate of sixteen indicates we freeze the whole network, including the stem and all the sixteen residual blocks of a ResNet-50, excepts the last head.}
  \label{fig:abl_freeze}
  \vskip -0.15in
\end{figure}

In our work, the strength of augmentation $\lambda$ is linear increases along with the training epoch $\lambda_t = (t/T) \times \lambda_0$, where $\lambda_t$ is the augmentation strength of the current epoch, $t$ is the index of the current epoch, $T$ is the total training epoch and $\lambda_0$ is a hyperparameter to control the overall augmentation strength. We ablate the strength of semantic data augmentation by conducting experiments on CUB-200-2011 with three different models (i.e., ResNet-50, DenseNet-161 and MobileNetV2). As shown in Fig~\ref{fig:abl_naive_train_and_lbd}(b), as $\lambda_0$ increase from zero, the accuracy gradually increase and reach its peak around $\lambda_0 = 10.0$. As a result, in our experiments we simply set $\lambda_0$ as 10.0 expect for the P2P-Net. Despite the performance of the proposed method is affected by the augmentation strength $\lambda_0$, our method always outperforms the baseline ($\lambda_0 = 0$ in \figurename~\ref{fig:abl_naive_train_and_lbd}(b)).

\subsubsection{Influence of growth scheduling of $\lambda$}

Considering the simple linear growth strategy of $\lambda$ may not be optimal, we ablate the growth scheduling of $\lambda$ by tuning the hyperparameter $\alpha$ in the scheduling function $\lambda_t = (t/T)^{\alpha} \times \lambda_0$. The scheduling of different $\alpha$ is illustrated in \figurename~\ref{fig:abl_alpha} (a) and \figurename~\ref{fig:abl_alpha} (b) shows the corresponding results demonstrated on a ResNet-50 model with CUB-200-2011. The results show that the simple linear increasing schedule is superior to convex or concave growing counterparts. As a result, we choose the linear increasing schedule for all the experiments.

\subsubsection{Impact of the frozen blocks}
\label{sec:froze}

\begin{table}[t]
\centering
\caption{Ablation study on the structure of CovNet}
\label{tab:abl_CovNet}
\resizebox{\linewidth}{!}{
\begin{tabular}{c c|c c|c}
    \toprule
    Depth & Width & FLOPs ($\times 10^6$) & Params ($\times 10^6$) & Top-1 Accuracy \\
    \midrule
    1 & 256  &  2.10 & 1.05 & 86.7 \\
    \cellcolor{lightgray!50}1 & \cellcolor{lightgray!50}512  &  \cellcolor{lightgray!50}4.19 & \cellcolor{lightgray!50}2.10 & \cellcolor{lightgray!50}86.7  \\
    1 & 1024 &  8.39 & 4.20 & 86.5  \\
    1 & 2048 & 16.78 & 8.39 & 86.2  \\
    \midrule
    2 & 512 & 4.72 & 2.36 & 86.2 \\
    3 & 512 & 5.24 & 2.63 & 86.4 \\
    4 & 512 & 5.77 & 2.89 & 86.0 \\
    \bottomrule
\end{tabular}
}
\vskip -0.13in
\end{table}

The meta-learning framework introduces extra training cost because it has two extra update steps. The pseudo update process takes about the same time as the final real update process, while the meta update process takes more time because it needs to compute second order gradient. In a 4 Nvidia V100 GPU server, the vanilla meta-learning algorithm take 6.5 hour to train a ResNet-50 model in CUB-200-2011 for 100 epoch, while the baseline method takes 1.7 hour.
We freeze the ResNet-50 network stem and the first $n$ residual blocks and record the corresponding accuracy and training time on the CUB-200-2011 dataset. The results in \figurename~\ref{fig:abl_freeze} show that the network performance is similar to the counterpart without freezing when $0 \le n \leq 12$, while the total training time is monotonically decreasing. As a result, we freeze the first ten residual blocks of ResNet-50 for all the fine-grained recognition experiments to get a good accuracy-efficiency trade-off. In this way, the training cost in CUB-200-2011 is reduced from 6.5 hour to 4.5 hour. It is worth mention that our method do not add any extra cost in inference.
For other neural network structures, including P2P-Net \cite{yang2022fine}, we do not freeze any part of the network structure because finding the proper number of frozen blocks need extra experiments.

\subsubsection{Impact of the CovNet Structure}
\label{sec:covnet_archi}

We ablate the structure of the covariance matrix prediction network by varying its depth (the number of hidden layers) and width (the number of neurons of the hidden layer). Experimental results on CUB-200-2011 dataset with ResNet-50, which is shown in Table~\ref{tab:abl_CovNet}, demonstrate that although different CovNet structures could affect the results, the whole method is still effective. In practice we set the depth as one and the width of the hidden layer as a quarter of the final feature dimension for all the model architectures except otherwise mentioned.

\vskip -0.2in
\subsection{Evaluation on Generic Recognition Benchmark}
\label{sec_imagenet}

\begin{table}[t]
\centering
\caption{Experiment results on ImageNet.}
\label{tab:imagenet}
\begin{tabular}{c l l}
    \toprule
    Classification Model & Data Augmentation & Top-1 Accuracy \\
    \midrule
    \multirow{3}{*}{ResNet-50}
        &\quad \quad Basic                              & $\quad \quad 76.4$ \\
        &\quad \quad ISDA~\cite{wang2019implicit}       & $\quad \quad 76.7_{(\uparrow 0.3)}$ \\
        &\quad \quad \cellcolor{lightgray!50}Our method & $\quad \quad \cellcolor{lightgray!50}77.4_{( \textcolor{blue}{\uparrow 1.0})}$ \\
    \midrule
    \multirow{3}{*}{ResNeXt-50 (32x4d)}
        &\quad \quad Basic                              & $\quad \quad 77.5$ \\
        &\quad \quad ISDA~\cite{wang2019implicit}       & $\quad \quad 78.1_{(\uparrow 0.5)}$ \\
        &\quad \quad \cellcolor{lightgray!50}Our method & $\quad \quad \cellcolor{lightgray!50}78.7_{( \textcolor{blue}{\uparrow 1.2})}$ \\
    \bottomrule
\end{tabular}
\vskip -0.2 in
\end{table}

As the proposed sample-wise feature-level augmentation approach does not rely on additional fine-grained annotations (such as bounding boxes, part annotations, and hierarchical labels), it can be easily adapted to general image classification.  We evaluate our proposed method on ImageNet~\cite{deng2009imagenet} with a ResNet-50 model. For a fair comparison, we keep the same training configuration as ISDA~\cite{wang2019implicit}, in which the model is trained from scratch for 120 epochs with a momentum of 0.9 and a weight decay of 0.0001. The augmentation strength is set as 7.5, which is also the same as that in ISDA~\cite{wang2019implicit}. The CovNet $\bm{\theta}_{g}$ has the same architecture configuration as in the CUB-200-2011 and is optimized every 100 iterations with a learning rate of 0.0005. The results in Table~\ref{tab:imagenet} show that, on the generic image recognition problem, our sample-wise semantic data augmentation method is also effective over the class-wise estimated semantic directions approach~\cite{wang2019implicit}.

\subsection{Visualization Results}
\label{sec_visualization}

\textbf{Augmented samples in the pixel space.}
To demonstrate that our method is able to generate meaningful semantically augmented samples, we present the Top-5 nearest neighbors of the augmented features in image space on four fine-grained datasets (CUB-200-2011, FGVC Aircraft, Stanford Cars and NABirds). As shown in \figurename~\ref{fig:pixelspace}, our feature-level data augmentation strategy is able to adjust the semantics of training sample, such as visual angels, background, pose of the birds, painting of the aircraft, color of the cars.

\textbf{Quality of learned feature.}
We compare the learned feature quality between training with basic data augmentation (baseline) and training with our method. We select the meta-classes, which contains more than five sub-classes, on CUB-200-2011. We extract the deep features using the pre-trained network learned with the baseline method and our feature-level data augmentation method, respectively. These high-dimensional deep features are downscaled using t-SNE \cite{van2008visualizing}, and results are illustrated in \figurename~\ref{fig:tsne_bsl_ours}. We can find that our method effectively enhances the intra-class compactnesss and the inter-class separability of the learned features.

\textbf{Comparison with image-level data augmentation.}
Finally, we visualize the feature of different data augmentation methods, including Random Erasing \cite{zhong2020random}, RandAugment \cite{cubuk2020randaugment}, and our method. For fair comparison we use the same feature extractor pre-trained with basic augmentation. For image-level augmentations (Random Erasing and RandAugment), we extract the feature of original images and that of the augmented images and reduce its dimension by t-SNE. For our feature-level data augmentation method, we extract the feature of original images, diversify them with a pre-trained CovNet, then downscale both of them using t-SNE. The result in \figurename~\ref{fig:vis_aug} (a) (b) reveals that a portion of the image-level augmented samples would loss its discriminative region and thus are clustered together to form a new clustering center. This phenomenon verifies the discriminative region loss problem we proposed in \figurename~\ref{fig:motivation}. Furthermore, our feature-level data augmentation method always produce appropriate semantic directions and help the training samples to be translated into reasonable locations. Our method provide an ingenious solution to avoid the discriminative region loss problem induced by image-level data augmentation methods on fine-grained images.

\section{Conclusion}

In this paper, we propose a meta-learning based implicit data augmentation method for fine-grained image recognition. Our approach aims to cope with the discriminative region loss problem in the fine-grained scenario, which is induced by the random editing behavior of image-level data augmentation techniques. We diversify the training samples in the feature space rather than the image space to alleviate this problem. The sample-wise meaningful semantic direction is predicted by the covariance prediction network, which is joint optimized with the classification network in a meta-learning manner. Experiment results over multiple fine-grained benchmarks and neural network structures show the effectiveness of our proposed method on the fine-grained recognition problem.

\vskip 0.5 cm
\noindent\textbf{Acknowledgement.} This work is supported in part by the National Key R\&D Program of China under Grant 2021ZD0140407, the National Natural Science Foundation of China under Grants 62022048 and 62276150,  Guoqiang Institute of Tsinghua University and Beijing Academy of Artificial Intelligence.

\newpage
\bibliographystyle{ieeetr}
\small{
    \bibliography{reference}

\begin{thebibliography}{100}

\bibitem{wah2011caltech}
C.~Wah, S.~Branson, P.~Welinder, P.~Perona, and S.~Belongie, ``The caltech-ucsd
  birds-200-2011 dataset,'' 2011.

\bibitem{van2015building}
G.~Van~Horn, S.~Branson, R.~Farrell, S.~Haber, J.~Barry, P.~Ipeirotis,
  P.~Perona, and S.~Belongie, ``Building a bird recognition app and large scale
  dataset with citizen scientists: The fine print in fine-grained dataset
  collection,'' in {\em CVPR}, 2015.

\bibitem{KhoslaYaoJayadevaprakashFeiFei_FGVC2011}
A.~Khosla, N.~Jayadevaprakash, B.~Yao, and L.~Fei-Fei, ``Novel dataset for
  fine-grained image categorization,'' in {\em CVPR Workshops}, 2011.

\bibitem{krause20133d}
J.~Krause, M.~Stark, J.~Deng, and L.~Fei-Fei, ``3d object representations for
  fine-grained categorization,'' in {\em ICCV}, 2013.

\bibitem{wei2022rpc}
X.-S. Wei, Q.~Cui, L.~Yang, P.~Wang, L.~Liu, and J.~Yang, ``Rpc: a large-scale
  and fine-grained retail product checkout dataset,'' {\em SCIS}, 2022.

\bibitem{bai2020products}
Y.~Bai, Y.~Chen, W.~Yu, L.~Wang, and W.~Zhang, ``Products-10k: A large-scale
  product recognition dataset,'' {\em arXiv:2008.10545}, 2020.

\bibitem{zhao2017survey}
B.~Zhao, J.~Feng, X.~Wu, and S.~Yan, ``A survey on deep learning-based
  fine-grained object classification and semantic segmentation,'' {\em IJAC},
  2017.

\bibitem{zheng2018survey}
M.~Zheng, Q.~Li, Y.-a. Geng, H.~Yu, J.~Wang, J.~Gan, and W.~Xue, ``A survey of
  fine-grained image categorization,'' in {\em ICSP}, 2018.

\bibitem{wei2021fine}
X.-S. Wei, Y.-Z. Song, O.~Mac~Aodha, J.~Wu, Y.~Peng, J.~Tang, J.~Yang, and
  S.~Belongie, ``Fine-grained image analysis with deep learning: A survey,''
  {\em IEEE TPAMI}, 2021.

\bibitem{lecun2015deep}
Y.~LeCun, Y.~Bengio, and G.~Hinton, ``Deep learning,'' {\em Nature}.

\bibitem{huang2022glance}
G.~Huang, Y.~Wang, K.~Lv, H.~Jiang, W.~Huang, P.~Qi, and S.~Song, ``Glance and
  focus networks for dynamic visual recognition,'' {\em IEEE TPAMI}, 2022.

\bibitem{wang2021adaptive}
Y.~Wang, Z.~Chen, H.~Jiang, S.~Song, Y.~Han, and G.~Huang, ``Adaptive focus for
  efficient video recognition,'' in {\em ICCV}, 2021.

\bibitem{wang2021not}
Y.~Wang, R.~Huang, S.~Song, Z.~Huang, and G.~Huang, ``Not all images are worth
  16x16 words: Dynamic transformers for efficient image recognition,'' in {\em
  NeurIPS}, 2021.

\bibitem{wang2020glance}
Y.~Wang, K.~Lv, R.~Huang, S.~Song, L.~Yang, and G.~Huang, ``Glance and focus: a
  dynamic approach to reducing spatial redundancy in image classification,'' in
  {\em NeurIPS}, 2020.

\bibitem{han2022latency}
Y.~Han, Z.~Yuan, Y.~Pu, C.~Xue, S.~Song, G.~Sun, and G.~Huang, ``Latency-aware
  spatial-wise dynamic networks,'' in {\em NeurIPS}, 2022.

\bibitem{han2022learning}
Y.~Han, Y.~Pu, Z.~Lai, C.~Wang, S.~Song, J.~Cao, W.~Huang, C.~Deng, and
  G.~Huang, ``Learning to weight samples for dynamic early-exiting networks,''
  in {\em ECCV}, 2022.

\bibitem{pu2023adaptive}
Y.~Pu, Y.~Wang, Z.~Xia, Y.~Han, Y.~Wang, W.~Gan, Z.~Wang, S.~Song, and
  G.~Huang, ``Adaptive rotated convolution for rotated object detection,'' in
  {\em ICCV}, 2023.

\bibitem{han2023dynamic}
Y.~Han, D.~Han, Z.~Liu, Y.~Wang, X.~Pan, Y.~Pu, C.~Deng, J.~Feng, S.~Song, and
  G.~Huang, ``Dynamic perceiver for efficient visual recognition,'' in {\em
  ICCV}, 2023.

\bibitem{zhang2018mixup}
H.~Zhang, M.~Cisse, Y.~N. Dauphin, and D.~Lopez-Paz, ``mixup: Beyond empirical
  risk minimization,'' in {\em ICLR}, 2018.

\bibitem{yun2019cutmix}
S.~Yun, D.~Han, S.~J. Oh, S.~Chun, J.~Choe, and Y.~Yoo, ``Cutmix:
  Regularization strategy to train strong classifiers with localizable
  features,'' in {\em ICCV}, 2019.

\bibitem{cubuk2020randaugment}
E.~D. Cubuk, B.~Zoph, J.~Shlens, and Q.~Le, ``Randaugment: Practical automated
  data augmentation with a reduced search space,'' in {\em NeurIPS}, 2020.

\bibitem{zhong2020random}
Z.~Zhong, L.~Zheng, G.~Kang, S.~Li, and Y.~Yang, ``Random erasing data
  augmentation,'' in {\em AAAI}, 2020.

\bibitem{liu2022convnet}
Z.~Liu, H.~Mao, C.-Y. Wu, C.~Feichtenhofer, T.~Darrell, and S.~Xie, ``A convnet
  for the 2020s,'' in {\em CVPR}, 2022.

\bibitem{tan2021efficientnetv2}
M.~Tan and Q.~Le, ``Efficientnetv2: Smaller models and faster training,'' in
  {\em ICML}, 2021.

\bibitem{wang2022efficienttrain}
Y.~Wang, Y.~Yue, R.~Lu, T.~Liu, Z.~Zhong, S.~Song, and G.~Huang,
  ``Efficienttrain: Exploring generalized curriculum learning for training
  visual backbones,'' in {\em ICCV}, 2022.

\bibitem{liu2021swin}
Z.~Liu, Y.~Lin, Y.~Cao, H.~Hu, Y.~Wei, Z.~Zhang, S.~Lin, and B.~Guo, ``Swin
  transformer: Hierarchical vision transformer using shifted windows,'' in {\em
  ICCV}, 2021.

\bibitem{touvron2021training}
H.~Touvron, M.~Cord, M.~Douze, F.~Massa, A.~Sablayrolles, and H.~J{\'e}gou,
  ``Training data-efficient image transformers \& distillation through
  attention,'' in {\em ICML}, 2021.

\bibitem{devries2017improved}
T.~DeVries and G.~W. Taylor, ``Improved regularization of convolutional neural
  networks with cutout,'' {\em arXiv:1708.04552}, 2017.

\bibitem{cubuk2019autoaugment}
E.~D. Cubuk, B.~Zoph, D.~Mane, V.~Vasudevan, and Q.~V. Le, ``Autoaugment:
  Learning augmentation policies from data,'' in {\em CVPR}, 2019.

\bibitem{wang2019implicit}
Y.~Wang, X.~Pan, S.~Song, H.~Zhang, G.~Huang, and C.~Wu, ``Implicit semantic
  data augmentation for deep networks,'' in {\em NeurIPS}, 2019.

\bibitem{wei2019deep}
X.-S. Wei, J.~Wu, and Q.~Cui, ``Deep learning for fine-grained image analysis:
  A survey,'' {\em arXiv:1907.03069}, 2019.

\bibitem{maji2013fine}
S.~Maji, E.~Rahtu, J.~Kannala, M.~Blaschko, and A.~Vedaldi, ``Fine-grained
  visual classification of aircraft,'' {\em arXiv:1306.5151}, 2013.

\bibitem{he2016deep}
K.~He, X.~Zhang, S.~Ren, and J.~Sun, ``Deep residual learning for image
  recognition,'' in {\em CVPR}, 2016.

\bibitem{huang2017densely}
G.~Huang, Z.~Liu, L.~Van Der~Maaten, and K.~Q. Weinberger, ``Densely connected
  convolutional networks,'' in {\em CVPR}, 2017.

\bibitem{tan2019efficientnet}
M.~Tan and Q.~Le, ``Efficientnet: Rethinking model scaling for convolutional
  neural networks,'' in {\em ICML}, 2019.

\bibitem{radosavovic2020designing}
I.~Radosavovic, R.~P. Kosaraju, R.~Girshick, K.~He, and P.~Doll{\'a}r,
  ``Designing network design spaces,'' in {\em CVPR}, 2020.

\bibitem{dosovitskiy2020image}
A.~Dosovitskiy, L.~Beyer, A.~Kolesnikov, D.~Weissenborn, X.~Zhai,
  T.~Unterthiner, M.~Dehghani, M.~Minderer, G.~Heigold, S.~Gelly, {\em et~al.},
  ``An image is worth 16x16 words: Transformers for image recognition at
  scale,'' in {\em ICLR}, 2020.

\bibitem{yang2022fine}
X.~Yang, Y.~Wang, K.~Chen, Y.~Xu, and Y.~Tian, ``Fine-grained object
  classification via self-supervised pose alignment,'' in {\em CVPR}, 2022.

\bibitem{wang2019survey}
Y.~Wang and Z.~Wang, ``A survey of recent work on fine-grained image
  classification techniques,'' {\em JVCIR}, 2019.

\bibitem{zhang2014part}
N.~Zhang, J.~Donahue, R.~Girshick, and T.~Darrell, ``Part-based r-cnns for
  fine-grained category detection,'' in {\em ECCV}, 2014.

\bibitem{yang2018learning}
Z.~Yang, T.~Luo, D.~Wang, Z.~Hu, J.~Gao, and L.~Wang, ``Learning to navigate
  for fine-grained classification,'' in {\em ECCV}, 2018.

\bibitem{ding2021ap}
Y.~Ding, Z.~Ma, S.~Wen, J.~Xie, D.~Chang, Z.~Si, M.~Wu, and H.~Ling, ``Ap-cnn:
  Weakly supervised attention pyramid convolutional neural network for
  fine-grained visual classification,'' {\em IEEE TIP}, 2021.

\bibitem{chang2020devil}
D.~Chang, Y.~Ding, J.~Xie, A.~K. Bhunia, X.~Li, Z.~Ma, M.~Wu, J.~Guo, and Y.-Z.
  Song, ``The devil is in the channels: Mutual-channel loss for fine-grained
  image classification,'' {\em IEEE TIP}, 2020.

\bibitem{koniusz2021power}
P.~Koniusz and H.~Zhang, ``Power normalizations in fine-grained image, few-shot
  image and graph classification,'' {\em IEEE TPAMI}, 2021.

\bibitem{xu2016webly}
Z.~Xu, S.~Huang, Y.~Zhang, and D.~Tao, ``Webly-supervised fine-grained visual
  categorization via deep domain adaptation,'' {\em IEEE TPAMI}, 2016.

\bibitem{deng2015leveraging}
J.~Deng, J.~Krause, M.~Stark, and L.~Fei-Fei, ``Leveraging the wisdom of the
  crowd for fine-grained recognition,'' {\em IEEE TPAMI}, 2015.

\bibitem{he2022transfg}
J.~He, J.-N. Chen, S.~Liu, A.~Kortylewski, C.~Yang, Y.~Bai, and C.~Wang,
  ``Transfg: A transformer architecture for fine-grained recognition,'' in {\em
  AAAI}, 2022.

\bibitem{yu2023difficulty}
Z.~Yu, S.~Li, Y.~Shen, C.~H. Liu, and S.~Wang, ``On the difficulty of unpaired
  infrared-to-visible video translation: Fine-grained content-rich patches
  transfer,'' in {\em CVPR}, 2023.

\bibitem{li2018domain}
S.~Li, S.~Song, G.~Huang, Z.~Ding, and C.~Wu, ``Domain invariant and class
  discriminative feature learning for visual domain adaptation,'' {\em IEEE
  TIP}, 2018.

\bibitem{ge2019weakly}
W.~Ge, X.~Lin, and Y.~Yu, ``Weakly supervised complementary parts models for
  fine-grained image classification from the bottom up,'' in {\em CVPR}, 2019.

\bibitem{wang2020graph}
Z.~Wang, S.~Wang, H.~Li, Z.~Dou, and J.~Li, ``Graph-propagation based
  correlation learning for weakly supervised fine-grained image
  classification,'' in {\em AAAI}, 2020.

\bibitem{liu2020filtration}
C.~Liu, H.~Xie, Z.-J. Zha, L.~Ma, L.~Yu, and Y.~Zhang, ``Filtration and
  distillation: Enhancing region attention for fine-grained visual
  categorization,'' in {\em AAAI}, 2020.

\bibitem{wang2018learning}
Y.~Wang, V.~I. Morariu, and L.~S. Davis, ``Learning a discriminative filter
  bank within a cnn for fine-grained recognition,'' in {\em CVPR}, 2018.

\bibitem{ding2019selective}
Y.~Ding, Y.~Zhou, Y.~Zhu, Q.~Ye, and J.~Jiao, ``Selective sparse sampling for
  fine-grained image recognition,'' in {\em ICCV}, 2019.

\bibitem{huang2020interpretable}
Z.~Huang and Y.~Li, ``Interpretable and accurate fine-grained recognition via
  region grouping,'' in {\em CVPR}, 2020.

\bibitem{zhang2019learning}
L.~Zhang, S.~Huang, W.~Liu, and D.~Tao, ``Learning a mixture of
  granularity-specific experts for fine-grained categorization,'' in {\em
  ICCV}, 2019.

\bibitem{zheng2019looking}
H.~Zheng, J.~Fu, Z.-J. Zha, and J.~Luo, ``Looking for the devil in the details:
  Learning trilinear attention sampling network for fine-grained image
  recognition,'' in {\em CVPR}, 2019.

\bibitem{zheng2019learning}
H.~Zheng, J.~Fu, Z.-J. Zha, J.~Luo, and T.~Mei, ``Learning rich part
  hierarchies with progressive attention networks for fine-grained image
  recognition,'' {\em IEEE TIP}, 2019.

\bibitem{ji2020attention}
R.~Ji, L.~Wen, L.~Zhang, D.~Du, Y.~Wu, C.~Zhao, X.~Liu, and F.~Huang,
  ``Attention convolutional binary neural tree for fine-grained visual
  categorization,'' in {\em CVPR}, 2020.

\bibitem{yu2018hierarchical}
C.~Yu, X.~Zhao, Q.~Zheng, P.~Zhang, and X.~You, ``Hierarchical bilinear pooling
  for fine-grained visual recognition,'' in {\em ECCV}, 2018.

\bibitem{wei2018grassmann}
X.~Wei, Y.~Zhang, Y.~Gong, J.~Zhang, and N.~Zheng, ``Grassmann pooling as
  compact homogeneous bilinear pooling for fine-grained visual
  classification,'' in {\em ECCV}, 2018.

\bibitem{zheng2019learningdeep}
H.~Zheng, J.~Fu, Z.-J. Zha, and J.~Luo, ``Learning deep bilinear transformation
  for fine-grained image representation,'' in {\em NeurIPS}, 2019.

\bibitem{min2020multi}
S.~Min, H.~Yao, H.~Xie, Z.-J. Zha, and Y.~Zhang, ``Multi-objective matrix
  normalization for fine-grained visual recognition,'' {\em IEEE TIP}, 2020.

\bibitem{sun2020text}
L.~Sun, X.~Guan, Y.~Yang, and L.~Zhang, ``Text-embedded bilinear model for
  fine-grained visual recognition,'' in {\em ACM MM}, 2020.

\bibitem{dubey2018maximum}
A.~Dubey, O.~Gupta, R.~Raskar, and N.~Naik, ``Maximum-entropy fine grained
  classification,'' in {\em NeurIPS}, 2018.

\bibitem{sun2020fine}
G.~Sun, H.~Cholakkal, S.~Khan, F.~Khan, and L.~Shao, ``Fine-grained
  recognition: Accounting for subtle differences between similar classes,'' in
  {\em AAAI}, 2020.

\bibitem{zhuang2020learning}
P.~Zhuang, Y.~Wang, and Y.~Qiao, ``Learning attentive pairwise interaction for
  fine-grained classification,'' in {\em AAAI}, 2020.

\bibitem{xu2023multi}
M.~Xu, L.~Qin, W.~Chen, S.~Pu, and L.~Zhang, ``Multi-view adversarial
  discriminator: Mine the non-causal factors for object detection in unseen
  domains,'' in {\em CVPR}, 2023.

\bibitem{shorten2019survey}
C.~Shorten and T.~M. Khoshgoftaar, ``A survey on image data augmentation for
  deep learning,'' {\em J. Big Data}, 2019.

\bibitem{perez2017effectiveness}
L.~Perez and J.~Wang, ``The effectiveness of data augmentation in image
  classification using deep learning,'' {\em arXiv:1712.04621}, 2017.

\bibitem{taylor2018improving}
L.~Taylor and G.~Nitschke, ``Improving deep learning with generic data
  augmentation,'' in {\em SSCI}, 2018.

\bibitem{inoue2018data}
H.~Inoue, ``Data augmentation by pairing samples for images classification,''
  {\em arXiv:1801.02929}, 2018.

\bibitem{summers2019improved}
C.~Summers and M.~J. Dinneen, ``Improved mixed-example data augmentation,'' in
  {\em WACV}, 2019.

\bibitem{takahashi2019data}
R.~Takahashi, T.~Matsubara, and K.~Uehara, ``Data augmentation using random
  image cropping and patching for deep cnns,'' {\em TCSVT}, 2019.

\bibitem{mikolajczyk2018data}
A.~Miko{\l}ajczyk and M.~Grochowski, ``Data augmentation for improving deep
  learning in image classification problem,'' in {\em IIPhDW}, 2018.

\bibitem{devries2017dataset}
T.~DeVries and G.~W. Taylor, ``Dataset augmentation in feature space,'' in {\em
  ICLR Workshop}, 2017.

\bibitem{wong2016understanding}
S.~C. Wong, A.~Gatt, V.~Stamatescu, and M.~D. McDonnell, ``Understanding data
  augmentation for classification: when to warp?,'' in {\em DICTA}, 2016.

\bibitem{bowles2018gan}
C.~Bowles, L.~Chen, R.~Guerrero, P.~Bentley, R.~Gunn, A.~Hammers, D.~A. Dickie,
  M.~V. Hern{\'a}ndez, J.~Wardlaw, and D.~Rueckert, ``Gan augmentation:
  Augmenting training data using generative adversarial networks,'' {\em
  arXiv:1810.10863}, 2018.

\bibitem{lim2018doping}
S.~K. Lim, Y.~Loo, N.-T. Tran, N.-M. Cheung, G.~Roig, and Y.~Elovici, ``Doping:
  Generative data augmentation for unsupervised anomaly detection with gan,''
  in {\em ICDM}, 2018.

\bibitem{zhang2019dada}
X.~Zhang, Z.~Wang, D.~Liu, and Q.~Ling, ``Dada: Deep adversarial data
  augmentation for extremely low data regime classification,'' in {\em ICASSP},
  2019.

\bibitem{verma2019manifold}
V.~Verma, A.~Lamb, C.~Beckham, A.~Najafi, I.~Mitliagkas, D.~Lopez-Paz, and
  Y.~Bengio, ``Manifold mixup: Better representations by interpolating hidden
  states,'' in {\em ICML}, 2019.

\bibitem{zhang2023style}
L.~Zhang, Z.~Liu, W.~Zhang, and D.~Zhang, ``Style uncertainty based self-paced
  meta learning for generalizable person re-identification,'' {\em IEEE TIP},
  2023.

\bibitem{simonyan2015very}
K.~Simonyan and A.~Zisserman, ``Very deep convolutional networks for
  large-scale image recognition,'' in {\em ICLR}, 2015.

\bibitem{li2021metasaug}
S.~Li, K.~Gong, C.~H. Liu, Y.~Wang, F.~Qiao, and X.~Cheng, ``{MetaSAug}: Meta
  semantic augmentation for long-tailed visual recognition,'' in {\em CVPR},
  2021.

\bibitem{li2021transferable}
S.~Li, M.~Xie, K.~Gong, C.~H. Liu, Y.~Wang, and W.~Li, ``Transferable semantic
  augmentation for domain adaptation,'' in {\em CVPR}, 2021.

\bibitem{finn2017model}
C.~Finn, P.~Abbeel, and S.~Levine, ``Model-agnostic meta-learning for fast
  adaptation of deep networks,'' in {\em ICML}, 2017.

\bibitem{franceschi2018bilevel}
L.~Franceschi, P.~Frasconi, S.~Salzo, R.~Grazzi, and M.~Pontil, ``Bilevel
  programming for hyperparameter optimization and meta-learning,'' in {\em
  ICML}, 2018.

\bibitem{liu2018darts}
H.~Liu, K.~Simonyan, and Y.~Yang, ``Darts: Differentiable architecture
  search,'' in {\em ICLR}, 2018.

\bibitem{snell2017prototypical}
J.~Snell, K.~Swersky, and R.~Zemel, ``Prototypical networks for few-shot
  learning,'' in {\em NeurIPS}, 2017.

\bibitem{metz2018meta}
L.~Metz, N.~Maheswaranathan, B.~Cheung, and J.~Sohl-Dickstein, ``Meta-learning
  update rules for unsupervised representation learning,'' in {\em ICLR}, 2018.

\bibitem{duan2016rl}
Y.~Duan, J.~Schulman, X.~Chen, P.~L. Bartlett, I.~Sutskever, and P.~Abbeel,
  ``Rl2: Fast reinforcement learning via slow reinforcement learning,'' {\em
  arXiv:1611.02779}, 2016.

\bibitem{houthooft2018evolved}
R.~Houthooft, Y.~Chen, P.~Isola, B.~Stadie, F.~Wolski, O.~Jonathan~Ho, and
  P.~Abbeel, ``Evolved policy gradients,'' in {\em NeurIPS}, 2018.

\bibitem{alet2019meta}
F.~Alet, M.~F. Schneider, T.~Lozano-Perez, and L.~P. Kaelbling, ``Meta-learning
  curiosity algorithms,'' in {\em ICLR}, 2019.

\bibitem{real2019regularized}
E.~Real, A.~Aggarwal, Y.~Huang, and Q.~V. Le, ``Regularized evolution for image
  classifier architecture search,'' in {\em AAAI}, 2019.

\bibitem{zoph2017neural}
B.~Zoph and Q.~V. Le, ``Neural architecture search with reinforcement
  learning,'' in {\em ICLR}, 2017.

\bibitem{lemke2015metalearning}
C.~Lemke, M.~Budka, and B.~Gabrys, ``Metalearning: a survey of trends and
  technologies,'' {\em Artificial intelligence review}, 2015.

\bibitem{hospedales2021meta}
T.~M. Hospedales, A.~Antoniou, P.~Micaelli, and A.~J. Storkey, ``Meta-learning
  in neural networks: A survey,'' {\em IEEE TPAMI}, 2021.

\bibitem{ravi2017optimization}
S.~Ravi and H.~Larochelle, ``Optimization as a model for few-shot learning,''
  in {\em ICLR}, 2017.

\bibitem{mishra2018simple}
N.~Mishra, M.~Rohaninejad, X.~Chen, and P.~Abbeel, ``A simple neural attentive
  meta-learner,'' in {\em ICLR}, 2018.

\bibitem{li2021meta}
S.~Li, W.~Ma, J.~Zhang, C.~H. Liu, J.~Liang, and G.~Wang, ``Meta-reweighted
  regularization for unsupervised domain adaptation,'' {\em IEEE TKDE}, 2021.

\bibitem{thrun1998learning}
S.~Thrun and L.~Pratt, ``Learning to learn: Introduction and overview,'' in
  {\em Learning to learn}, 1998.

\bibitem{andrychowicz2016learning}
M.~Andrychowicz, M.~Denil, S.~Gomez, M.~W. Hoffman, D.~Pfau, T.~Schaul,
  B.~Shillingford, and N.~De~Freitas, ``Learning to learn by gradient descent
  by gradient descent,'' in {\em NeurIPS}, 2016.

\bibitem{shu2019meta}
J.~Shu, Q.~Xie, L.~Yi, Q.~Zhao, S.~Zhou, Z.~Xu, and D.~Meng, ``Meta-weight-net:
  Learning an explicit mapping for sample weighting,'' in {\em NeurIPS}, 2019.

\bibitem{sandler2018mobilenetv2}
M.~Sandler, A.~Howard, M.~Zhu, A.~Zhmoginov, and L.-C. Chen, ``Mobilenetv2:
  Inverted residuals and linear bottlenecks,'' in {\em CVPR}, 2018.

\bibitem{lin2015bilinear}
T.-Y. Lin, A.~RoyChowdhury, and S.~Maji, ``Bilinear cnn models for fine-grained
  visual recognition,'' in {\em ICCV}, 2015.

\bibitem{fu2017look}
J.~Fu, H.~Zheng, and T.~Mei, ``Look closer to see better: Recurrent attention
  convolutional neural network for fine-grained image recognition,'' in {\em
  CVPR}, 2017.

\bibitem{zheng2017learning}
H.~Zheng, J.~Fu, T.~Mei, and J.~Luo, ``Learning multi-attention convolutional
  neural network for fine-grained image recognition,'' in {\em ICCV}, 2017.

\bibitem{wei2018mask}
X.-S. Wei, C.-W. Xie, J.~Wu, and C.~Shen, ``Mask-cnn: Localizing parts and
  selecting descriptors for fine-grained bird species categorization,'' {\em
  Pattern Recognition}, 2018.

\bibitem{chen2019destruction}
Y.~Chen, Y.~Bai, W.~Zhang, and T.~Mei, ``Destruction and construction learning
  for fine-grained image recognition,'' in {\em CVPR}, 2019.

\bibitem{wang2020weakly}
Z.~Wang, S.~Wang, S.~Yang, H.~Li, J.~Li, and Z.~Li, ``Weakly supervised
  fine-grained image classification via guassian mixture model oriented
  discriminative learning,'' in {\em CVPR}, 2020.

\bibitem{du2020fine}
R.~Du, D.~Chang, A.~K. Bhunia, J.~Xie, Z.~Ma, Y.-Z. Song, and J.~Guo,
  ``Fine-grained visual classification via progressive multi-granularity
  training of jigsaw patches,'' in {\em ECCV}, 2020.

\bibitem{du2021progressive}
R.~Du, J.~Xie, Z.~Ma, D.~Chang, Y.-Z. Song, and J.~Guo, ``Progressive learning
  of category-consistent multi-granularity features for fine-grained visual
  classification,'' {\em IEEE TPAMI}, 2021.

\bibitem{rao2021counterfactual}
Y.~Rao, G.~Chen, J.~Lu, and J.~Zhou, ``Counterfactual attention learning for
  fine-grained visual categorization and re-identification,'' in {\em ICCV},
  2021.

\bibitem{liu2022convolutional}
K.~Liu, K.~Chen, and K.~Jia, ``Convolutional fine-grained classification with
  self-supervised target relation regularization,'' {\em IEEE TIP}, 2022.

\bibitem{ke2023granularity}
X.~Ke, Y.~Cai, B.~Chen, H.~Liu, and W.~Guo, ``Granularity-aware distillation
  and structure modeling region proposal network for fine-grained image
  classification,'' {\em Pattern Recognition}, 2023.

\bibitem{van2008visualizing}
L.~Van~der Maaten and G.~Hinton, ``Visualizing data using t-sne.,'' {\em JMLR},
  2008.

\bibitem{deng2009imagenet}
J.~Deng, W.~Dong, R.~Socher, L.-J. Li, K.~Li, and L.~Fei-Fei, ``Imagenet: A
  large-scale hierarchical image database,'' in {\em CVPR}, 2009.

\bibitem{mairal2013stochastic}
J.~Mairal, ``Stochastic majorization-minimization algorithms for large-scale
  optimization,'' in {\em NeurIPS}, 2013.

\bibitem{bertsekas2016nonlinear}
D.~Bertsekas, {\em Nonlinear Programming}.
\newblock Athena Scientific, 2016.

\end{thebibliography}
}

%

\begin{IEEEbiography}[{\includegraphics[width=1in,height=1.25in,clip,keepaspectratio]{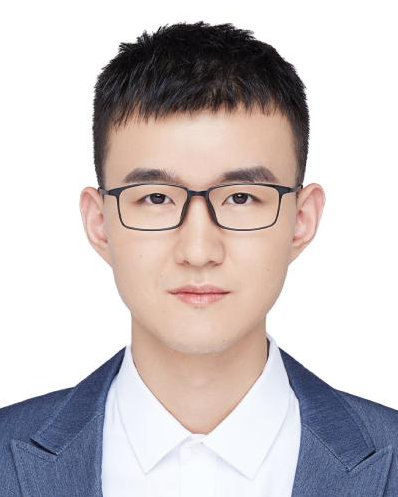}}]{Yifan Pu}
received the B.S. degree in automation from Beihang University, Beijing, China, in 2020. He is currently pursuing the M.S. degree with the Department of Automation, Tsinghhua University, Beijing, China. His research interests include computer vision, machine learning and deep learning.
\end{IEEEbiography}

\vspace{-10mm}
\begin{IEEEbiography}[{\includegraphics[width=1in,height=1.25in,clip,keepaspectratio]{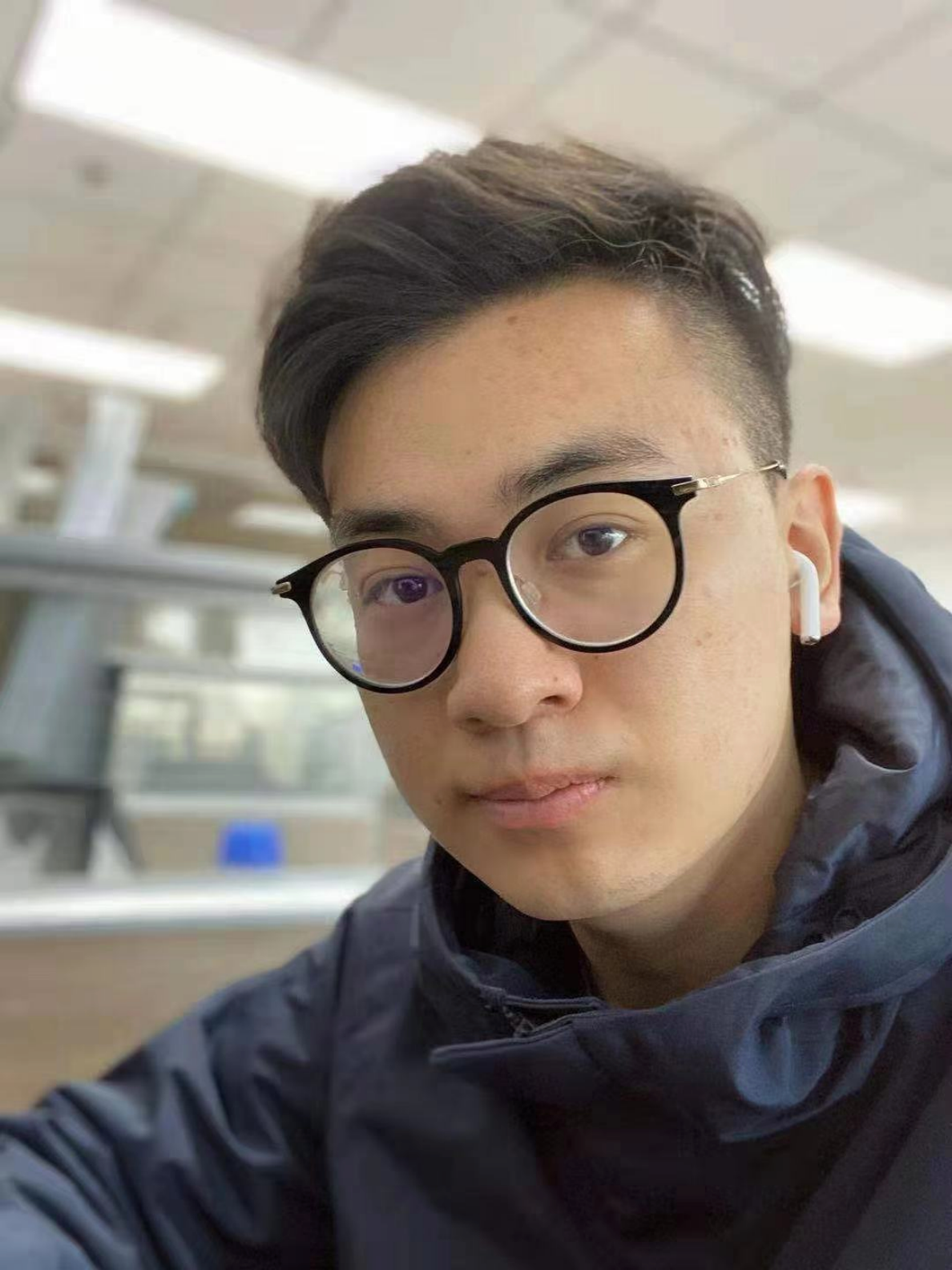}}]{Yizeng Han}
received the B.S. degree from the Department of Automation, Tsinghua University, Beijing, China, in 2018. And he is currently pursuing the Ph.D. degree in control science and engineering with the Department of Automation, Institute of System Integration in Tsinghua University. His current research interests include computer vision and deep learning, especially in dynamic neural networks.
\end{IEEEbiography}

\vspace{-10mm}
\begin{IEEEbiography}[{\includegraphics[width=1in,height=1.25in,clip,keepaspectratio]{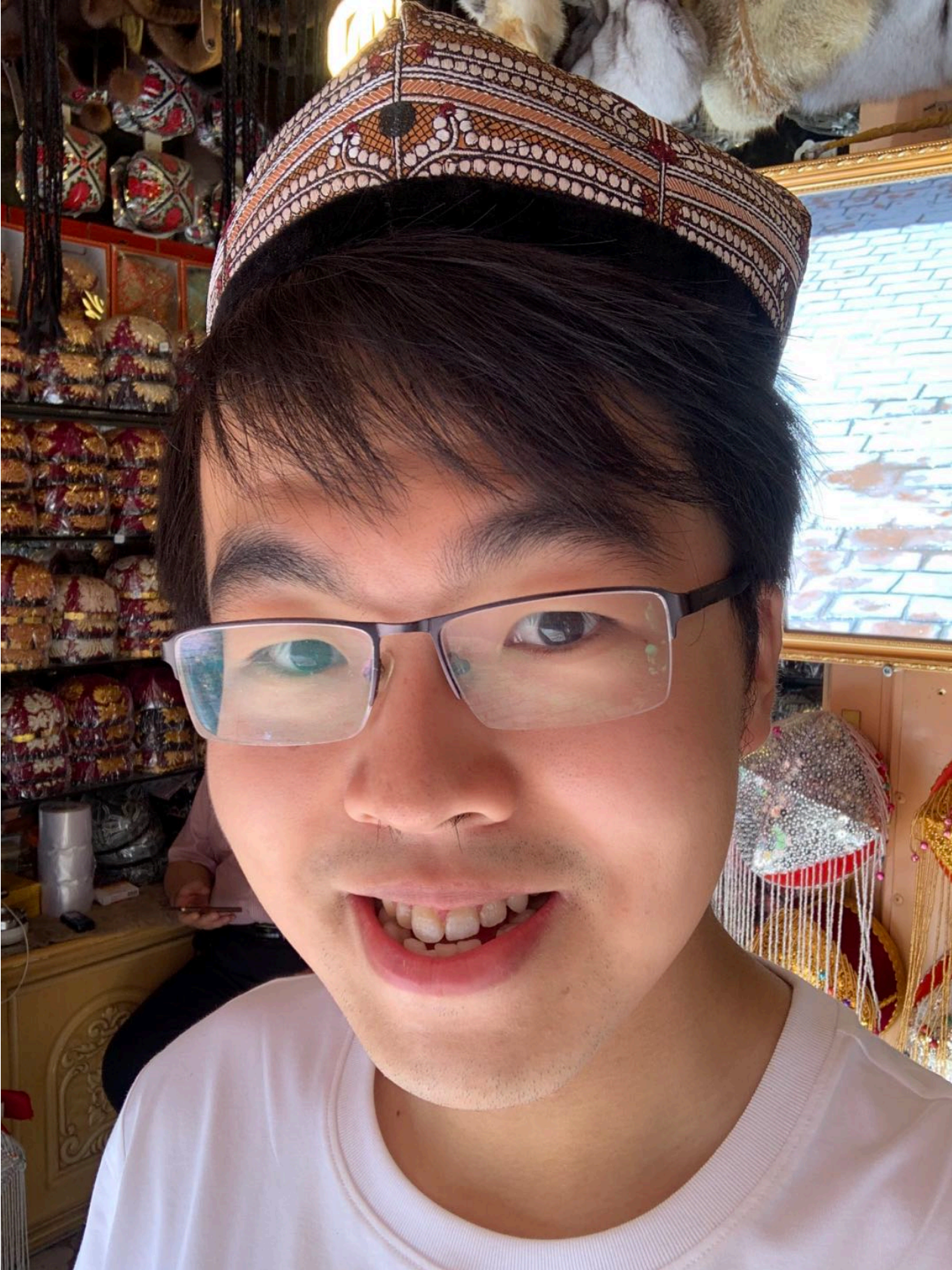}}]{Yulin Wang}
received his B.S. degree in Automation from Beihang University, Beijing, China, in 2019. He is currently pursuing the Ph.D. degree in the Department of Automation, Tsinghhua university. He was a Visiting Student at U.C. Berkeley, Berkeley, CA, USA, in 2018. His research interests include computer vision and deep learning.
\end{IEEEbiography}



\vspace{-10mm}
\begin{IEEEbiography}[{\includegraphics[width=1in,height=1.25in,clip,keepaspectratio]{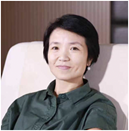}}]{Junlan Feng}(Fellow, IEEE)
received her Ph.D. on Speech Recognition from Chinese Academy of Sciences in 2001. She had been a principal researcher at AT\&T Labs Research and has been the chief scientist of China Mobile Research since 2013.
\end{IEEEbiography}

\vspace{-10mm}
\begin{IEEEbiography}[{\includegraphics[width=1in,height=1.25in,clip,keepaspectratio]{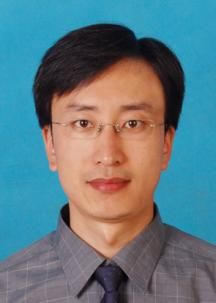}}]{Chao Deng}
received the M.S. degree and the Ph.D. degree from Harbin Institute of Technology, Harbin, China, in 2003 and 2009 respectively. He is currently a deputy general manager with AI center of China Mobile Research Institute. His research interests include machine learning and artificial intelligence for ICT operations.
\end{IEEEbiography}

\vspace{-10mm}
\begin{IEEEbiography}[{\includegraphics[width=1in,height=1.25in,clip,keepaspectratio]{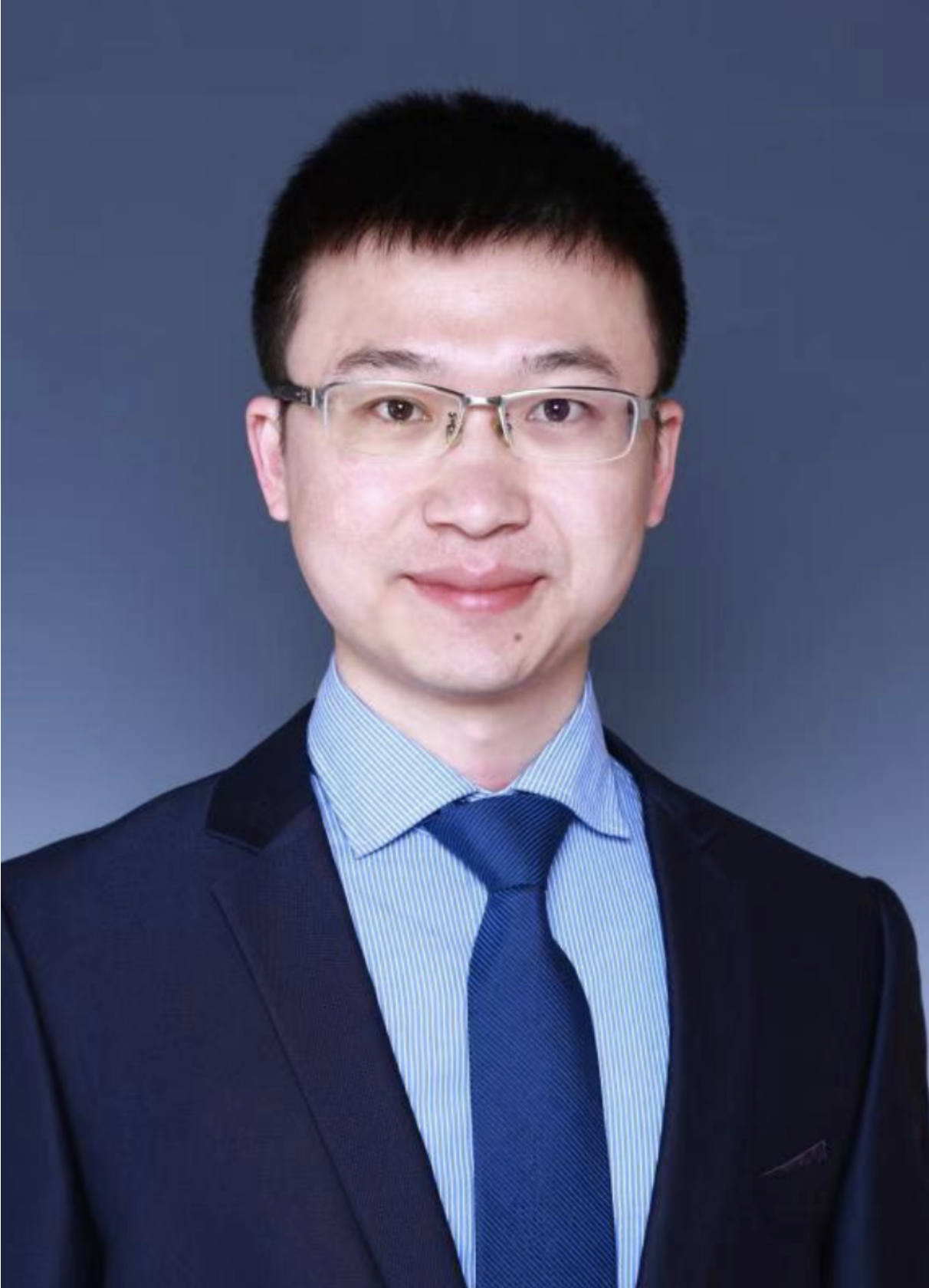}}]{Gao Huang}
received the B.S. degree in automation from Beihang University, Beijing, China, in 2009, and the Ph.D. degree in automation from Tsinghua University, Beijing, in 2015.

He was a Visiting Research Scholar with the Department of Computer Science and Engineering, Washington University in St. Louis, St. Louis, MO, USA, in 2013 and a Post-Doctoral Researcher with the Department of Computer Science, Cornell University, Ithaca, NY, USA, from 2015 to 2018. He is currently an Associate Professor with the Department of Automation, Tsinghua University, Beijing. His current research interests include machine learning and
deep learning.
\end{IEEEbiography}







\onecolumn
\appendix[Convergence Proof of the proposed method]
\label{sec:appendix}

\noindent
\textbf{Lemma 1. (Deterministic Lemma on Non-negative Converging Series) }(Lemma A.5 in \cite{mairal2013stochastic}) Let $(a_n)_{n \geq 1}$, $(b_n)_{n \geq 1}$ be two non-negative real sequences such that the series $\sum_{n=1}^{\infty} a_n$ diverges, the serious $\sum_{n=1}^{\infty} a_n b_n$ converges, and there exists $K > 0$ such that $\| b_{n+1} - b_{n} \| \leq K a_n$. Then, the sequence $(b_n)_{n \geq 1}$ converges to 0.
\\

\noindent
\textbf{Lemma 2. (Descent Lemma) }(Proposition A.24 in \cite{bertsekas2016nonlinear}) Lef $f: \mathbb{R}^n \rightarrow \mathbb{R}$ be continuously differentiable, and let $\bm{x}$ and  $\bm{y}$ be two vectors in $\mathbb{R}^n$. Suppose that $f$ is Lipschitz continuous, that is
$|| \nabla f(\bm{x}) - \nabla f(\bm{y})||_2 \leq L || \bm{x} - \bm{y} ||_2$, where $L$ is some scalar. Then
\begin{align}
\begin{split}
f(\bm{y}) - f(\bm{x}) \leq \left \langle \nabla f(\bm{x}), \bm{y} - \bm{x} \right \rangle + \frac{L}{2} || \bm{y} - \bm{x} ||_2^2.
\end{split}
\end{align}

\noindent \emph{Proof.}
The Lipschitz continuous condition can be rewritten as $|| \nabla f(\bm{x} + \mu \bm{z}) - \nabla f(\bm{x})||_2 \leq L \mu || \bm{z} ||, \forall \mu \in [0,1]$.
Let $t$ be a scalar parameter and let $g(t)=f(\bm{x}+t\bm{z})$. The chain rule yields 
\begin{align}
\begin{split}
\frac{dg(t)}{dt}
= \left \langle \nabla f(\bm{x}+t\bm{z}), \bm{z} \right \rangle.
\end{split}
\end{align}

We have

\begin{align}
\begin{split}
f(\bm{x}+\bm{z}) - f(\bm{x}) &= g(1) - g(0)
= \int_0^1 \frac{dg(t)}{dt} dt
= \int_0^1 \left \langle \nabla f(\bm{x}+t\bm{z}), \bm{z} \right \rangle dt \\
& \leq \int_0^1 \left \langle \nabla f(\bm{x}), \bm{z} \right \rangle dt
+ \left | \int_0^1 \left \langle \nabla f(\bm{x}+t\bm{z})- \nabla f(\bm{x}), \bm{z} \right \rangle dt \right | \\
& \leq \int_0^1 \left \langle \nabla f(\bm{x}), \bm{z} \right \rangle dt
+ \int_0^1 \lVert \nabla f(\bm{x}+t\bm{z})- \nabla f(\bm{x}) \rVert_2 \cdot \lVert \bm{z} \rVert_2 dt \\
& = \left \langle \nabla f(\bm{x}), \bm{z} \right \rangle
+ \lVert \bm{z} \rVert \int_0^1 \lVert \nabla f(\bm{x}+t\bm{z})- \nabla f(\bm{x}) \rVert_2  dt \\
& \leq \left \langle \nabla f(\bm{x}), \bm{z} \right \rangle
+ \lVert \bm{z} \rVert_2 \int_0^1 L t \lVert \bm{z} \rVert_2 dt \\
& = \left \langle \nabla f(\bm{x}), \bm{z} \right \rangle
+ L \lVert \bm{z} \rVert^2_2 \int_0^1 t  dt \\
& = \left \langle \nabla f(\bm{x}), \bm{z} \right \rangle
+ \frac{L}{2} \lVert \bm{z} \rVert^2_2.  \\
\label{eq:descent_lemma_mid}
\end{split}
\end{align}

Let $\bm{y} = \bm{x} + \bm{z} $, Eq.~(\ref{eq:descent_lemma_mid}) can be written as 

\begin{align}
\begin{split}
f(\bm{y}) - f(\bm{x})
= \left \langle \nabla f(\bm{x}), \bm{y} - \bm{x} \right \rangle
+ \frac{L}{2} \lVert \bm{y} - \bm{x} \rVert_2^2.  \\
\end{split}
\end{align}

Q.E.D. $\blacksquare$

\noindent
\textbf{Theorem 1.} Suppose the ISDA loss function  $\ell^{\text{ISDA}}$ and the cross-entropy loss function $\ell^{\text{CE}}$ are both differentiable, Lipschitz continuous with constant $L$ and have $\rho$-bounded gradients with respect to the training/meta data. The learning rate satisfies $a_t = \min\{1, \frac{k}{T}\}$, for some $k>0$, such that $\frac{k}{T} < 1$. The meta learning rate $\beta_t (1<t<N)$ is monotone descent sequence. $\beta_t = \min \{\frac{1}{L}, \frac{c}{\sigma\sqrt{T}}\}$ for some $c>0$, such that $\frac{\sigma\sqrt{T}}{c} \geq L$ and $\sum_{t=1}^{\infty} \beta_t \leq \infty$, $\sum_{t=1}^{\infty} \beta_t^2 \leq \infty$. Then  

\noindent 1) the proposed algorithm can always achieve
\begin{align}
\begin{split}
\min _{0 \leq t \leq T} \mathbb{E}\left[\left\|\nabla \mathcal{L}^{\text{meta}}\left(\bm{\theta}_{g}^{(t)}\right)\right\|_{2}^{2}\right] \leq \mathcal{O}\left(\frac{1}{\sqrt{T}}\right)
\end{split}
\end{align}
\noindent
in $T$ steps;

\noindent
2) the training loss is convergent
\begin{align}
\begin{split}
\lim_{t \to \infty} \mathop{\mathbb{E}} \left [ \| \nabla \mathcal{L}^{\text{train}}(\bm{\theta}_{f}^{(t)};\bm{\theta}_{g}^{(t)})\|_2^2 \right ] = 0.
\end{split}
\end{align}

\noindent \emph{Proof.}
The update process of $\bm{\theta}_g$ in each iteration is 

\begin{equation}
\bm{\theta}_{g}^{(t+1)}=\bm{\theta}_{g}^{(t)} - \left.\beta_t \nabla_{\bm{\theta}_{g}} \mathcal{L}^{\text{meta}}\left(\tilde{\bm{\theta}}_{f}^{(t)}(\bm{\theta}_{g})\right)\right|_{\bm{\theta}_{g}^{(t)}}.
\end{equation}

\noindent 
This can be written as

\begin{equation}
\bm{\theta}_{g}^{(t+1)}=\bm{\theta}_{g}^{(t)}-\left.\beta_t \nabla_{\bm{\theta}_{g}} \mathcal{L}^{\text{meta}}\left(\tilde{\bm{\theta}}_{f}^{(t)}(\bm{\theta}_{g})\right)\right|_{\bm{\Xi}_{t}},
\end{equation}

\noindent
where $\bm{\Xi}_{t}$ is a mini-batch of metadata. Since $\bm{\Xi}_{t}$ is drawn uniformly from the entire data set, we can rewrite the update equation as

\begin{equation}
\bm{\theta}_{g}^{(t+1)}=\bm{\theta}_{g}^{(t)} - \beta_t [ \nabla_{\bm{\theta}_{g}} \mathcal{L}^{\text{meta}}\left(\tilde{\bm{\theta}}_{f}^{(t)}(\bm{\theta}_{g})\right) + \bm{\xi}^{(t)}],
\end{equation}

\noindent
where $\bm{\xi}^{(t)} = \left. \nabla_{\bm{\theta}_{g}} \mathcal{L}^{\text{meta}}\left(\tilde{\bm{\theta}}_{f}^{(t)}(\bm{\theta}_{g})\right)\right|_{\bm{\Xi}_{t}}  -  \nabla_{\bm{\theta}_{g}} \mathcal{L}^{\text{meta}}\left(\tilde{\bm{\theta}}_{f}^{(t)}(\bm{\theta}_{g})\right)$ are i.i.d random variable with finite variance $\sigma^2$, since $\bm{\Xi}_{t}$ are drawn i.i.d with a finite number of samples. Furthermore, $\mathop{\mathbb{E}}[\bm{\xi}^{(t)}]=\bm{0}$ because samples are drawn uniformly at random. We can split the difference of meta loss between two adjacent time steps as

\begin{equation}
\label{appeq:two_term_meta}
\begin{aligned}
& \mathcal{L}^{\text{meta}}\left(\tilde{\bm{\theta}}_{f}^{(t+1)}(\bm{\theta}_{g}^{(t+1)})\right) - \mathcal{L}^{\text{meta}}\left(\tilde{\bm{\theta}}_{f}^{(t)}(\bm{\theta}_{g}^{(t)})\right) \\
= \{ & \mathcal{L}^{\text{meta}}\left(\tilde{\bm{\theta}}_{f}^{(t+1)}(\bm{\theta}_{g}^{(t+1)})\right) - \mathcal{L}^{\text{meta}}\left(\tilde{\bm{\theta}}_{f}^{(t)}(\bm{\theta}_{g}^{(t+1)})\right) \}
+ \{ \mathcal{L}^{\text{meta}}\left(\tilde{\bm{\theta}}_{f}^{(t)}(\bm{\theta}_{g}^{(t+1)})\right) - \mathcal{L}^{\text{meta}}\left(\tilde{\bm{\theta}}_{f}^{(t)}(\bm{\theta}_{g}^{(t)})\right) \}
\end{aligned}
\end{equation}

\noindent
For the first term in Eq.~\eqref{appeq:two_term_meta}, by Lemma 2 we have

\begin{equation}
\label{appeq:two_term_meta_descent_lemma_1}
\begin{aligned}
& \mathcal{L}^{\text{meta}}\left(\tilde{\bm{\theta}}_{f}^{(t+1)}(\bm{\theta}_{g}^{(t+1)})\right) - \mathcal{L}^{\text{meta}}\left(\tilde{\bm{\theta}}_{f}^{(t)}(\bm{\theta}_{g}^{(t+1)})\right) \\
\leq & \left \langle \nabla_{\bm{\theta}_{g}} \mathcal{L}^{\text{meta}}\left(\tilde{\bm{\theta}}_{f}^{(t)}(\bm{\theta}_{g}^{(t+1)})\right) , \tilde{\bm{\theta}}_{f}^{(t+1)}(\bm{\theta}_{g}^{(t+1)}) - \tilde{\bm{\theta}}_{f}^{(t)}(\bm{\theta}_{g}^{(t+1)})) \right \rangle + \frac{L}{2} \left \| \tilde{\bm{\theta}}_{f}^{(t+1)}(\bm{\theta}_{g}^{(t+1)}) - \tilde{\bm{\theta}}_{f}^{(t)}(\bm{\theta}_{g}^{(t+1)})) \right \|^2_2.
\end{aligned}
\end{equation}

\noindent
Since

\begin{equation}
\begin{aligned}
& \tilde{\bm{\theta}}_{f}^{(t+1)}(\bm{\theta}_{g}^{(t+1)}) - \tilde{\bm{\theta}}_{f}^{(t)}(\bm{\theta}_{g}^{(t+1)}) \\
= & \bm{\theta}_{f}^{(t+1)} - \left.\alpha_{t+1} \nabla_{\bm{\theta}_{f}} \mathcal{L}^{\text{train}}(\bm{\theta}_{f}; \bm{\theta}_{g}^{(t+1)})\right|_{\bm{\theta}_{f}^{(t+1)}}- \bm{\theta}_{f}^{(t)}  + \left.\alpha_t \nabla_{\bm{\theta}_{f}} \mathcal{L}^{\text{train}}(\bm{\theta}_{f}; \bm{\theta}_{g}^{(t+1)})\right|_{\bm{\theta}_{f}^{(t)}} \\
= & \bm{\theta}_{f}^{(t+1)} - \bm{\theta}_{f}^{(t)} - \left.\alpha_{t+1} \nabla_{\bm{\theta}_{f}} \mathcal{L}^{\text{train}}(\bm{\theta}_{f}; \bm{\theta}_{g}^{(t+1)})\right|_{\bm{\theta}_{f}^{(t+1)}}  + \left.\alpha_t \nabla_{\bm{\theta}_{f}} \mathcal{L}^{\text{train}}(\bm{\theta}_{f}; \bm{\theta}_{g}^{(t+1)})\right|_{\bm{\theta}_{f}^{(t)}} \\
= & -\left.\alpha_t \nabla_{\bm{\theta}_{f}} \mathcal{L}^{\text{train}}(\bm{\theta}_{f}; \bm{\theta}_{g}^{(t+1)})\right|_{\bm{\theta}_{f}^{(t)}} - \left.\alpha_{t+1} \nabla_{\bm{\theta}_{f}} \mathcal{L}^{\text{train}}(\bm{\theta}_{f}; \bm{\theta}_{g}^{(t+1)})\right|_{\bm{\theta}_{f}^{(t+1)}}  + \left.\alpha_t \nabla_{\bm{\theta}_{f}} \mathcal{L}^{\text{train}}(\bm{\theta}_{f}; \bm{\theta}_{g}^{(t+1)})\right|_{\bm{\theta}_{f}^{(t)}} \\
= &  - \left.\alpha_{t+1} \nabla_{\bm{\theta}_{f}} \mathcal{L}^{\text{train}}(\bm{\theta}_{f}; \bm{\theta}_{g}^{(t+1)})\right|_{\bm{\theta}_{f}^{(t+1)}} \\
= &  - \left.\alpha_{t+1} \frac{1}{N_1} \sum_{i=1}^{N_1} \nabla_{\bm{\theta}_{f}} \ell^{\text{ISDA}}(\bm{\theta}_{f}; \bm{\theta}_{g}^{(t+1)})\right|_{\bm{\theta}_{f}^{(t+1)}}, \\
\end{aligned}
\end{equation}

\noindent
we have

\begin{equation}
\centering
\begin{aligned}
& \left \| \tilde{\bm{\theta}}_{f}^{(t+1)}(\bm{\theta}_{g}^{(t+1)}) - \tilde{\bm{\theta}}_{f}^{(t)}(\bm{\theta}_{g}^{(t+1)}) \right \|_2
= \alpha_{t+1}  \frac{1}{N_1} \left \| \sum_{i=1}^{N_1} \left. \nabla_{\bm{\theta}_{f}} \ell^{\text{ISDA}}(\bm{\theta}_{f}; \bm{\theta}_{g}^{(t+1)})\right|_{\bm{\theta}_{f}^{(t+1)}} \right \|_2 \\
\leq & \alpha_{t+1}  \frac{1}{N_1} \sum_{i=1}^{N_1} \left \| \left. \nabla_{\bm{\theta}_{f}} \ell^{\text{ISDA}}(\bm{\theta}_{f}; \bm{\theta}_{g}^{(t+1)})\right|_{\bm{\theta}_{f}^{(t+1)}} \right \|_2
\leq \alpha_{t+1}  \frac{1}{N_1} \sum_{i=1}^{N_1} \rho
= \alpha_{t+1} \rho
\leq \alpha_{t} \rho.
\end{aligned}
\end{equation}
 
\noindent
Therefore, Eq.~\eqref{appeq:two_term_meta_descent_lemma_1} satisfies
 
\begin{equation}
\label{appeq:two_term_meta_result_1}
\centering
\begin{aligned}
& \mathcal{L}^{\text{meta}}\left(\tilde{\bm{\theta}}_{f}^{(t+1)}(\bm{\theta}_{g}^{(t+1)})\right) - \mathcal{L}^{\text{meta}}\left(\tilde{\bm{\theta}}_{f}^{(t)}(\bm{\theta}_{g}^{(t+1)})\right) \\
\leq & \left \| \mathcal{L}^{\text{meta}}\left(\tilde{\bm{\theta}}_{f}^{(t+1)}(\bm{\theta}_{g}^{(t+1)})\right) - \mathcal{L}^{\text{meta}}\left(\tilde{\bm{\theta}}_{f}^{(t)}(\bm{\theta}_{g}^{(t+1)})\right) \right \|_2 \\
\leq & \left \| \left \langle \nabla_{\bm{\theta}_{g}} \mathcal{L}^{\text{meta}}\left(\tilde{\bm{\theta}}_{f}^{(t)}(\bm{\theta}_{g}^{(t+1)})\right) , \tilde{\bm{\theta}}_{f}^{(t+1)}(\bm{\theta}_{g}^{(t+1)}) - \tilde{\bm{\theta}}_{f}^{(t)}(\bm{\theta}_{g}^{(t+1)})) \right \rangle + \frac{L}{2} \left \| \tilde{\bm{\theta}}_{f}^{(t+1)}(\bm{\theta}_{g}^{(t+1)}) - \tilde{\bm{\theta}}_{f}^{(t)}(\bm{\theta}_{g}^{(t+1)})) \right \|^2_2 \right \|_2 \\
\leq & \left \| \left \langle \nabla_{\bm{\theta}_{g}} \mathcal{L}^{\text{meta}}\left(\tilde{\bm{\theta}}_{f}^{(t)}(\bm{\theta}_{g}^{(t+1)})\right) , \tilde{\bm{\theta}}_{f}^{(t+1)}(\bm{\theta}_{g}^{(t+1)}) - \tilde{\bm{\theta}}_{f}^{(t)}(\bm{\theta}_{g}^{(t+1)})) \right \rangle \right \|_2 +  \left\| \frac{L}{2} \left \| \tilde{\bm{\theta}}_{f}^{(t+1)}(\bm{\theta}_{g}^{(t+1)}) - \tilde{\bm{\theta}}_{f}^{(t)}(\bm{\theta}_{g}^{(t+1)})) \right \|^2_2 \right \|_2 \\
\leq & \left \| \nabla_{\bm{\theta}_{g}} \mathcal{L}^{\text{meta}}\left(\tilde{\bm{\theta}}_{f}^{(t)}(\bm{\theta}_{g}^{(t+1)})\right) \right \|_2 \left \| \tilde{\bm{\theta}}_{f}^{(t+1)}(\bm{\theta}_{g}^{(t+1)}) - \tilde{\bm{\theta}}_{f}^{(t)}(\bm{\theta}_{g}^{(t+1)})) \right \|_2 +  \frac{L}{2} \left\| \tilde{\bm{\theta}}_{f}^{(t+1)}(\bm{\theta}_{g}^{(t+1)}) - \tilde{\bm{\theta}}_{f}^{(t)}(\bm{\theta}_{g}^{(t+1)})) \right \|^2_2 \\ 
\leq & \alpha_t \rho^2 +  \frac{L}{2} \alpha_t^2 \rho^2. \\ 
\end{aligned}
\end{equation}

\noindent
For the second term in Eq.~\eqref{appeq:two_term_meta}, by Lemma 2 we can also have

\begin{equation}
\label{appeq:two_term_meta_result_2}
\centering
\begin{aligned}
&\mathcal{L}^{\text{meta}}\left(\tilde{\bm{\theta}}_{f}^{(t)}(\bm{\theta}_{g}^{(t+1)})\right) - \mathcal{L}^{\text{meta}}\left(\tilde{\bm{\theta}}_{f}^{(t)}(\bm{\theta}_{g}^{(t)})\right) \\
\leq & \left \langle \nabla_{\bm{\theta}_{g}} \mathcal{L}^{\text{meta}}\left(\tilde{\bm{\theta}}_{f}^{(t)}(\bm{\theta}_{g}^{(t)})\right),  \bm{\theta}_{g}^{(t+1)} - \bm{\theta}_{g}^{(t)} \right \rangle  + \frac{L}{2} \left \| \bm{\theta}_{g}^{(t+1)} - \bm{\theta}_{g}^{(t)} \right \| ^2_2 \\
\leq & \left \langle \nabla_{\bm{\theta}_{g}} \mathcal{L}^{\text{meta}}\left(\tilde{\bm{\theta}}_{f}^{(t)}(\bm{\theta}_{g}^{(t)})\right),  -\beta_t [ \nabla_{\bm{\theta}_{g}} \mathcal{L}^{\text{meta}}\left(\tilde{\bm{\theta}}_{f}^{(t)}(\bm{\theta}_{g})\right) + \bm{\xi}^{(t)}] \right \rangle  + \frac{L}{2} \left \| -\beta_t [ \nabla_{\bm{\theta}_{g}} \mathcal{L}^{\text{meta}}\left(\tilde{\bm{\theta}}_{f}^{(t)}(\bm{\theta}_{g})\right) + \bm{\xi}^{(t)}] \right \| ^2_2 \\
= & -(\beta_{t}-\frac{L \beta_{t}^{2}}{2})\left\| \nabla_{\bm{\theta}_{g}} \mathcal{L}^{\text{meta}}\left(\tilde{\bm{\theta}}_{f}^{(t)}(\bm{\theta}_{g}^{(t)})\right) \right\|_{2}^{2}+\frac{L \beta_{t}^{2}}{2} \left\| \bm{\xi}^{(t)} \right\|_{2}^{2}-\left(\beta_{t}-L \beta_{t}^{2}\right)\left\langle \nabla_{\bm{\theta}_{g}} \mathcal{L}^{\text{meta}}\left(\tilde{\bm{\theta}}_{f}^{(t)}(\bm{\theta}_{g}^{(t)})\right), \bm{\xi}^{(t)} \right\rangle.
\end{aligned}
\end{equation}

\noindent
Adding Eq.~\eqref{appeq:two_term_meta_result_1} and Eq.~\eqref{appeq:two_term_meta_result_2} together, we get 

\begin{equation}
\label{appeq:add_together}
\centering
\begin{aligned}
& \mathcal{L}^{\text{meta}}\left(\tilde{\bm{\theta}}_{f}^{(t+1)}(\bm{\theta}_{g}^{(t+1)})\right) - \mathcal{L}^{\text{meta}}\left(\tilde{\bm{\theta}}_{f}^{(t)}(\bm{\theta}_{g}^{(t)})\right) \\
\leq & \alpha_t \rho^2 +  \frac{L}{2} \alpha_t^2 \rho^2 -(\beta_{t}-\frac{L \beta_{t}^{2}}{2})\left\| \nabla_{\bm{\theta}_{g}} \mathcal{L}^{\text{meta}}\left(\tilde{\bm{\theta}}_{f}^{(t)}(\bm{\theta}_{g}^{(t)})\right) \right\|_{2}^{2}+\frac{L \beta_{t}^{2}}{2} \left\| \bm{\xi}^{(t)} \right\|_{2}^{2}-\left(\beta_{t}-L \beta_{t}^{2}\right)\left\langle \nabla_{\bm{\theta}_{g}} \mathcal{L}^{\text{meta}}\left(\tilde{\bm{\theta}}_{f}^{(t)}(\bm{\theta}_{g}^{(t)})\right), \bm{\xi}^{(t)} \right\rangle.
\end{aligned}
\end{equation}

\noindent
Rearranging the terms in Eq.~\eqref{appeq:add_together}, we can obtain

\begin{equation}
\label{appeq:rearrange}
\centering
\begin{aligned}
& (\beta_{t}-\frac{L \beta_{t}^{2}}{2})\left\| \nabla_{\bm{\theta}_{g}} \mathcal{L}^{\text{meta}}\left(\tilde{\bm{\theta}}_{f}^{(t)}(\bm{\theta}_{g}^{(t)})\right) \right\|_{2}^{2} \\
\leq & \alpha_t \rho^2 + \frac{L}{2} \alpha_t^2 \rho^2 - \mathcal{L}^{\text{meta}}\left(\tilde{\bm{\theta}}_{f}^{(t+1)}(\bm{\theta}_{g}^{(t+1)})\right) + \mathcal{L}^{\text{meta}}\left(\tilde{\bm{\theta}}_{f}^{(t)}(\bm{\theta}_{g}^{(t)})\right)  +\frac{L \beta_{t}^{2}}{2} \left\| \bm{\xi}^{(t)} \right\|_{2}^{2}-\left(\beta_{t}-L \beta_{t}^{2}\right) \left \langle \nabla_{\bm{\theta}_{g}} \mathcal{L}^{\text{meta}}\left(\tilde{\bm{\theta}}_{f}^{(t)}(\bm{\theta}_{g}^{(t)})\right), \bm{\xi}^{(t)} \right\rangle.
\end{aligned}
\end{equation}

\noindent
Summing up the Eq.~\eqref{appeq:rearrange} over $T$ time steps, we get

\begin{equation}
\centering
\begin{aligned}
& \sum_{t=1}^{T} \left \{ (\beta_{t}-\frac{L \beta_{t}^{2}}{2})\left\| \nabla_{\bm{\theta}_{g}} \mathcal{L}^{\text{meta}}\left(\tilde{\bm{\theta}}_{f}^{(t)}(\bm{\theta}_{g}^{(t)})\right) \right\|_{2}^{2} \right \} \\
\leq & \sum_{t=1}^{T} \left \{\alpha_t \rho^2 + \frac{L}{2} \alpha_t^2 \rho^2 \right \}
+ \sum_{t=1}^{T} \left \{ - \mathcal{L}^{\text{meta}}\left(\tilde{\bm{\theta}}_{f}^{(t+1)}(\bm{\theta}_{g}^{(t+1)})\right) + \mathcal{L}^{\text{meta}}\left(\tilde{\bm{\theta}}_{f}^{(t)}(\bm{\theta}_{g}^{(t)})\right)   \right \}  \\
& + \sum_{t=1}^{T} \left \{ \frac{L \beta_{t}^{2}}{2}\left\| \bm{\xi}^{(t)}\right\|_{2}^{2} \right \} 
- \sum_{t=1}^{T} \left \{ \left(\beta_{t}-L \beta_{t}^{2}\right) \left \langle \nabla_{\bm{\theta}_{g}} \mathcal{L}^{\text{meta}}\left(\tilde{\bm{\theta}}_{f}^{(t)}(\bm{\theta}_{g}^{(t)})\right), \bm{\xi}^{(t)} \right\rangle \right \} \\
\leq & \sum_{t=1}^{T} \left \{\alpha_t \rho^2 + \frac{L}{2} \alpha_t^2 \rho^2 \right \}
+ \left \{ - \mathcal{L}^{\text{meta}}\left(\tilde{\bm{\theta}}_{f}^{(T+1)}(\bm{\theta}_{g}^{(T+1)})\right) + \mathcal{L}^{\text{meta}}\left(\tilde{\bm{\theta}}_{f}^{(1)}(\bm{\theta}_{g}^{(1)})\right)   \right \}  \\
& + \sum_{t=1}^{T} \left \{ \frac{L \beta_{t}^{2}}{2}\left\| \bm{\xi}^{(t)}\right\|_{2}^{2} \right \} 
- \sum_{t=1}^{T} \left \{ \left(\beta_{t}-L \beta_{t}^{2}\right) \left \langle \nabla_{\bm{\theta}_{g}} \mathcal{L}^{\text{meta}}\left(\tilde{\bm{\theta}}_{f}^{(t)}(\bm{\theta}_{g}^{(t)})\right), \bm{\xi}^{(t)} \right\rangle \right \} \\
\leq & \sum_{t=1}^{T} \left \{\alpha_t \rho^2 + \frac{L}{2} \alpha_t^2 \rho^2 \right \}
+ \mathcal{L}^{\text{meta}}\left(\tilde{\bm{\theta}}_{f}^{(1)}(\bm{\theta}_{g}^{(1)})\right)   \\
& + \sum_{t=1}^{T} \left \{ \frac{L \beta_{t}^{2}}{2}\left\| \bm{\xi}^{(t)}\right\|_{2}^{2} \right \} 
- \sum_{t=1}^{T} \left \{ \left(\beta_{t}-L \beta_{t}^{2}\right) \left \langle \nabla_{\bm{\theta}_{g}} \mathcal{L}^{\text{meta}}\left(\tilde{\bm{\theta}}_{f}^{(t)}(\bm{\theta}_{g}^{(t)})\right), \bm{\xi}^{(t)} \right\rangle \right \}.
\end{aligned}
\end{equation}

\noindent
That is

\begin{equation}
\label{appeq:thatis}
\centering
\begin{aligned}
& \sum_{t=1}^{T} \left \{ (\beta_{t}-\frac{L \beta_{t}^{2}}{2})\left\| \nabla_{\bm{\theta}_{g}} \mathcal{L}^{\text{meta}}\left(\tilde{\bm{\theta}}_{f}^{(t)}(\bm{\theta}_{g}^{(t)})\right) \right\|_{2}^{2} \right \} \\
\leq & \sum_{t=1}^{T} \left \{ \alpha_t \rho^2 + \frac{L}{2} \alpha_t^2 \rho^2 \right \}
+ \mathcal{L}^{\text{meta}}\left(\tilde{\bm{\theta}}_{f}^{(1)}(\bm{\theta}_{g}^{(1)})\right) + \sum_{t=1}^{T} \left \{ \frac{L \beta_{t}^{2}}{2}\left\| \bm{\xi}^{(t)}\right\|_{2}^{2} \right \} 
- \sum_{t=1}^{T} \left \{ \left(\beta_{t}-L \beta_{t}^{2}\right) \left \langle \nabla_{\bm{\theta}_{g}} \mathcal{L}^{\text{meta}}\left(\tilde{\bm{\theta}}_{f}^{(t)}(\bm{\theta}_{g}^{(t)})\right), \bm{\xi}^{(t)} \right \rangle \right \}.
\end{aligned}
\end{equation}

\noindent
Taking expectations with respect to $\bm{\xi}^{(t)}$ on both sides of Eq.~\eqref{appeq:thatis}, we can then obtain

\begin{equation}
\centering
\begin{aligned}
& \sum_{t=1}^{T} \left \{ (\beta_{t}-\frac{L \beta_{t}^{2}}{2}) \mathop{\mathbb{E}_{\bm{\xi}^{(t)}}} \left\| \nabla_{\bm{\theta}_{g}} \mathcal{L}^{\text{meta}}\left(\tilde{\bm{\theta}}_{f}^{(t)}(\bm{\theta}_{g}^{(t)})\right) \right\|_{2}^{2} \right \} \\
\leq & \sum_{t=1}^{T} \left \{ \alpha_t \rho^2 + \frac{L}{2} \alpha_t^2 \rho^2 \right \} 
+ \mathcal{L}^{\text{meta}}\left(\tilde{\bm{\theta}}_{f}^{(1)}(\bm{\theta}_{g}^{(1)})\right)
+ \sum_{t=1}^{T} \left \{ \frac{L \beta_{t}^{2}}{2} \mathop{\mathbb{E}_{\bm{\xi}^{(t)}}} \left\| \bm{\xi}^{(t)}\right\|_{2}^{2} \right \} \\
& - \sum_{t=1}^{T}  \mathop{\mathbb{E}_{\bm{\xi}^{(t)}}} \left \{ \left(\beta_{t}-L \beta_{t}^{2}\right) \left \langle \nabla_{\bm{\theta}_{g}} \mathcal{L}^{\text{meta}}\left(\tilde{\bm{\theta}}_{f}^{(t)}(\bm{\theta}_{g}^{(t)})\right), \bm{\xi}^{(t)} \right \rangle \right \} \\
= & \sum_{t=1}^{T} \left \{ \alpha_t \rho^2 + \frac{L}{2} \alpha_t^2 \rho^2 \right \} 
+ \mathcal{L}^{\text{meta}}\left(\tilde{\bm{\theta}}_{f}^{(1)}(\bm{\theta}_{g}^{(1)})\right)
+ \sum_{t=1}^{T} \left \{ \frac{L \beta_{t}^{2}}{2} \mathop{\mathbb{E}_{\bm{\xi}^{(t)}}} \left\| \bm{\xi}^{(t)}\right\|_{2}^{2} \right \} \\
\leq & \sum_{t=1}^{T} \left \{ \alpha_t \rho^2 + \frac{L}{2} \alpha_t^2 \rho^2 \right \} 
+ \mathcal{L}^{\text{meta}}\left(\tilde{\bm{\theta}}_{f}^{(1)}(\bm{\theta}_{g}^{(1)})\right)
+ \frac{L \sigma^2 }{2} \sum_{t=1}^{T} \left \{ \beta_{t}^{2} \right \}.
\end{aligned}
\end{equation}

\noindent
Take a step further, we have

\begin{equation}
\centering
\begin{aligned}
& \min_t \mathop{\mathbb{E}} \left [ \left\| \nabla_{\bm{\theta}_{g}} \mathcal{L}^{\text{meta}}\left(\tilde{\bm{\theta}}_{f}^{(t)}(\bm{\theta}_{g}^{(t)})\right) \right\|_{2}^{2} \right] \sum_{t=1}^{T} \left \{ (\beta_{t}-\frac{L \beta_{t}^{2}}{2})   \right \} \\
\leq & 
\sum_{t=1}^{T} \left \{ (\beta_{t}-\frac{L \beta_{t}^{2}}{2}) \mathop{\mathbb{E}_{\bm{\xi}^{(t)}}} \left\| \nabla_{\bm{\theta}_{g}} \mathcal{L}^{\text{meta}}\left(\tilde{\bm{\theta}}_{f}^{(t)}(\bm{\theta}_{g}^{(t)})\right) \right\|_{2}^{2} \right \}  \\
\leq & 
\sum_{t=1}^{T} \left \{ \alpha_t \rho^2 + \frac{L}{2} \alpha_t^2 \rho^2 \right \} 
+ \mathcal{L}^{\text{meta}}\left(\tilde{\bm{\theta}}_{f}^{(1)}(\bm{\theta}_{g}^{(1)})\right)
+ \frac{L \sigma^2 }{2} \sum_{t=1}^{T} \left \{ \beta_{t}^{2} \right \}.
\end{aligned}
\end{equation}

\noindent
Move the term $\sum_{t=1}^{T} \left \{ (\beta_{t}-\frac{L \beta_{t}^{2}}{2})   \right \}$ into the denominate, we have

\begin{equation}
\label{appeq:three_terms}
\centering
\begin{aligned}
& \min_t \mathop{\mathbb{E}} \left [ \left\| \nabla_{\bm{\theta}_{g}} \mathcal{L}^{\text{meta}}\left(\tilde{\bm{\theta}}_{f}^{(t)}(\bm{\theta}_{g}^{(t)})\right) \right\|_{2}^{2} \right] \\
\leq
& \frac{
\sum_{t=1}^{T} \left \{ \alpha_t \rho^2 + \frac{L}{2} \alpha_t^2 \rho^2 \right \} 
+ \mathcal{L}^{\text{meta}}\left(\tilde{\bm{\theta}}_{f}^{(1)}(\bm{\theta}_{g}^{(1)})\right)
+ \frac{L \sigma^2 }{2} \sum_{t=1}^{T} \left \{ \beta_{t}^{2} \right \}
}{
\sum_{t=1}^{T} \left \{ (\beta_{t}-\frac{L \beta_{t}^{2}}{2}) \right \}
} \\
\leq
& \frac{
\sum_{t=1}^{T} \left \{ 2 \alpha_t \rho^2 + L \alpha_t^2 \rho^2 \right \} 
+ 2 \mathcal{L}^{\text{meta}}\left(\tilde{\bm{\theta}}_{f}^{(1)}(\bm{\theta}_{g}^{(1)})\right)
+ L \sigma^2 \sum_{t=1}^{T} \left \{ \beta_{t}^{2} \right \}
}{
\sum_{t=1}^{T} \left \{ ( 2 \beta_{t} - L \beta_{t}^{2}) \right \}
} \\
\leq
& \frac{
\sum_{t=1}^{T} \left \{ 2 \alpha_t \rho^2 + L \alpha_t^2 \rho^2 \right \} 
+ 2 \mathcal{L}^{\text{meta}}\left(\tilde{\bm{\theta}}_{f}^{(1)}(\bm{\theta}_{g}^{(1)})\right)
+ L \sigma^2 \sum_{t=1}^{T} \left \{ \beta_{t}^{2} \right \}
}{
\sum_{t=1}^{T} \left \{ \beta_{t} \right \}
} \\
\leq
& \frac{
2T \alpha_1 \rho^2 +  L T\alpha_1^2 \rho^2 
+ 2 \mathcal{L}^{\text{meta}}\left(\tilde{\bm{\theta}}_{f}^{(1)}(\bm{\theta}_{g}^{(1)})\right)
+ L \sigma^2 \sum_{t=1}^{T} \left \{ \beta_{t}^{2} \right \}
}{
\sum_{t=1}^{T} \left \{ \beta_{t}   \right \}
} \\
& =
\frac{2T\alpha_1 \rho^2 + LT\alpha_1^2 \rho^2}{
\sum_{t=1}^{T} \left \{ \beta_{t}   \right \}
}
+ \frac{2 \mathcal{L}^{\text{meta}}\left(\tilde{\bm{\theta}}_{f}^{(1)}(\bm{\theta}_{g}^{(1)})\right)}{
\sum_{t=1}^{T} \left \{ \beta_{t}   \right \}
}
+ \frac{L \sigma^2 \sum_{t=1}^{T} \left \{ \beta_{t}^2 \right \}}{
\sum_{t=1}^{T} \left \{ \beta_{t}   \right \}
}
\end{aligned}
\end{equation}

\noindent
The second inequality holds for $2\beta_{t} - L \beta_{t}^{2} \geq \beta_{t} \iff \beta_t \geq L \beta_{t}^{2} \iff \beta_t(\beta_t-\frac{1}{L}) \leq 0 \iff 0 \leq \beta_t \leq \frac{1}{L}$.
We deal with the three terms in Eq.~\eqref{appeq:three_terms} separately. For the first term,

\begin{equation}
\centering
\begin{aligned}
\frac{2T\alpha_1 \rho^2 + LT\alpha_1^2 \rho^2}{\sum_{t=1}^{T} \beta_{t} }
\leq \frac{2T\alpha_1 \rho^2 + LT\alpha_1^2 \rho^2}{T \beta_{T} }
=\frac{2\alpha_1 \rho^2 + L\alpha_1^2 \rho^2}{\beta_{T} }
= O(\frac{1}{\sqrt{T}})
\end{aligned}
\end{equation}

\noindent
For the second term,

\begin{equation}
\centering
\begin{aligned}
\frac{2 \mathcal{L}^{\text{meta}}\left(\tilde{\bm{\theta}}_{f}^{(1)}(\bm{\theta}_{g}^{(1)})\right)}{\sum_{t=1}^{T} \beta_{t} }
=\frac{C}{\sum_{t=1}^{T} \frac{1}{\sqrt t} },
\end{aligned}
\end{equation}

\noindent
for some constant $C$. By Stolz's formula,

\begin{equation}
\centering
\begin{aligned}
\lim_{T \rightarrow \infty} \frac{\sqrt T}{\sum_{t=1}^{T} \frac{1}{\sqrt t}}
=\lim_{T \rightarrow \infty} \frac{\sqrt {T+1} - \sqrt{T} }{\frac{1}{\sqrt {T+1} }}
=\lim_{T \rightarrow \infty} \frac{1}{(\sqrt {T+1} + \sqrt{T})\frac{1}{\sqrt {T+1} }}
=\frac{1}{2},
\end{aligned}
\end{equation}

\noindent
the second term also has order $O(\frac{1}{\sqrt{T}})$. For the third term, again by Stolz's formula,

\begin{equation}
\centering
\begin{aligned}
\lim_{T \rightarrow \infty} \frac{\sum_{t=1}^{T} \left \{ \beta_{t}^2 \right \}}{\sum_{t=1}^{T} \left \{ \beta_{t}   \right \}}
= \lim_{T \rightarrow \infty} \frac{\beta_{T}^2}{\beta_{T}}
= \lim_{T \rightarrow \infty} \beta_{T} 
= \lim_{T \rightarrow \infty} \frac{1}{\sqrt{T}} 
= 0.
\end{aligned}
\end{equation}

\noindent
Therefore, the third term goes to 0 with order $O(\frac{1}{\sqrt{T}})$. In short, we can conclude that our algorithm can always achieve $\min _{0 \leq t \leq T} \mathbb{E}\left[\left\| \nabla_{\bm{\theta}_{g}} \mathcal{L}^{\text{meta}}\left(\tilde{\bm{\theta}}_{f}^{(t)}(\bm{\theta}_{g}^{(t)})\right) \right\|_{2}^{2}\right] \leq \mathcal{O}\left(\frac{1}{\sqrt{T}}\right)$
in $T$ steps.

Next, we proof the convergence of the training loss. Because the learning rate satisfies $a_t = \min\{1, \frac{k}{T}\}$, for some $k>0$, such that $\frac{k}{T} < 1$, we have $\sum_{t=0}^{\infty} \alpha_{t}=\infty$, $\sum_{t=0}^{\infty} \alpha_{t}^{2}<\infty$.

The update process of $\bm{\theta}_{f}$ 

\begin{equation}
\bm{\theta}_{f}^{(t+1)}(\bm{\theta}_{g}^{(t+1)})=\bm{\theta}_{f}^{(t)} -  \left.\alpha_t \nabla_{\bm{\theta}_{f}} \mathcal{L}^{\text{train}}(\bm{\theta}_{f}; \bm{\theta}_{g}^{(t+1)})\right|_{\bm{\theta}_{f}^{(t)}},
\end{equation}

\noindent
can be written as

\begin{equation}
\bm{\theta}_{f}^{(t+1)}(\bm{\theta}_{g}^{(t+1)})=\bm{\theta}_{f}^{(t)} - \left.\alpha_t \nabla_{\bm{\theta}_{f}} \mathcal{L}^{\text{train}}(\bm{\theta}_{f}; \bm{\theta}_{g}^{(t+1)})\right|_{\bm{\Psi}^{(t)}}.
\end{equation}

\noindent
Since the mini-batch ${\bm{\Psi}^{(t)}}$ is drawn uniformly at random, we can rewrite the update equation as

\begin{equation}
\bm{\theta}_{f}^{(t+1)}(\bm{\theta}_{g}^{(t+1)})=\bm{\theta}_{f}^{(t)} - \alpha_t \left[ \nabla_{\bm{\theta}_{f}} \mathcal{L}^{\text{train}}(\bm{\theta}_{f}^{(t)}; \bm{\theta}_{g}^{(t+1)}) + \bm{\psi}^{(t)} \right],
\end{equation}

\noindent
where $ \bm{\psi}^{(t)} = \left. \nabla_{\bm{\theta}_{f}} \mathcal{L}^{\text{train}}(\bm{\theta}_{f}; \bm{\theta}_{g}^{(t+1)})\right|_{\bm{\Psi}^{(t)}} - \nabla_{\bm{\theta}_{f}} \mathcal{L}^{\text{train}}(\bm{\theta}_{f}^{(t)}; \bm{\theta}_{g}^{(t+1)}) $. Note that $\bm{\psi}^{(t)}$ is i.i.d random variable with finite variance, since $\bm{\Psi}^{(t)}$ are drawn i.i.d. with finite number of samples. Furthermore, $\mathbb{E}\left[\bm{\psi}^{(t)}\right]=\bm{0}$, since samples are drawn uniformly at random, and $\mathbb{E}\left[\left\|\bm{\psi}^{(t)}\right\|_{2}^{2}\right] \leq \sigma^{2}$. Observe that

\begin{equation}
\label{appeq:two_term_train}
\centering
\begin{aligned}
& \mathcal{L}^{\text{train}}(\bm{\theta}_{f}^{(t+1)}; \bm{\theta}_{g}^{(t+2)}) - \mathcal{L}^{\text{train}}(\bm{\theta}_{f}^{(t)}; \bm{\theta}_{g}^{(t+1)}) \\
= \{ & \mathcal{L}^{\text{train}}(\bm{\theta}_{f}^{(t+1)}; \bm{\theta}_{g}^{(t+2)}) - \mathcal{L}^{\text{train}}(\bm{\theta}_{f}^{(t+1)} ; \bm{\theta}_{g}^{(t+1)}) \}
+ \{ \mathcal{L}^{\text{train}}(\bm{\theta}_{f}^{(t+1)} ; \bm{\theta}_{g}^{(t+1)}) - \mathcal{L}^{\text{train}}(\bm{\theta}_{f}^{(t)} ; \bm{\theta}_{g}^{(t+1)}) \}.
\end{aligned}
\end{equation}

\noindent
For the first term in Eq.~\eqref
{appeq:two_term_train}, by Lemma 2 we have

\begin{equation}
\label{appeq:two_term_train_result_1}
\centering
\begin{aligned}
&\mathcal{L}^{\text{train}}(\bm{\theta}_{f}^{(t+1)}; \bm{\theta}_{g}^{(t+2)}) - \mathcal{L}^{\text{train}}(\bm{\theta}_{f}^{(t+1)} ; \bm{\theta}_{g}^{(t+1)}) \\
&\leq 
\left \langle \left.\nabla_{\bm{\theta}_{f}}\mathcal{L}^{\text{train}}(\bm{\theta}_{f}^{(t+1)} ; \bm{\theta}_{g}) \right |_{\bm{\theta}_{g}^{(t+1)}}, \bm{\theta}_{g}^{(t+2)} - \bm{\theta}_{g}^{(t+1)} \right \rangle + \frac{L}{2} \left \| \bm{\theta}_{g}^{(t+2)} - \bm{\theta}_{g}^{(t+1)} \right \|_2^2
\\
&\leq 
\left \langle \left.\nabla_{\bm{\theta}_{f}}\mathcal{L}^{\text{train}}(\bm{\theta}_{f}^{(t)} ; \bm{\theta}_{g}) \right |_{\bm{\theta}_{g}^{(t)}}, \bm{\theta}_{g}^{(t+1)} - \bm{\theta}_{g}^{(t)} \right \rangle + \frac{L}{2} \left \| \bm{\theta}_{g}^{(t+1)} - \bm{\theta}_{g}^{(t)} \right \|_2^2
\\
&= 
\left \langle \left.\nabla_{\bm{\theta}_{f}}\mathcal{L}^{\text{train}}(\bm{\theta}_{f}^{(t)} ; \bm{\theta}_{g}) \right |_{\bm{\theta}_{g}^{(t)}}, - \beta_t [ \nabla_{\bm{\theta}_{g}} \mathcal{L}^{\text{meta}}\left(\tilde{\bm{\theta}}_{f}^{(t)}(\bm{\theta}_{g})\right) + \bm{\xi}^{(t)}] \right \rangle + \frac{L}{2} \left \|  - \beta_t [ \nabla_{\bm{\theta}_{g}} \mathcal{L}^{\text{meta}}\left(\tilde{\bm{\theta}}_{f}^{(t)}(\bm{\theta}_{g})\right) + \bm{\xi}^{(t)}] \right \|_2^2
\\
&= 
\left \langle \left.\nabla_{\bm{\theta}_{f}}\mathcal{L}^{\text{train}}(\bm{\theta}_{f}^{(t)} ; \bm{\theta}_{g}) \right |_{\bm{\theta}_{g}^{(t)}}, - \beta_t [ \nabla_{\bm{\theta}_{g}} \mathcal{L}^{\text{meta}}\left(\tilde{\bm{\theta}}_{f}^{(t)}(\bm{\theta}_{g})\right) + \bm{\xi}^{(t)}] \right \rangle + \frac{L \beta_t^2}{2} \left \|   \nabla_{\bm{\theta}_{g}} \mathcal{L}^{\text{meta}}\left(\tilde{\bm{\theta}}_{f}^{(t)}(\bm{\theta}_{g})\right) + \bm{\xi}^{(t)} \right \|_2^2
\\
&= 
\left \langle \left.\nabla_{\bm{\theta}_{f}}\mathcal{L}^{\text{train}}(\bm{\theta}_{f}^{(t)} ; \bm{\theta}_{g}) \right |_{\bm{\theta}_{g}^{(t)}}, - \beta_t [ \nabla_{\bm{\theta}_{g}} \mathcal{L}^{\text{meta}}\left(\tilde{\bm{\theta}}_{f}^{(t)}(\bm{\theta}_{g})\right) + \bm{\xi}^{(t)}] \right \rangle
+ \frac{L \beta_t^2}{2} \left \| \nabla_{\bm{\theta}_{g}} \mathcal{L}^{\text{meta}}\left(\tilde{\bm{\theta}}_{f}^{(t)}(\bm{\theta}_{g})\right) \right \| ^2
+ \frac{L \beta_t^2}{2} \left \| \bm{\xi}^{(t)} \right \| ^2 \\
& + L \beta_t^2 \left \langle \nabla_{\bm{\theta}_{g}} \mathcal{L}^{\text{meta}}\left(\tilde{\bm{\theta}}_{f}^{(t)}(\bm{\theta}_{g})\right), \bm{\xi}^{(t)} \right \rangle
\end{aligned}
\end{equation}

\noindent
The second inequality in Eq.~\eqref{appeq:two_term_train_result_1} holds for the difference of $\bm{\theta}_{g}^{(t+2)} - \bm{\theta}_{g}^{(t+1)}$ and $\bm{\theta}_{g}^{(t+1)} - \bm{\theta}_{g}^{(t)}$ is the gradient \[- \left.\beta_{t+1} \nabla_{\bm{\theta}_{g}} \mathcal{L}^{\text{meta}}\left(\tilde{\bm{\theta}}_{f}^{(t+1)}(\bm{\theta}_{g})\right)\right|_{\bm{\theta}_{g}^{(t+1)}} + \left.\beta_t \nabla_{\bm{\theta}_{g}} \mathcal{L}^{\text{meta}}\left(\tilde{\bm{\theta}}_{f}^{(t)}(\bm{\theta}_{g})\right)\right|_{\bm{\theta}_{g}^{(t)}},\] 

\noindent
which will be controlled since the gradient is bounded. Also, the difference between $\bm{\theta}_{f}^{(t+1)}$ and $\bm{\theta}_{f}^{(t)}$ is 
$ - \left.\alpha_t \nabla_{\bm{\theta}_{f}} \mathcal{L}^{\text{train}}(\bm{\theta}_{f}; \bm{\theta}_{g}^{(t+1)})\right|_{\bm{\theta}_{f}^{(t)}}$.
If we assume that $\nabla_{\bm{\theta}_{f}}\mathcal{L}^{\text{train}}$ is Lipschitz in both variable, then the difference between $ \left.\nabla_{\bm{\theta}_{f}}\mathcal{L}^{\text{train}}(\bm{\theta}_{f}^{(t+1)} ; \bm{\theta}_{g}) \right |_{\bm{\theta}_{g}^{(t+1)}}$ 
and
$ \left.\nabla_{\bm{\theta}_{f}}\mathcal{L}^{\text{train}}(\bm{\theta}_{f}^{(t)} ; \bm{\theta}_{g}) \right |_{\bm{\theta}_{g}^{(t)}}$
can be also controlled by a constant time of 
\[- \left.\alpha_t \nabla_{\bm{\theta}_{f}} \mathcal{L}^{\text{train}}(\bm{\theta}_{f}; \bm{\theta}_{g}^{(t+1)})\right|_{\bm{\theta}_{f}^{(t)}} - \left.\beta_{t+1} \nabla_{\bm{\theta}_{g}} \mathcal{L}^{\text{meta}}\left(\tilde{\bm{\theta}}_{f}^{(t+1)}(\bm{\theta}_{g})\right)\right|_{\bm{\theta}_{g}^{(t+1)}} + \left.\beta_t \nabla_{\bm{\theta}_{g}} \mathcal{L}^{\text{meta}}\left(\tilde{\bm{\theta}}_{f}^{(t)}(\bm{\theta}_{g})\right)\right|_{\bm{\theta}_{g}^{(t)}}.\] 
The assumption means that $\ell^{\text{ISDA}}$ is second-order derivable almost everywhere with bounded derivative with respect to $\bm{\theta}_g$, and is first-order derivable almost everywhere with bounded derivative with respect to $\bm{\theta}_f$.
As a result, all the cost of substituting $t+1$ by $t$ is some gradient given above. Therefore, we can do the substitution without affect the convergence.

\noindent
For the second term in Eq.~\eqref
{appeq:two_term_train}, by Lemma 2 we have

\begin{equation}
\label{appeq:two_term_train_result_2}
\begin{aligned}
&\mathcal{L}^{\text{train}}(\bm{\theta}_{f}^{(t+1)} ; \bm{\theta}_{g}^{(t+1)}) - \mathcal{L}^{\text{train}}(\bm{\theta}_{f}^{(t)} ; \bm{\theta}_{g}^{(t+1)})\\
&\leq \left \langle \nabla \mathcal{L}^{\text{train}}(\bm{\theta}_{f}^{(t)} ; \bm{\theta}_{g}^{(t+1)}), \bm{\theta}_{f}^{(t+1)} - \bm{\theta}_{f}^{(t)} \right \rangle + \frac{L}{2} \left \| \bm{\theta}_{f}^{(t+1)} - \bm{\theta}_{f}^{(t)} \right \|_2^2 \\
&= \left \langle \nabla \mathcal{L}^{\text{train}}(\bm{\theta}_{f}^{(t)} ; \bm{\theta}_{g}^{(t+1)}), - \alpha_t \left[ \nabla_{\bm{\theta}_{f}} \mathcal{L}^{\text{train}}(\bm{\theta}_{f}^{(t)}; \bm{\theta}_{g}^{(t+1)}) + \bm{\psi}^{(t)} \right] \right \rangle + \frac{L}{2} \left \|  \alpha_t \left[ \nabla_{\bm{\theta}_{f}} \mathcal{L}^{\text{train}}(\bm{\theta}_{f}^{(t)}; \bm{\theta}_{g}^{(t+1)}) + \bm{\psi}^{(t)} \right] \right \|_2^2 \\
&= \left \langle \nabla \mathcal{L}^{\text{train}}(\bm{\theta}_{f}^{(t)} ; \bm{\theta}_{g}^{(t+1)}), - \alpha_t \left[ \nabla_{\bm{\theta}_{f}} \mathcal{L}^{\text{train}}(\bm{\theta}_{f}^{(t)}; \bm{\theta}_{g}^{(t+1)}) + \bm{\psi}^{(t)} \right] \right \rangle + \frac{L \alpha_t^2}{2} \left \|    \nabla_{\bm{\theta}_{f}} \mathcal{L}^{\text{train}}(\bm{\theta}_{f}^{(t)}; \bm{\theta}_{g}^{(t+1)}) + \bm{\psi}^{(t)}  \right \|_2^2 \\
& = -(\alpha_t-\frac{L\alpha_t^2}{2}) \left \| \nabla_{\bm{\theta}_{f}} \mathcal{L}^{\text{train}}(\bm{\theta}_{f}^{(t)}; \bm{\theta}_{g}^{(t+1)}) \right \|_2^2 + \frac{L\alpha_t^2}{2} \left \| \bm{\psi}^{(t)} \right \|_2^2 - (\alpha_t-L\alpha_t^2) \left \langle \nabla_{\bm{\theta}_{f}} \mathcal{L}^{\text{train}}(\bm{\theta}_{f}^{(t)}; \bm{\theta}_{g}^{(t+1)}), \bm{\psi}^{(t)} \right \rangle.
\end{aligned}
\end{equation}

\noindent
Adding Eq.~\eqref{appeq:two_term_train_result_1} and Eq.~\eqref{appeq:two_term_train_result_2} together, we get

\begin{equation}
\label{appeq:add_together_train}
\begin{aligned}
& \mathcal{L}^{\text{train}}(\bm{\theta}_{f}^{(t+1)} ; \bm{\theta}_{g}^{(t+2)}) - \mathcal{L}^{\text{train}}(\bm{\theta}_{f}^{(t)} ; \bm{\theta}_{g}^{(t+1)})\\
\leq 
&\left \langle \left.\nabla_{\bm{\theta}_{f}}\mathcal{L}^{\text{train}}(\bm{\theta}_{f}^{(t)} ; \bm{\theta}_{g}) \right |_{\bm{\theta}_{g}^{(t)}}, - \beta_t [ \nabla_{\bm{\theta}_{g}} \mathcal{L}^{\text{meta}}\left(\tilde{\bm{\theta}}_{f}^{(t)}(\bm{\theta}_{g})\right) + \bm{\xi}^{(t)}] \right \rangle
+ \frac{L \beta_t^2}{2} \left \| \nabla_{\bm{\theta}_{g}} \mathcal{L}^{\text{meta}}\left(\tilde{\bm{\theta}}_{f}^{(t)}(\bm{\theta}_{g})\right) \right \| ^2
+ \frac{L \beta_t^2}{2} \left \| \bm{\xi}^{(t)} \right \| ^2 \\
+ & L \beta_t^2 \left \langle \nabla_{\bm{\theta}_{g}} \mathcal{L}^{\text{meta}}\left(\tilde{\bm{\theta}}_{f}^{(t)}(\bm{\theta}_{g})\right), \bm{\xi}^{(t)} \right \rangle
-(\alpha_t-\frac{L\alpha_t^2}{2}) \left \| \nabla_{\bm{\theta}_{f}} \mathcal{L}^{\text{train}}(\bm{\theta}_{f}^{(t)}; \bm{\theta}_{g}^{(t+1)}) \right \|_2^2 + \frac{L\alpha_t^2}{2} \left \| \bm{\psi}^{(t)} \right \|_2^2 \\
- & (\alpha_t-L\alpha_t^2) \left \langle \nabla_{\bm{\theta}_{f}} \mathcal{L}^{\text{train}}(\bm{\theta}_{f}^{(t)}; \bm{\theta}_{g}^{(t+1)}), \bm{\psi}^{(t)} \right \rangle.
\end{aligned}
\end{equation}

\noindent
Taking expectation of both of the random variables in turn of both sides of Eq.~\eqref{appeq:add_together_train}, we have 

\begin{equation}
\label{appeq:exp_train}
\begin{aligned}
&\mathop{\mathbb{E}}[\mathcal{L}^{\text{train}}(\bm{\theta}_{f}^{(t+1)} ; \bm{\theta}_{g}^{(t+2)})] - \mathop{\mathbb{E}}[\mathcal{L}^{\text{train}}(\bm{\theta}_{f}^{(t)} ; \bm{\theta}_{g}^{(t+1)})] \\
&\leq
\mathop{\mathbb{E}} \left [ \left \langle \left.\nabla_{\bm{\theta}_{f}}\mathcal{L}^{\text{train}}(\bm{\theta}_{f}^{(t)} ; \bm{\theta}_{g}) \right |_{\bm{\theta}_{g}^{(t)}}, - \beta_t [ \nabla_{\bm{\theta}_{g}} \mathcal{L}^{\text{meta}}\left(\tilde{\bm{\theta}}_{f}^{(t)}(\bm{\theta}_{g})\right) + \bm{\xi}^{(t)}] \right \rangle \right ]
+ \frac{L \beta_t^2}{2} \mathop{\mathbb{E}} \left [ \left \|  \nabla_{\bm{\theta}_{g}} \mathcal{L}^{\text{meta}}\left(\tilde{\bm{\theta}}_{f}^{(t)}(\bm{\theta}_{g})\right) \right \| ^2 \right ] \\
& + \frac{L \beta_t^2}{2} \mathop{\mathbb{E}} \left [ \left \|  \bm{\xi}^{(t)} \right \| ^2 \right ] 
-(\alpha_t-\frac{L\alpha_t^2}{2}) \mathop{\mathbb{E}} \left [ \left \| \nabla_{\bm{\theta}_{f}} \mathcal{L}^{\text{train}}(\bm{\theta}_{f}^{(t)}; \bm{\theta}_{g}^{(t+1)}) \right \|_2^2 \right  ]
+ \frac{L\alpha_t^2}{2} \mathop{\mathbb{E}} \left [ \left \|\bm{\psi}^{(t)} \right \|_2^2 \right ] \\
\end{aligned}
\end{equation}

\noindent
Summing up the above inequalities over $t=1, 2, ..., \infty$ in both sides of Eq.~\eqref{appeq:exp_train}, we obtain

\begin{equation}
\label{appeq:sumup_train}
\begin{aligned}
& \lim_{t \to \infty}  \mathop{\mathbb{E}}[\mathcal{L}^{\text{train}}(\bm{\theta}_{f}^{(t+1)} ; \bm{\theta}_{g}^{(t+2)})] - \mathop{\mathbb{E}}[\mathcal{L}^{\text{train}}(\bm{\theta}_{f}^{(1)} ; \bm{\theta}_{g}^{(2)})] \\
&\leq
\sum_{t=1}^{\infty} \mathop{\mathbb{E}} \left [ \left \langle \left.\nabla_{\bm{\theta}_{f}}\mathcal{L}^{\text{train}}(\bm{\theta}_{f}^{(t)} ; \bm{\theta}_{g}) \right |_{\bm{\theta}_{g}^{(t)}}, - \beta_t [ \nabla_{\bm{\theta}_{g}} \mathcal{L}^{\text{meta}}\left(\tilde{\bm{\theta}}_{f}^{(t)}(\bm{\theta}_{g})\right) + \bm{\xi}^{(t)}] \right \rangle \right ]
+ \sum_{t=1}^{\infty} \frac{L \beta_t^2}{2} \mathop{\mathbb{E}} \left [ \left \|  \nabla_{\bm{\theta}_{g}} \mathcal{L}^{\text{meta}}\left(\tilde{\bm{\theta}}_{f}^{(t)}(\bm{\theta}_{g})\right) \right \| ^2 \right ] \\
& + \sum_{t=1}^{\infty} \frac{L \beta_t^2}{2} \mathop{\mathbb{E}} \left [ \left \|  \bm{\xi}^{(t)} \right \| ^2 \right ]
- \sum_{t=1}^{\infty} (\alpha_t-\frac{L\alpha_t^2}{2}) \mathop{\mathbb{E}} \left [ \left \| \nabla_{\bm{\theta}_{f}} \mathcal{L}^{\text{train}}(\bm{\theta}_{f}^{(t)}; \bm{\theta}_{g}^{(t+1)}) \right \|_2^2 \right  ]
+ \sum_{t=1}^{\infty} \frac{L\alpha_t^2}{2} \mathop{\mathbb{E}} \left [ \left \|\bm{\psi}^{(t)} \right \|_2^2 \right ] \\
\end{aligned}
\end{equation}

\noindent
Re-arrange the terms in Eq.~\eqref{appeq:sumup_train}, we get

\begin{equation}
\label{appeq:rearranged}
\begin{aligned}
& \sum_{t=1}^{\infty} \alpha_t \mathop{\mathbb{E}} \left [ \left \| \nabla_{\bm{\theta}_{f}} \mathcal{L}^{\text{train}}(\bm{\theta}_{f}^{(t)}; \bm{\theta}_{g}^{(t+1)}) \right \|_2^2 \right ]
+ \sum_{t=1}^{\infty} \mathop{\mathbb{E}} \left [ \left \langle \left.\nabla_{\bm{\theta}_{f}}\mathcal{L}^{\text{train}}(\bm{\theta}_{f}^{(t)} ; \bm{\theta}_{g}) \right |_{\bm{\theta}_{g}^{(t)}}, \beta_t  \nabla_{\bm{\theta}_{g}} \mathcal{L}^{\text{meta}}\left(\tilde{\bm{\theta}}_{f}^{(t)}(\bm{\theta}_{g})\right) \right \rangle \right ] \\
\leq
& \mathop{\mathbb{E}}[\mathcal{L}^{\text{train}}(\bm{\theta}_{f}^{(1)} ; \bm{\theta}_{g}^{(2)})] - \lim_{t \to \infty}  \mathop{\mathbb{E}}[\mathcal{L}^{\text{train}}(\bm{\theta}_{f}^{(t+1)} ; \bm{\theta}_{g}^{(t+2)})] 
+ \sum_{t=1}^{\infty} \frac{L \beta_t^2}{2} \mathop{\mathbb{E}} \left [ \left \|  \nabla_{\bm{\theta}_{g}} \mathcal{L}^{\text{meta}}\left(\tilde{\bm{\theta}}_{f}^{(t)}(\bm{\theta}_{g})\right) \right \| ^2 \right ] \\
& + \sum_{t=1}^{\infty} \frac{L \beta_t^2}{2} \mathop{\mathbb{E}} \left [ \left \|  \bm{\xi}^{(t)} \right \| ^2 \right ] 
+ \sum_{t=1}^{\infty} (\frac{L\alpha_t^2}{2}) \mathop{\mathbb{E}} \left [ \left \| \nabla_{\bm{\theta}_{f}} \mathcal{L}^{\text{train}}(\bm{\theta}_{f}^{(t)}; \bm{\theta}_{g}^{(t+1)}) \right \|_2^2 \right  ]
+ \sum_{t=1}^{\infty} \frac{L\alpha_t^2}{2} \mathop{\mathbb{E}} \left [ \left \|\bm{\psi}^{(t)} \right \|_2^2 \right ] \\
\leq
& \mathop{\mathbb{E}}[\mathcal{L}^{\text{train}}(\bm{\theta}_{f}^{(1)} ; \bm{\theta}_{g}^{(2)})]
+ \sum_{t=1}^{\infty} \frac{L \beta_t^2}{2} \mathop{\mathbb{E}} \left [ \left \|  \nabla_{\bm{\theta}_{g}} \mathcal{L}^{\text{meta}}\left(\tilde{\bm{\theta}}_{f}^{(t)}(\bm{\theta}_{g})\right) \right \| ^2 \right ] \\
& + \sum_{t=1}^{\infty} \frac{L \beta_t^2}{2} \mathop{\mathbb{E}} \left [ \left \|  \bm{\xi}^{(t)} \right \| ^2 \right ] 
+ \sum_{t=1}^{\infty} (\frac{L\alpha_t^2}{2}) \mathop{\mathbb{E}} \left [ \left \| \nabla_{\bm{\theta}_{f}} \mathcal{L}^{\text{train}}(\bm{\theta}_{f}^{(t)}; \bm{\theta}_{g}^{(t+1)}) \right \|_2^2 \right  ]
+ \sum_{t=1}^{\infty} \frac{L\alpha_t^2}{2} \mathop{\mathbb{E}} \left [ \left \|\bm{\psi}^{(t)} \right \|_2^2 \right ] \\
\leq
& \mathop{\mathbb{E}}[\mathcal{L}^{\text{train}}(\bm{\theta}_{f}^{(1)} ; \bm{\theta}_{g}^{(2)})]
+ \sum_{t=1}^{\infty} \frac{L \beta_t^2}{2} (\rho^2 + \sigma^2)
+ \sum_{t=1}^{\infty} \frac{L\alpha_t^2}{2} (\rho^2 + \sigma^2) \\
\leq & +\infty.
\end{aligned}
\end{equation}

\noindent
The second term in Eq.~\eqref{appeq:rearranged} satisfies

\begin{equation}
\begin{aligned}
& \sum_{t=1}^{\infty} \mathop{\mathbb{E}} \left [ \left \langle \left.\nabla_{\bm{\theta}_{f}}\mathcal{L}^{\text{train}}(\bm{\theta}_{f}^{(t)} ; \bm{\theta}_{g}) \right |_{\bm{\theta}_{g}^{(t)}}, \beta_t  \nabla_{\bm{\theta}_{g}} \mathcal{L}^{\text{meta}}\left(\tilde{\bm{\theta}}_{f}^{(t)}(\bm{\theta}_{g})\right) \right \rangle \right ] \\ 
=
& \sum_{t=1}^{\infty} \beta_t  \mathop{\mathbb{E}} \left [ \left \langle \left. \nabla_{\bm{\theta}_{f}}\mathcal{L}^{\text{train}}(\bm{\theta}_{f}^{(t)} ; \bm{\theta}_{g}) \right |_{\bm{\theta}_{g}^{(t)}},  \nabla_{\bm{\theta}_{g}} \mathcal{L}^{\text{meta}}\left(\tilde{\bm{\theta}}_{f}^{(t)}(\bm{\theta}_{g})\right) \right \rangle \right ] \\ 
\leq
& \sum_{t=1}^{\infty} \beta_t \mathop{\mathbb{E}} \left [\left \|  \left. \nabla_{\bm{\theta}_{f}}\mathcal{L}^{\text{train}}(\bm{\theta}_{f}^{(t)} ; \bm{\theta}_{g}) \right |_{\bm{\theta}_{g}^{(t)}} \right \| \cdot \left \|  \nabla_{\bm{\theta}_{g}} \mathcal{L}^{\text{meta}}\left(\tilde{\bm{\theta}}_{f}^{(t)}(\bm{\theta}_{g})\right)  \right \|\right ] \\ 
\leq & \rho^2 \sum_{t=1}^{\infty} \beta_t
\leq
+\infty,
\end{aligned}
\end{equation}

\noindent
which implies that the first term in Eq.~\eqref{appeq:rearranged} also satisfies $\sum_{t=1}^{\infty} \alpha_t \mathop{\mathbb{E}} \left [ \left \| \nabla_{\bm{\theta}_{f}} \mathcal{L}^{\text{train}}(\bm{\theta}_{f}^{(t)}; \bm{\theta}_{g}^{(t+1)}) \right \|_2^2 \right ] < \infty$. By Lemma 1, to substantiate \\
$\lim_{t \to \infty} \mathop{\mathbb{E}} \left [ \left \| \nabla_{\bm{\theta}_{f}} \mathcal{L}^{\text{train}}(\bm{\theta}_{f}^{(t)}; \bm{\theta}_{g}^{(t+1)}) \right \|_2^2 \right ] = 0$, since 
$\sum_{t=0}^{\infty} \alpha_{t}=\infty$, it needs to prove

\begin{equation}
\label{appeq:need_to_prove}
\begin{aligned}
\left | 
\mathop{\mathbb{E}} \left [ \left \| \nabla_{\bm{\theta}_{f}} \mathcal{L}^{\text{train}}(\bm{\theta}_{f}^{(t+1)}; \bm{\theta}_{g}^{(t+2)}) \right \|_2^2 \right ] 
- \mathop{\mathbb{E}} \left [ \left \| \nabla_{\bm{\theta}_{f}} \mathcal{L}^{\text{train}}(\bm{\theta}_{f}^{(t)}; \bm{\theta}_{g}^{(t+1)}) \right \|_2^2 \right ] 
\right | 
\leq C \alpha_t
\end{aligned}
\end{equation}

\noindent
for some constant $C$. The left of Eq.~\eqref{appeq:need_to_prove}

\begin{equation}
\begin{aligned}
& \left | 
\mathop{\mathbb{E}} \left [ \left \| \nabla_{\bm{\theta}_{f}} \mathcal{L}^{\text{train}}(\bm{\theta}_{f}^{(t+1)}; \bm{\theta}_{g}^{(t+2)}) \right \|_2^2 \right ] 
- \mathop{\mathbb{E}} \left [ \left \| \nabla_{\bm{\theta}_{f}} \mathcal{L}^{\text{train}}(\bm{\theta}_{f}^{(t)}; \bm{\theta}_{g}^{(t+1)}) \right \|_2^2 \right ] 
\right | \\
= &
\left | 
\mathop{\mathbb{E}} \left [ \left \| \nabla_{\bm{\theta}_{f}} \mathcal{L}^{\text{train}}(\bm{\theta}_{f}^{(t+1)}; \bm{\theta}_{g}^{(t+2)}) \right \|_2 \right ]^2
- \mathop{\mathbb{E}} \left [ \left \| \nabla_{\bm{\theta}_{f}} \mathcal{L}^{\text{train}}(\bm{\theta}_{f}^{(t)}; \bm{\theta}_{g}^{(t+1)}) \right \|_2 \right ]^2 -\rho + \rho
\right | \\
= &
\left | \mathbb{E}\left[ \left( \left \| \nabla_{\bm{\theta}_{f}} \mathcal{L}^{\text{train}}(\bm{\theta}_{f}^{(t+1)}; \bm{\theta}_{g}^{(t+2)}) \right \|_2 + \left \| \nabla_{\bm{\theta}_{f}} \mathcal{L}^{\text{train}}(\bm{\theta}_{f}^{(t)}; \bm{\theta}_{g}^{(t+1)}) \right \|_2 \right ) \left ( \left \| \nabla_{\bm{\theta}_{f}} \mathcal{L}^{\text{train}}(\bm{\theta}_{f}^{(t+1)}; \bm{\theta}_{g}^{(t+2)}) \right \|_2 - \left \| \nabla_{\bm{\theta}_{f}} \mathcal{L}^{\text{train}}(\bm{\theta}_{f}^{(t)}; \bm{\theta}_{g}^{(t+1)}) \right \|_2 \right ) \right ] \right |\\
& \text{Since} \left| E[A \cdot B] \right| \leq E[\left| A \cdot \right|\left| B \right | ] \text{:} \\
\leq &
\mathbb{E}\left[ \left | \left \| \nabla_{\bm{\theta}_{f}} \mathcal{L}^{\text{train}}(\bm{\theta}_{f}^{(t+1)}; \bm{\theta}_{g}^{(t+2)}) \right \|_2 + \left \| \nabla_{\bm{\theta}_{f}} \mathcal{L}^{\text{train}}(\bm{\theta}_{f}^{(t)}; \bm{\theta}_{g}^{(t+1)}) \right \|_2 \right | \cdot \left | \left \| \nabla_{\bm{\theta}_{f}} \mathcal{L}^{\text{train}}(\bm{\theta}_{f}^{(t+1)}; \bm{\theta}_{g}^{(t+2)}) \right \|_2 - \left \| \nabla_{\bm{\theta}_{f}} \mathcal{L}^{\text{train}}(\bm{\theta}_{f}^{(t)}; \bm{\theta}_{g}^{(t+1)}) \right \|_2 \right ) \right | \\
& \text{Because} \left| (\|a\|+\|b\|)(\|a\|-\|b\|) \right| \leq\|a+b\|\|a-b\| \text{:}\\
\leq & \mathbb{E}\left[ \left \|\nabla_{\bm{\theta}_{f}} \mathcal{L}^{\text{train}}(\bm{\theta}_{f}^{(t+1)}; \bm{\theta}_{g}^{(t+2)}) + \nabla_{\bm{\theta}_{f}} \mathcal{L}^{\text{train}}(\bm{\theta}_{f}^{(t)}; \bm{\theta}_{g}^{(t+1)}) \right\|_2 \left\| \nabla_{\bm{\theta}_{f}} \mathcal{L}^{\text{train}}(\bm{\theta}_{f}^{(t+1)}; \bm{\theta}_{g}^{(t+2)}) - \nabla_{\bm{\theta}_{f}} \mathcal{L}^{\text{train}}(\bm{\theta}_{f}^{(t)}; \bm{\theta}_{g}^{(t+1)}) \right\|_2\right]\\
& \text{Because} \|a+b\|\|a-b\| \leq (\|a\|+\|b\|)\|a-b\| \text{:}\\
\leq & \mathbb{E}\left[ \left ( \left \|\nabla_{\bm{\theta}_{f}} \mathcal{L}^{\text{train}}(\bm{\theta}_{f}^{(t+1)}; \bm{\theta}_{g}^{(t+2)}) \right\|_2 + \left \| \nabla_{\bm{\theta}_{f}} \mathcal{L}^{\text{train}}(\bm{\theta}_{f}^{(t)}; \bm{\theta}_{g}^{(t+1)}) \right\|_2 \right ) \left\| \nabla_{\bm{\theta}_{f}} \mathcal{L}^{\text{train}}(\bm{\theta}_{f}^{(t+1)}; \bm{\theta}_{g}^{(t+2)}) - \nabla_{\bm{\theta}_{f}} \mathcal{L}^{\text{train}}(\bm{\theta}_{f}^{(t)}; \bm{\theta}_{g}^{(t+1)}) \right\|_2\right]\\
\leq & \mathbb{E}\left[ 2 \rho \left\| \nabla_{\bm{\theta}_{f}} \mathcal{L}^{\text{train}}(\bm{\theta}_{f}^{(t+1)}; \bm{\theta}_{g}^{(t+2)}) - \nabla_{\bm{\theta}_{f}} \mathcal{L}^{\text{train}}(\bm{\theta}_{f}^{(t)}; \bm{\theta}_{g}^{(t+1)}) \right\|_2\right]\\
& \text{For } \mathcal{L}^{\text{train}} \text{ is Lipschitz continuous:}\\
\leq & 2 \rho L \mathbb{E}\left[ \left \|(\bm{\theta}_{f}^{(t+1)}, \bm{\theta}_{g}^{(t+2)}) - (\bm{\theta}_{f}^{(t)}, \bm{\theta}_{g}^{(t+1)}) \right \|_2\right] \\
\leq & 2 \rho L \mathbb{E}\left[ \left \|(\bm{\theta}_{f}^{(t+1)} -\bm{\theta}_{f}^{(t)}, \bm{\theta}_{g}^{(t+2)} - \bm{\theta}_{g}^{(t+1)}) \right \|_2\right] \\
= & 2 \rho L \mathbb{E}\left[ \left \| \left (- \alpha_t \left[ \nabla_{\bm{\theta}_{f}} \mathcal{L}^{\text{train}}(\bm{\theta}_{f}; \bm{\theta}_{g}^{(t+1)}) + \bm{\psi}^{(t)} \right], - \beta_t \left [ \nabla_{\bm{\theta}_{g}} \mathcal{L}^{\text{meta}}\left(\tilde{\bm{\theta}}_{f}^{(t)}(\bm{\theta}_{g})\right) + \bm{\xi}^{(t)} \right ] \right ) \right \|_2\right] \\
= & 2 \alpha_t \beta_t \rho L \mathbb{E}\left[ \left \| \left (  \nabla_{\bm{\theta}_{f}} \mathcal{L}^{\text{train}}(\bm{\theta}_{f}; \bm{\theta}_{g}^{(t+1)}) + \bm{\psi}^{(t)},  \nabla_{\bm{\theta}_{g}} \mathcal{L}^{\text{meta}}\left(\tilde{\bm{\theta}}_{f}^{(t)}(\bm{\theta}_{g})\right) + \bm{\xi}^{(t)} \right ) \right \|_2\right] \\
\leq & 2 \alpha_t \beta_t \rho L \mathbb{E}\left[ \sqrt{ \left \| \nabla_{\bm{\theta}_{f}} \mathcal{L}^{\text{train}}(\bm{\theta}_{f}; \bm{\theta}_{g}^{(t+1)}) + \bm{\psi}^{(t)} \right \|_2^2} +\sqrt{ \left \| \nabla_{\bm{\theta}_{g}} \mathcal{L}^{\text{meta}}\left(\tilde{\bm{\theta}}_{f}^{(t)}(\bm{\theta}_{g})\right) + \bm{\xi}^{(t)} \right \|_2^2} \right] \\
& \text{By the inequality} \sqrt{a} + \sqrt{b} \leq 2 \cdot \sqrt{\frac{a+b}{2}} \text{ we have:}\\
\leq & 2 \sqrt{2} \alpha_t \beta_t \rho L \sqrt{\mathbb{E}\left[ \left \| \nabla_{\bm{\theta}_{f}} \mathcal{L}^{\text{train}}(\bm{\theta}_{f}; \bm{\theta}_{g}^{(t+1)}) + \bm{\psi}^{(t)} \right \|_2^2  \right]+\mathbb{E}\left[ \left \| \nabla_{\bm{\theta}_{g}} \mathcal{L}^{\text{meta}}\left(\tilde{\bm{\theta}}_{f}^{(t)}(\bm{\theta}_{g})\right) + \bm{\xi}^{(t)} \right \|_2^2 \right] }\\
\leq & 2 \sqrt{2} \alpha_t \beta_t \rho L \sqrt{\mathbb{E}\left[ \left \| \nabla_{\bm{\theta}_{f}} \mathcal{L}^{\text{train}}(\bm{\theta}_{f}; \bm{\theta}_{g}^{(t+1)}) \right \|_2^2  \right] + \mathbb{E}\left[ \left \| \bm{\psi}^{(t)} \right \|_2^2  \right] + \mathbb{E}\left[ \left \| \nabla_{\bm{\theta}_{g}} \mathcal{L}^{\text{meta}}\left(\tilde{\bm{\theta}}_{f}^{(t)}(\bm{\theta}_{g})\right) \right \|_2^2 \right] + \mathbb{E}\left[ \left \| \bm{\xi}^{(t)} \right \|_2^2 \right] }\\
\leq & 2 \sqrt{2} \alpha_t \beta_t \rho L \sqrt{2\sigma^2+2\rho^2}\\
= & 4 \alpha_t \beta_t \rho L \sqrt{\sigma^2+\rho^2}\\
\leq & 4 \beta_1 \rho L \sqrt{\sigma^2+\rho^2} \cdot \alpha_t\\
\end{aligned}
\end{equation}

\noindent
According to the above inequality, we can conclude that our algorithm can achieve
$\lim_{t \to \infty} \mathop{\mathbb{E}} \left [ \| \nabla \mathcal{L}^{\text{train}}(\bm{\theta}_{f}^{(t)};\bm{\theta}_{g}^{(t)})\|_2^2 \right ] = 0.$

\noindent
Q.E.D. $\blacksquare$

\end{document}